%% file: emergence.tex
\documentclass[10pt]{amsart} 
\usepackage[utf8]{inputenc}

\title[Transformers as scores of entropic laws of random measures]{Emergence of the transformer architecture as the score of entropic laws of random measures}

\author{Giacomo Greco}
\address{Università degli studi di Roma Tor Vergata}
\curraddr{RoMaDS - Department of Mathematics, 00133RM Rome, Italy.}
\email{greco@mat.uniroma2.it}
\thanks{We warmly thank Domenico Marinucci for many fruitful discussions and continuous support.\newline
 While this work was written GG was associated to INdAM (Istituto Nazionale di Alta Matematica “Francesco
Severi”) and the group GNAMPA.
GG is supported by the MUR Department of Excellence Programme 2023-2027 MatMod@Tov (CUP: E83C23000330006). 
This work was partially supported by the research project METRO (Fondi di Ateneo 2024 Tor Vergata, CUP: E83C25000630005) 
and by the INdAM-GNAMPA research project ``The most likely Markovian field'' (CUP: E53C25002010001)}

\date{\today}

\input{header}

\begin{document}

\begin{abstract}
We show that the transformer architecture emerges as the score of last-token laws in a natural statistical model of language, 
where sentences are empirical measures of tokens, a language is a law on $\mcP(\mbbS^d)$, and our prior belief is that sentences carry no structure at all. 
By the maximum-entropy principle, we select the law closest to this prior in relative entropy among those reproducing an observed sentence-level statistic. 
We discuss the relevance of the statistic in the posterior interaction between tokens and we build a statistic, able to capture that interaction,
with Luce's choice axiom truncated at the lowest relational order. This procedure gives the softmax choice rule, with the attention cross-moment as its mean parameter.
The query--key matrix is then the Lagrange multiplier of a maximum-entropy problem inside each sentence, and the value matrix the multiplier of the constraint over the language, so that the two are untied. 
Under the optimal entropic law we then show that a transformer head appears in the score of the last-token law. For ordered sentences, causal masking follows from the masked reference choice rule, and with a next-token cross-moment the score equals one causally masked attention head. 
We also discuss multi-head attention, weight-tied unembedding, and a heuristic reading of residual blocks as projected Langevin steps.
\end{abstract}


\maketitle
%\tableofcontents

\section{Introduction}\label{sec:intro}

\begin{figure}[t]
  \centering
  \includegraphics[width=.85\textwidth]{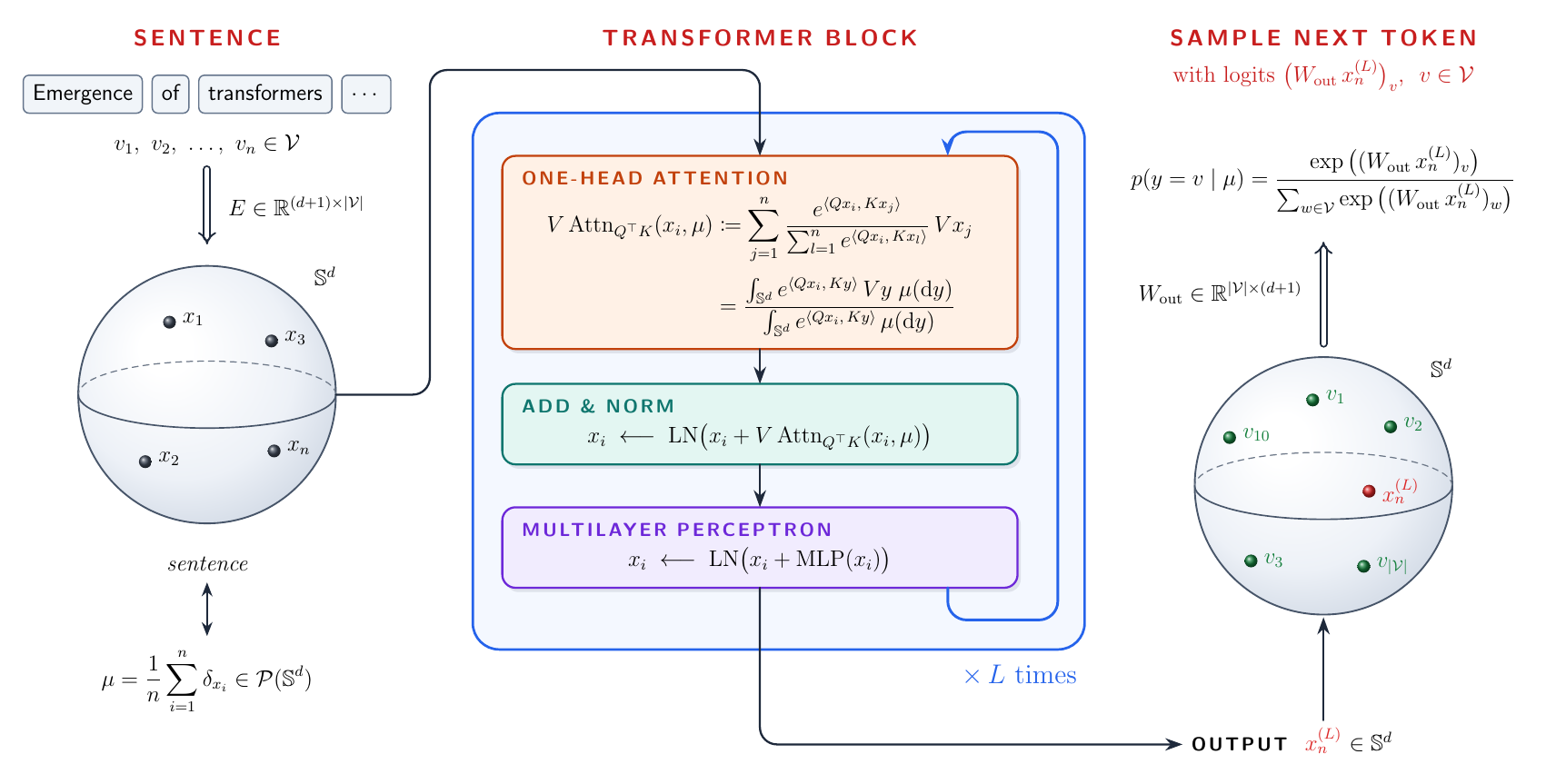}
  \caption{Transformer architecture as a three-step procedure. In the first step an embedding operator maps each token, represented by a one-hot vector in $\R^{|\mcV|}$, to a vector of $\mbbR^{d+1}$ through a learned matrix $E\in\R^{(d+1)\times|\mcV|}$, possibly combined with normalization and with a positional encoding which keeps track of token order.
Then, in the second step the sequence of vectors $x_1,\dots,x_n$ goes through a stack of $L$ identical layers. Each layer consists of a residual attention sub-layer, in which every token updates its representation by attending to the other tokens of the sentence, and a residual MLP which acts on each token separately. 
The output is then projected back to $\mbbS^d$ via $\mathrm{LN}$. 
Finally, an unembedding matrix $W_{\mathrm{out}}\in\R^{|\mcV|\times(d+1)}$ maps the output representation of the last token to a vector of logits, one per element of the vocabulary. Their softmax is the law of the next token, from which a token is sampled and appended to the text.
}
  \label{fig:pipeline}
\end{figure}

Transformers \cite{vaswani2017attention} are nowadays state-of-the-art neural networks, behind modern large language models \cite{devlin2019bert}. These networks process sentences by splitting them into 
tokens from a finite vocabulary $\mcV$, and then process them all at once in a three-stage procedure (see \cite{phuong2022formal}). 
In the first step tokens are embedded into a possibly lower-dimensional space.
Then, in the second step the corresponding sequence of vectors $x_1,\dots,x_n\in\R^{d+1}$ goes through a stack of identical layers, each consisting of a residual attention sub-layer and a residual multi-layer perceptron (MLP). 
Finally, an unembedding procedure maps the output representation of the last token to a vector of logits, one per element of the vocabulary. Their softmax is the law of the next token, from which a token is sampled and appended to the text. 
We describe this three-stage process in \Cref{fig:pipeline,fig:pipeline:masked}.

The core mechanism of the transformer architecture is the second step: the only place where tokens interact is exactly in the attention sub-layer. In its simplest form, a single attention head with query, key and value matrices $Q,K,V\in\mbbM_{d+1}$, maps the collection of tokens $x_1,\dots,x_n\in\R^{d+1}$ to

\begin{equation}\label{eq:intro:transformer}
x_i\longmapsto\sum_{j=1}^n\frac{e^{\ip{Qx_i}{Kx_j}}}{\sum_{\ell=1}^n e^{\ip{Qx_i}{Kx_\ell}}}\,Vx_j .
\end{equation}
Two structural features of this map are by now well understood. Normalization layers place tokens (up to a learned scale) on a sphere, and, in the absence of positional information,
the map is permutation-equivariant: it only sees the unordered collection of tokens. We therefore model sentences as unordered collections of points on the unit sphere $x_1,\dots,x_n\in\mbbS^d\subset\mbbR^{d+1}$, which, together with their length $n$, are in bijection with finitely supported probability measures
 on the sphere $\mu_n=\frac1n\sum_i\delta_{x_i}$.

Transformers can therefore be seen as maps between probability measures on the sphere, and their composition in layers can be seen as a discretization of a continuity equation driven by the attention vector field \cite{sander2022sinkformers,geshkovski2023emergence,geshkovski2025perspective,geshkovski2026measure}.
Most works attempt to explain why the transformer architecture works by studying these dynamics, describing how transformers work for a fixed given attention mechanism. Here we would like to explain why transformers work with a complementary statistical question:
\medskip
\begin{center}
\emph{Where does the transformer architecture come from? Which probabilistic model of language has softmax self-attention, with its query--key matrix $\Lambda=Q^\top K$ and its value matrix $V$, as its natural inference rule, and which variational principle trains these two matrices?}
\end{center}
\medskip

In light of token normalization and permutation-equivariance, our answer is formulated at the level of laws of measures, which leads us to an entropic minimization problem on $\mcP(\mcP(\mbbS^d))$.
In particular, we posit that there exists an underlying, unobserved \emph{language law} governing how
sentences/measures $\mu$ arise. Formally, this is a probability measure $\mbfP^\star \;\in\; \mcP(\mcP(\mbbS^d))$
on the space of probability measures on $\mbbS^d$, a distribution over distributions. Therefore each sentence can be considered as a random (atomic) probability measure over the token
space and sampling one sentence amounts to generating a sample from $\mbfP^\star\in\mcP(\mcP(\mbbS^d))$.

\medskip

In what follows we focus on the core step of the transformer architecture and directly assume that tokens are elements of $\mbbS^d$ (i.e., the initial embedding is the identity). We postpone the discussion of embedding and unembedding steps to \Cref{sec:unembedding}.
We first address the unordered model, where tokens within a sentence do not carry positional information (as in \Cref{fig:pipeline}), while the causal model with positional encoding is treated in \Cref{sub:intro:ordered} (see also \Cref{fig:pipeline:masked}). 
Lastly, the multi-head extension of the transformer architecture is discussed in \Cref{sec:multihead} (see also \Cref{fig:multihead}).

\subsection{The choice of the prior} In the absence of data, our prior belief is that a sentence carries no structure at all. 
Therefore, in order to sample a sentence in an uninformative way, it is enough to consider a random length $N\sim\nu$ sampled according to a probability measure on $\{1,2,\dots\}$ 
(e.g., a Poisson distribution shifted by one), together with an independent collection of tokens $X_1,\,X_2,\dots \overset{i.i.d.}{\sim}\underline{\vol}$, and consider the corresponding sentence $\tfrac{1}{N}\sum_{i=1}^N\delta_{X_i}$.
Our prior will be the law of a random sentence/measure, built in this way
\begin{equation}
  \label{eq:prior}
  \mbfN\;\coloneqq\; \mathrm{Law}\Big(\tfrac1N\sum_{i=1}^N\delta_{X_i}\Big)\;\in\;\mcP(\mcP(\mbbS^d)),\qquad N\sim\nu,\qquad X_1,X_2,\dots\overset{iid}\sim\uvol,\ \text{independent of }N\,.
\end{equation}
Here $\uvol$ is the normalized volume, the only rotation-invariant probability measure on $\mbbS^d$. Its predictive law is trivial (because of independence):
\begin{equation}\label{eq:BM-intro}
  X_{n+1} \mid X_1,\dots,X_n,\ N> n \;\sim\;\underline{\vol},\qquad \mbbP^{\mbfN}\big(N=n\,\big|\,X_{1:n},\ N\ge n\big)=\frac{\nu(n)}{\nu([n,\infty))}.
\end{equation}
The next token is a fresh uniform draw, independent of the context, and the sentence stops with the hazard rate of $\nu$, independently of its tokens.

\medskip

In this paper we look for the minimal modification of this null model, in relative entropy (a.k.a. Kullback--Leibler divergence),
that reproduces a vector of sentence-level statistics $\Phi=(\varphi_1,\dots,\varphi_k):\mcP(\mbbS^d)\to\R^k$ observed in data:
\begin{equation}\label{eq:entropic-intro}
\text{minimize } \rel{\mrmP}{\mbfN}\quad\text{over } \mrmP\in\mcP(\mcP(\mbbS^d))\quad\text{subject to}\quad \mbbE_\mrmP[\Phi(\mu)]=\mbbE_{\mbfP^\star}[\Phi(\mu)] \,,
\end{equation}
where $\KL$ is the Kullback--Leibler divergence, defined for any couple of probabilities $\mrmP,\,\mrmQ$ as $\rel{\mrmP}{\mrmQ}:=\mbbE_\mrmP[\log\tfrac{\De\mrmP}{\De\mrmQ}]$ whenever $\mrmP\ll\mrmQ$, and set equal to $+\infty$ otherwise.
The solution is the Gibbs tilt $\mrmP^\Phi\propto e^{\ip{V^\Phi}{\Phi}}\mbfN$ \cite{csiszar1975}, for some vector of Lagrange multipliers $V^\Phi\in\R^k$, as soon as $\Phi$ is bounded, continuous and non-degenerate and the constraint lies in the interior of the convex support of $\Phi$ under $\mbfN$ (\Cref{thm:iproj}).
In particular, this means that under the Gibbs tilt $\mrmP^\Phi$, see \Cref{prop:last-token}, the last-token predictive law reads as
\begin{equation*}
\mrmP^\Phi\big(X_{n}\in\mathrm{d}y\,\big|\,N=n,\ X_{1:n-1}=x_{1:n-1}\big)
\,\propto\,\exp\big(\langle V^\Phi,\,\Phi(\mu_y)\rangle\big)\,\underline{\vol}(\mathrm{d}y)\,,\quad\text{with }\mu_y\coloneqq \tfrac1n\big(\sum_{i=1}^{n-1}\delta_{x_i}+\delta_y\big)\,.
\end{equation*}
Therefore, in this model, we could sample the last token via a Langevin sampler associated to the score of its predictive law
\begin{equation}\label{eq:intro:score:last:token}
\nabla_y\log p(y\mid X_{1:n-1}=x_{1:n-1},\,N=n)=\nabla_y\langle V^\Phi,\,\Phi(\mu_y)\rangle=\frac1n\,P^\perp_y\,\nabla_z\,\langle V^\Phi,\,\delta\Phi(\mu_y)\rangle\,(z)\Big|_{z=y}\,,
\end{equation}
where $\delta\Phi=(\delta\varphi_1,\,\dots,\,\delta\varphi_k)$ denotes the first variation of the functional $\Phi$ and $\nabla_z\langle V^\Phi,\,\delta\Phi(\mu)\rangle(z)$ its gradient.
This shows that the whole content of the model lies in the choice of the statistic $\Phi$, and that its first variation is what should capture how tokens collectively behave and attend to each other within each sentence.

\subsection{The choice of relevant statistics for the semantic structure of tokens}\label{sec:intro:choice}
The semantic structure of tokens cannot be captured by the null model, where the tokens of a sentence are i.i.d.\ draws from the uniform measure $\uvol$.
Statistics depending on $\mu$ only through its weights detect only the length of the sentence. Linear statistics $\Phi(\mu)=\int h\,\De\mu$ do not create interactions either: given $N=n$, the tilt $e^{\ip{V^\Phi}{\Phi}}$ factorizes as $\prod_{i\le n}e^{\ip{V^\Phi}{h(x_i)}/n}$, so the tokens remain independent and the next token does not depend on the context. Interaction between tokens therefore requires statistics that involve at least pairs of tokens of the same sentence.
Among these, the most natural choice would be the covariance among tokens $\Phi(\mu)=\Cov(\mu)$; however, as we show in \Cref{prop:additive},
any statistic that is additive over pairs (i.e.\ of the form
$\iint h\,\De\mu\,\De\mu=n^{-2}\sum_{i,j}h(X_i,X_j)$) would produce reinforcements in which every key
contributes with a fixed weight that does not depend on the other keys: there is no competition among keys. For moment statistics such as the covariance, the query even attends to the context only through the empirical mean of the other tokens. In particular, in the covariance case we would have 
\begin{equation*}
\nabla_z\langle V^\Phi,\,\delta\Phi(\mu_y)\rangle\,(z)\Big|_{z=y}=\frac{n-1}{n}\,(V^\Phi+(V^\Phi)^\top)\big(y-\tfrac1{n-1}\sum_{i=1}^{n-1}x_i\big)\,,
\end{equation*}
which shows that the previous tokens influence the next token $y$ only via their empirical mean, and the next token candidate $y$ does not attend to each key individually.

On the contrary, linguistic structure is relational: the role of a token is determined by which other tokens of the sentence it relates to, that is, 
by a choice among competitors. We model this as a random choice of a key by a query, a \emph{choice rule} for the sentence $\mu$, which we model as a coupling in
\[
\Pi_1(\mu)=\big\{\pi\in\mcP(\mbbS^d\times\mbbS^d):\ \pi(\cdot\times\mbbS^d)=\mu\big\},
\]
the first coordinate being the query and the second the key; we write $\pi(\De x,\De y)=\mu(\De x)\,\pi_x(\De y)$. Keys are tokens of the sentence: $\pi_x\ll\mu$ for $\mu$-almost every $x$, which is forced as soon as $\rel{\pi}{\mu\otimes\mu}<\infty$. The uniform choice $\mu\otimes\mu$, in which the key is drawn from the sentence irrespective of the query, models the choice rule without relational content.

The weakest form of competition we can consider in our model is Luce's choice axiom \cite{luce1959individual}. In this model, 
the odds of choosing the key $y$ rather than $y'$ depend on the query and on the two keys, and not on the other tokens of the sentence. 
In our model, for a finite sentence with positive choice probabilities, Luce's choice axiom holds if and only if there is a \emph{choice function} $h:\mbbS^d\times\mbbS^d\to\R$ \cite[Chapter~1, Theorem~3]{luce1959individual}, the same for all sentences, such that for any sentence $\mu$ its associated choice rule equals
\begin{equation}\label{eq:score-rule}
\pi^h_\mu(\De x,\De y)=\mu(\De x)\,\frac{e^{h(x,y)}\,\mu(\De y)}{\int e^{h(x,y')}\,\mu(\De y')}\,.
\end{equation}
The choice function should encode relational preferences and syntactical relations between tokens, and in particular different choice functions might focus on different aspects of our language.

A priori the choice function $h$ could be highly non-linear, 
with spherical harmonics decomposition in $\mrmL^2(\uvol\otimes\uvol)$ equal to $h=\sum_{\ell,\ell'\ge0}h_{\ell\ell'}$ with $h_{\ell\ell'}\in\msY_\ell\otimes\msY_{\ell'}$, where $\msY_\ell$ denotes the set of spherical harmonics of degree $\ell$ on $\mbbS^d$.
In \Cref{prop:harmonic}, we argue that the lowest-order term in this expansion that captures tokens' relations, is $h_{11}\in\msY_1\otimes\msY_1=\{(x,y)\mapsto x^\top\Lambda y:\ \Lambda\in\Mat\}$, which corresponds to a bilinear form  induced by a matrix $\Lambda\in\mbbM_{d+1}$.
Therefore, in view of implementation, we approximate the choice function $h$ with $h_{11}=\langle x,\Lambda y\rangle$ and consider choice rules induced by the exponential bilinear kernel
$k_\Lambda(x,y)=e^{x^\top\Lambda y}$ (\Cref{prop:harmonic}). We refer to the matrix $\Lambda$ as the choice matrix or attention matrix. The corresponding choice rule is then given by the Gibbs coupling
\begin{equation}\label{eq:rule-intro}
\pi^\Lambda_\mu(\De x,\De y)=\mu(\De x)\,\frac{k_\Lambda(x,y)\,\mu(\De y)}{Z^\Lambda_\mu(x)},\qquad Z^\Lambda_\mu(x)=\int k_\Lambda(x,y)\,\mu(\De y)\,.
\end{equation}

In light of this choice rule, we consider covariance statistics associated to $\pi^\Lambda_\mu$. 
More precisely, in this paper the statistic of interest is the attention cross-moment 
\begin{equation}\label{eq:S-intro}
S_\Lambda(\mu)\coloneqq\mbbE_{\pi^\Lambda_\mu}[XY^\top]=\mbbE_\mu[ X\,\Attn_\Lambda(X;\mu)^\top]\in\mbbM_{ d+1}\,, 
\end{equation}
where the attention vector field is defined via conditional expectation 
\begin{equation}
\label{eq:attn-intro}
\Attn_\Lambda(x;\mu)\coloneqq\mbbE_{\pi^\Lambda_\mu}[Y|X=x]=\frac{\int y\,k_\Lambda(x,y)\,\mu(\De y)}{Z^\Lambda_\mu(x)}\,.
\end{equation}
For each fixed sentence $\mu$, the attention cross-moment $S_\Lambda(\mu)$ is a matrix-valued statistic, whose entries are the (non-centered) cross-moments between the query and key directions of the sentence under the choice rule $\pi^\Lambda_\mu$.
It is the mean parameter (the expectation of the sufficient statistic $xy^\top$) of the conditional exponential family of choice rules $\{\pi^\Lambda_\mu\}_\Lambda$, and it is therefore a natural choice for capturing the relational structure of tokens in a sentence. 
Moreover, the choice rule $\pi^\Lambda_\mu$ is the least informative choice rule with its prescribed cross-moment $S_\Lambda(\mu)$ fixed, 
that is a conditional entropic projection as in conditional maximum-entropy models of natural language \cite{berger1996maximum}, with $\Lambda$ the Lagrange multiplier (\Cref{prop:condmaxent}).
Finally, we take the last term appearing in~\eqref{eq:attn-intro} as a definition for the attention vector field, whenever $x\not\in\supp(\mu)$.

\medskip

In this paper the statistic of \eqref{eq:entropic-intro} is the attention cross-moment, $\Phi=S_\Lambda$. Its Lagrange multiplier is a matrix $V^\star=V^\star(\Lambda)\in\mbbM_{d+1}$, and the minimizer reads
\begin{equation}\label{eq:model-intro}
\mrmP^{S_\Lambda}=\mrmP_{\Lambda,V^\star},\qquad \mrmP_{\Lambda,V^\star}\propto\exp\big(\ip{V^\star}{S_\Lambda}\big)\,\mbfN\,.
\end{equation}
With this choice attention appears twice: inside the statistic, and in the score of the law of new tokens (\Cref{thm:last}), whereas $V^\star$ corresponds to the value matrix (\Cref{thm:last}-\ref{item:last:att:c}).  In particular, the readout
\begin{equation*}
\ip{V^\star}{S_\Lambda(\mu)}=\mbbE_\mu[ X^\top\,V^\star\,\Attn_\Lambda(X;\mu)]=\int x^\top V^\star\,\Attn_\Lambda(x;\mu)\,\mu(\De x)
\end{equation*}
is the average, over the tokens $x$ of the sentence, of the pairing between $x$ and the output, through $V^\star$, of the attention head whose query is $x$. 
On a generic sentence in our model $\mu=\tfrac1n\sum_{i=1}^n\delta_{x_i}$, for an attention matrix $\Lambda=Q^\top\,K$, the readout explicitly becomes 
\begin{equation*}
\langle V^\star,\,S_{\Lambda}\big(\tfrac1n\sum_{i=1}^n\delta_{x_i}\big)\rangle=\tfrac1{n}\sum_{i=1}^n\langle x_i,\,V^\star\,\Attn_\Lambda(x_i;\mu)\rangle=\tfrac1{n}\sum_{i=1}^n\Big\langle  x_i,\,\sum_{j=1}^n\frac{e^{\ip{Qx_i}{Kx_j}}}{\sum_{\ell=1}^n e^{\ip{Qx_i}{Kx_\ell}}}\,V^\star x_j \Big\rangle\,,
\end{equation*}
and $V^\star\,\Attn_\Lambda(x_i;\mu)$ corresponds exactly to the transformer mechanism \eqref{eq:intro:transformer} with value matrix $V^\star$ and with query matrix $Q$ and key matrix $K$ (with $\Lambda=Q^\top\,K$). The same transformer architecture will appear as the leading (query) term of the 
score of the posterior distribution \eqref{eq:intro:score:last:token} which can then be used in a Langevin sampler for the last-token generation. 

\medskip

Let us conclude by noticing that a different statistic induced by the choice rule can be considered. 
Indeed, $\pi^\Lambda_\mu$ can be variationally characterized as the optimizer of an entropy-regularized problem (cf.~\Cref{prop:gibbs-attn}), whose optimal value equals the attention free energy $F_\Lambda(\mu)=\int\log Z^\Lambda_\mu\,\mathrm{d}\mu$.
The attention cross-moment corresponds to the gradient in $\Lambda$ of this scalar statistic (\Cref{prop:gibbs-attn}-\ref{item:attention:EOT:b}). 
In \Cref{rem:free-energy} we also consider a different model in which the language is constrained through $F_\Lambda$ instead of $S_\Lambda$. There, the Gibbs tilt \eqref{eq:model-intro} is replaced with $e^{\lambda^\star F_\Lambda}\mbfN$, 
which as a side-effect necessarily ties the value matrix to the attention matrix $V=\lambda^\star\Lambda$.  Hence, in this case the values correspond to the keys, 
whereas in the cross-moment model the value matrix $V$ is free. 
We refer the reader to \Cref{rem:free-energy}, where we discuss what changes in the free-energy setting.

\bigskip

We can summarize what we have discussed so far and will show below in the following informal statement.

\begin{metathm}[Unordered model]
Let $\mbfN\in\mcP(\mcP(\mbbS^d))$ be the structureless null model (a random number $N\sim\nu$ of i.i.d.\ uniform tokens on $\mbbS^d$) and consider the $\KL$ minimization problem
\begin{equation*}
\text{minimize } \rel{\mrmP}{\mbfN}\quad\text{over } \mrmP\in\mcP(\mcP(\mbbS^d))\quad\text{subject to}\quad \mbbE_\mrmP[\Phi(\mu)]=C_{\Phi}\,,
\end{equation*}
with $C_\Phi$ being the empirical value of the statistic $\Phi$ observed from data samples. Assume that $\nu$ charges at least one length $n\geq 3$. Then, the following statements hold true:
\begin{enumerate}[(i)]
\item \emph{Why attention.} Statistics $\Phi$ that are polynomial in sentences $\mu\in\mcP(\mbbS^d)$, or additive over pairs of tokens, give reinforcements without competition between keys. 
Statistics induced by choice rule couplings~\eqref{eq:rule-intro} preserve relational structure and the lowest-order relational choice rules correspond to couplings
 $\pi^\Lambda_\mu(\De x,\De y)=\mu(\De x)\,e^{\langle x,\Lambda y\rangle}\mu(\De y)/Z^\Lambda_\mu(x)\in\Pi_1(\mu)$ induced by attention choice matrices $\Lambda\in\mbbM_{d+1}$. 
The corresponding mean parameter is then the attention cross-moment $S_\Lambda$, which is non-degenerate for $\Lambda\neq0$.
\item \emph{Why this model.} For any fixed attention choice matrix $\Lambda\in\mbbM_{d+1}$, the minimal modification of $\mbfN$ in relative entropy that reproduces $\mbbE[S_\Lambda]$ is the Gibbs tilt 
$\mrmP_{\Lambda,V^\star}\propto e^{\langle V^\star,S_\Lambda(\mu)\rangle}\mbfN$, with value matrix $V^\star$ being the corresponding Lagrange multiplier (for $\Lambda\neq0$ and an interior constraint, \Cref{thm:fixed}). 
In particular query--key attention and value matrices are untied.
\item \label{item:meta:sym:iii}\emph{Transformer as last-token prediction score.} Conditionally on $N=n$ and on the first $n-1$ tokens $X_{1:n-1}=x_{1:n-1}$, 
under $\mrmP_{\Lambda,V^\star}$, the law of the last token $X_n$ has density, with respect to $\uvol$,
\begin{equation*}
p(y)\;\propto\;\exp\big(\langle V^\star,S_\Lambda(\mu_y)\rangle\big)\,,\quad\text{ with }\mu_y=\frac1n\big(\sum_{j<n}\delta_{x_j}+\delta_y\big)\,.
\end{equation*}
Its score on $\mbbS^d$ is
\begin{equation*}
\nabla_{\mbbS^d}\log p(y)=\frac1n\,P_y^\perp\,\nabla_z\,\langle V^\star,\delta S_\Lambda(\mu_y)\rangle\,(z)\Big|_{z=y},
\qquad P_y^\perp=\mathrm{Id}-yy^\top,
\end{equation*}
where (\Cref{thm:last})
\begin{equation*}
\nabla_z\,\langle V^\star,\delta S_\Lambda(\mu_y)\rangle\,(z)= V^\star\,\mathrm{Attn}_\Lambda(z;\mu_y)
+\int \frac{e^{\langle x,\Lambda z\rangle}}{\int e^{\langle x,\Lambda z'\rangle}\mu_y(\mathrm dz')}\,V^{\star\top}\,x\,\mu_y(\mathrm dx)+\rho_{\mu_y}(z).
\end{equation*}
The first term is the transformer architecture~\eqref{eq:intro:transformer}, the second is a key-side term, and $\rho_{\mu_y}$ is the correction due to the dependence of the attention weights on $z$.
The score is therefore $n^{-1}$ times the tangential part of the full gradient of the first variation of $\langle V^\star,S_\Lambda\rangle$ at the completed sentence $\mu_y$.
\end{enumerate}
\end{metathm}

The first term, $V^\star\,\mathrm{Attn}_\Lambda(z;\mu_y)$, is the query part: the last token $z$ is pulled 
toward what it reads from the sentence, a softmax-weighted average of the other tokens mapped by $V^\star$. 
The second is the key part: $z$ is also pulled toward the tokens that attend to it, weighted by how much attention each of them pays to $z$. 
The third, $\rho_{\mu_y}(z)$, is the correction due to the attention weights themselves depending on $z$: moving $z$ changes where $z$ looks,
and where the other tokens look, and this changes the readout. The presence of these terms is a
consequence of the candidate being, in the statistic, both a key of the other tokens and a query: permutation invariance symmetrizes the roles of query and key.

\subsection{The choice of the optimal query--key attention matrix $\Lambda^\star$ and the value matrix $V^\star$}
The query--key attention matrix $\Lambda$ and the value matrix $V^\star$ play different roles. Inside a sentence, $\Lambda$ plays the role of the Lagrange multiplier of 
the choice problem (\Cref{prop:condmaxent}): it decides which keys a query selects. It is fixed language-wide, and it cannot be recovered from a single sentence. 
Over the language, $V^\star$ is the Lagrange multiplier of the same cross-moment averaged 
over sentences (\Cref{thm:fixed}): it decides which pairings between a token and the content its attention head reads are over-represented in the language, relative to the null model.
Query--key attention and value matrices are therefore the inner and outer multipliers of the same cross-moment.

Both matrices are learned from the language by one variational principle: we aim at maximizing the expected log-likelihood ratio of the model against the null model,
\begin{equation}\label{eq:loglik-intro}
\mcL(\Lambda,V)=\mbbE_{\mbfP^\star}\big[\ip{V}{S_\Lambda}\big]-\log\mbbE_{\mbfN}\big[e^{\ip{V}{S_\Lambda}}\big]\,.
\end{equation}
Equivalently, we look for $(\Lambda,V)$ optimizing the minimax-entropy principle \cite{zhu1997minimax}
\[
\sup_{\Lambda,V}\mcL(\Lambda,V)=\sup_\Lambda\ \inf_{\mrmP:\ \mbbE_\mrmP S_\Lambda=\mbbE_{\mbfP^\star}S_\Lambda}\rel{\mrmP}{\mbfN}\,.
\]
In this way, the query--key attention matrix selects the attention cross-moment that is least compatible with the null model, and the model is the least informative one that reproduces it. 
This principle was introduced in natural language processing long before large language models, in the maximum-entropy models of \cite{berger1996maximum,rosenfeld1996maximum}. 
We combine this principle with the minimax-entropy models of \cite{zhu1997minimax}: the second term in the log-likelihood \eqref{eq:loglik-intro} 
penalizes readouts that are already large under the null model: 
a readout that is equally large on the data and on structureless sentences gains nothing.  
In this paper the matrix $\Lambda$ is kept fixed: we study the model $\mrmP_{\Lambda,V^\star}$, its predictive laws and its samplers, and we only discuss the training problem for the optimal attention matrix $\Lambda$ in \Cref{rem:training}.

\subsection{Recovering the exact transformer architecture: positions, causal masking and next-token readout}\label{sub:intro:ordered}
By the Meta-Theorem for the unordered model, the score of the last token contains the transformer architecture \eqref{eq:intro:transformer} as its query part, 
but it also contains the key-side term and the correction $\rho_{\mu_y}$. 
The key-side term, and the part of $\rho_{\mu_y}$ coming from the other queries, appear because the tokens of the sentence attend to $y$ 
as one of their keys. The rest of $\rho_{\mu_y}$ appears because $y$ is itself a query, whose attention weights move with $y$.

Decoder transformers, which generate text, switch off both mechanisms, and we now show how to do the same in our setting.
In practice transformers keep track of the position of each token, and the attention is causally masked \cite{vaswani2017attention}: 
a token only attends to tokens that come before it, and to itself, and never sees future tokens. 
Order cannot be produced by our null model $\mbfN$, which is exchangeable, nor by a statistic of the sentence $\mu$, 
which is invariant under permutations of the tokens. 
Therefore our sentences must remember the positions of their tokens. 
One way of doing this is by lifting our sentence space to $\mcP(\mbbS^d\times \mbbN)$, the space of ordered sentences
\begin{equation*}
\overset{\rightarrow}{\mu}=\frac1N\sum_{i=1}^N\delta_{(X_i,i)}\in\mcP(\mbbS^d\times\N)\,,
\end{equation*}
where we record each token together with its position. Clearly this measure is in bijection with the ordered sequence $(X_1,\dots,X_N)$, and its image under $(x,i)\mapsto x$ is our unordered sentence $\mu=\tfrac1N\sum_{i=1}^N\delta_{X_i}\in\mcP(\mbbS^d)$.

The null model in this ordered setting becomes the law $\arrowN\in\mcP(\mcP(\mbbS^d\times\N))$ of $\arrowmu$, 
under which the length and the ordered tokens have the law $\nu(n)\,\uvol^{\otimes n}$ on $\{N=n\}$.

\medskip

In practice, in order to prevent tokens from seeing the future, a causal masking mechanism 
is implemented in the attention softmax mechanism, which switches to zero all the attention weights appearing in~\eqref{eq:intro:transformer} 
when the key token $(X_j,j)$ comes after the query token $(X_i,i)$ in our sentence, i.e., for $i<j$. 
In our model, the choice rule is the only place where relational structure is encoded; thus in order to enforce this property we should introduce the masking there.
As a consequence, now Luce's axiom should hold among the competitors that are not later than the query, and the uniform choice $\mu\otimes\mu$ should be replaced by the uniform choice among the keys $(y,j)$ with $j\le i$, for the query $(x,i)$. 
Then, by arguing as in the unordered model, we are led to consider the lowest-order relational choice rule as the causal Gibbs coupling
\begin{equation}\label{eq:causal-rule-intro}
\pi^{\Lambda,\mathrm M}_{\arrowmu}(\De x\,\De i,\De y\,\De j)\coloneqq \arrowmu(\De x\,\De i)\,\frac{\mathds1_{j\le i}\,e^{\ip{x}{\Lambda y}}\,\arrowmu(\De y\,\De j)}{\int\mathds1_{j'\le i}\,e^{\ip{x}{\Lambda y'}}\,\arrowmu(\De y'\,\De j')}=\sum_{i=1}^n\sum_{j=1 }^n\Pi^{\mathrm M}_{ij}\;\delta_{(x_i,i)}\otimes\delta_{(x_j,j)}
\;\in\mcP\big((\mbbS^d\times\mbbN)^2\big)\,,
\end{equation}
with $\Pi^{\mathrm M}_{ij}\coloneqq n^{-1}\frac{\mathds{1}_{j\leq i}e^{\ip{x_i}{\Lambda x_j}}}{\sum_{\ell=1}^{n}\mathds{1}_{\ell\leq i}e^{\ip{x_i}{\Lambda x_\ell}}}$.
Notice that this coupling can equivalently be rewritten as 
\begin{equation}\label{eq:causal-rule-intro:again}
\pi^{\Lambda,\mathrm M}_{\arrowmu}=\sum_{i=1}^n\sum_{j=1}^n\,\frac1n\,\frac{e^{\ip{x_i}{\Lambda x_j}+M_{ij}}}{\sum_{\ell=1}^ne^{\ip{x_i}{\Lambda x_\ell}+M_{i\ell}}}\;\delta_{(x_i,i)}\otimes\delta_{(x_j,j)}\,,
\end{equation}
where $M\in\{0,-\infty\}^{n\times n}$ is the causal additive mask defined as $M_{ij}= (-\infty)\,\mathds1_{i<j}$,
with the convention $e^{-\infty}=0$.
We refer to the above as the masked choice rule, which is again the least informative rule with its query--key cross-moment, 
the masked entries being forced by the finiteness of the relative entropy with respect to the masked uniform choice, in which each query chooses uniformly among the tokens not later than itself (\Cref{prop:causal-rule}).

Under this choice rule the attention that the token $(x_i,i)$ gives to the already seen tokens, the attention vector field, is defined as the conditional expectation
\begin{equation*}
  \begin{aligned}
\Attn_\Lambda((x_i,i);\arrowmu)\coloneqq\mbbE_{((X,I),(Y,J))\sim\pi^{\Lambda,\mathrm M}_{\arrowmu}}\big[\,Y\,\big|\,(X,I)=(x_i,i)\big]=\sum_{j\le i}\frac{e^{\ip{x_i}{\Lambda x_j}}}{\sum_{\ell\le i}e^{\ip{x_i}{\Lambda x_\ell}}}\,x_j\\
=\sum_{j=1}^n\frac{e^{\ip{x_i}{\Lambda x_j}+M_{ij}}}{\sum_{\ell=1}^ne^{\ip{x_i}{\Lambda x_\ell}+M_{i\ell}}}\,x_j\,,
\end{aligned}
\end{equation*}
and we take the latter expression as a definition for any $(x,i)\not\in\supp(\arrowmu)$.

\medskip

We can then associate to this choice rule the cross-moment statistic. However, the attention cross-moment considered in~\eqref{eq:S-intro}
pairs each token with what it reads; for generation, instead, we are interested in how well the next token is predicted by what the current token reads, i.e., its attention vector field. 
That is, we are interested in the cross-moment between the next token $X_{i+1}$ and the key chosen by $X_i$ according to the choice rule, namely $\mbbE_{(X,Y)\sim\pi^{\Lambda,\mathrm M}_{\arrowmu}}[Y|X=X_i]$.
 This leads, for $n\geq 2$, to the shifted masked cross-moment

\begin{equation}\label{eq:shifted:cross-moment}
S^{\to}_\Lambda(\arrowmu)=\frac1{n-1}\sum_{i=1}^{n-1}x_{i+1}\,\Attn_\Lambda((x_i,i);\arrowmu)^\top=\mbbE_{\pi^{\Lambda,\rightarrow}_{\arrowmu}}[XY^\top]\,,
\end{equation}
where we have introduced the next-token coupling  $\pi^{\Lambda,\rightarrow}_{\arrowmu}=(\iota_{\arrowmu}\times\iota_{\arrowmu})_\#\big(\tfrac{n}{n-1} J\Pi^{\mathrm M}\big)$ (cf.~\eqref{eq:shifted-coupling}), with 
$\iota_{\arrowmu}(i)=(x_i,i)$ being the lift of positions to tokens in $\arrowmu$, $J\in\mbbM_{n}$ the lower shift matrix, $J_{ij}=\mathds 1_{\{i=j+1\}}$,
 and $\Pi^{\mathrm M}=(\Pi^{\mathrm M}_{ij})\in\mbbM_n$ is the matrix associated to $\pi^{\Lambda,\mathrm M}_{\arrowmu}$, so that $\pi^{\Lambda,\mathrm M}_{\arrowmu}=(\iota_{\arrowmu}\times\iota_{\arrowmu})_\#\Pi^{\mathrm M}$.

Two reasons support this choice. Firstly, it is a classical feature of the maximum-entropy language models of \cite{rosenfeld1996maximum,berger1996maximum} 
to pair a summary of the history with the next word, and 
the shifted cross-moment is their relational version. The history up to position $i$ is summarized by what $x_i$ reads 
through the attention vector field $\Attn_\Lambda((x_i,i);\arrowmu)$, and it is paired with the next token $x_{i+1}$. 
By contrast, $S_\Lambda$ describes the sentence as a bag of tokens.
Secondly, no token is paired with what it reads itself, whereas in $S_\Lambda$ each token is paired with its own read, which contains the token itself (\Cref{rem:shifted-selfpair}).

The masked choice rule is unchanged: by \Cref{prop:causal-rule} it is still the least informative masked rule given its own query--key cross-moment $\frac1n\sum_ix_i\Attn_\Lambda((x_i,i);\arrowmu)^\top$, and $\Lambda$ still plays the role of the corresponding Lagrange multiplier inside each sentence. 
What changes is the pairing of its output at the level of the language: the statistic $S^\to_\Lambda$ pairs what $x_i$ reads with the next token $x_{i+1}$ instead of with $x_i$ itself. 
For every $V$, the model $\mrmP^\to_{\Lambda,V}\propto e^{\ip{V}{S^\to_\Lambda}}\arrowN$ is the entropic projection of the null model onto its own mean of $S^\to_\Lambda$, 
with $V$ as the Lagrange multiplier (\Cref{sec:shifted-def}), and we denote by $V^\star=V^\star(\Lambda)$ the multiplier associated to our constraint $\mbbE_{\mrmP}[S^\rightarrow_\Lambda]=C\in\mbbM_{d+1}$, with the latter matrix being what we observe from data samples.

\bigskip

With this statistic the transformer architecture appears exactly, without correction terms. Indeed, for $n\geq 2$, conditionally on $N=n$ and on the first $n-1$ tokens $X_{1:n-1}=x_{1:n-1}$, 
we will show in \Cref{thm:shifted-last} that the last-token predictive law equals
\begin{equation*}
\mrmP^{\to}_{\Lambda,V^\star}\big(X_n\in\De y\,\big|\,N=n,\ X_{1:n-1}=x_{1:n-1}\big)\;\propto\;\exp\Big(\frac1{n-1}\,y^\top V^\star\Attn_\Lambda((x_{n-1},n-1);\arrowmu)\Big)\,\uvol(\De y)\,,
\end{equation*}
a von Mises--Fisher law with mean direction vector $V^\star\Attn_\Lambda((x_{n-1},n-1);\arrowmu)$ (normalized) and concentration parameter $\norm{ V^\star\Attn_\Lambda((x_{n-1},n-1);\arrowmu)}/(n-1)$. 
The score associated to this distribution then simply corresponds to the projection onto the tangent space $T_y\mbbS^d$ of the von Mises--Fisher mean direction vector,
and if we decompose the attention matrix as $\Lambda=Q^\top K$ we get
\begin{equation}\label{eq:shifted-score-intro}
\nabla_{\mbbS^d}\log p(y)=\frac1{n-1}\,P^\perp_y\,V^\star \Attn_\Lambda((x_{n-1},n-1);\arrowmu)=\frac1{n-1}\,P^\perp_y\sum_{j=1}^{n}\frac{e^{\ip{Qx_{n-1}}{Kx_j}+M_{n-1,j}}}{\sum_{\ell=1}^{n}e^{\ip{Qx_{n-1}}{Kx_\ell}+M_{n-1,\ell}}}\,V^\star x_j\,.
\end{equation}
This is exactly one causally masked attention head \eqref{eq:intro:transformer} (e.g., \cite[Figure 2]{vaswani2017attention},\cite[Algorithm 4]{phuong2022formal}), with query matrix $Q$, key matrix $K$ and value matrix $V^\star$, evaluated at the last position of the context, followed by the projection onto the tangent space of the sphere. 
There is no key-side term and no correction term, because the candidate $y$ is neither a key nor a query of the statistic: the mask removes it from the keys of every query, and the shifted pairing never uses it as a query.

\bigskip

We can summarize this part in the following informal statement.

\begin{metathm}[Causal model]
Let $\arrowN\in\mcP(\mcP(\mbbS^d\times\N))$ be the structureless null prior with positions (law of ordered sentences $\tfrac1N\sum_{i=1}^N\delta_{(X_i,i)}$, with $N\sim\nu$ and independent $X_1,X_2,\dots\overset{i.i.d.}{\sim}\uvol$), and consider the $\KL$ minimization problem 
\begin{equation}\label{eq:meta:min:ord}
\text{minimize } \rel{\mrmP}{\arrowN}\quad\text{over } \mrmP\in\mcP(\mcP(\mbbS^d\times\N))\quad\text{subject to}\quad \mbbE_\mrmP[S^\rightarrow_\Lambda(\arrowmu)]=\widehat{S^\rightarrow_\Lambda}\,,
\end{equation}
with $\widehat{S^\rightarrow_\Lambda}$ being the empirical shifted masked cross-moment observed from samples (or its true mean value~$\mbbE [S^\rightarrow_\Lambda]$). Assume that $\nu$ charges at least one length $n\ge2$. Then, the following statements hold true:
\begin{enumerate}[(i)]
\item \label{item:meta:ord:a} \emph{Causal choice rule.} Assume Luce's axiom among the competitors not later than the query, with a choice function that does not depend on positions. 
With the mask in the reference coupling, the lowest-order relational choice rule is causally masked softmax attention  $\pi^{\Lambda,\mathrm M}_{\arrowmu}$ defined in ~\eqref{eq:causal-rule-intro}. 
It is the unique minimizer of the relative entropy with respect to the masked uniform choice, given its own query--key cross-moment, with $\Lambda$ the Lagrange multiplier (\Cref{prop:causal-rule}).
\item \label{item:meta:ord:b} \emph{Next-token model.}  For any fixed attention choice matrix $\Lambda\in\mbbM_{d+1}$, the minimal modification of $\arrowN$ in relative entropy that reproduces $\mbbE[S^\rightarrow_\Lambda]$ is the Gibbs tilt 
$\mrmP^\to_{\Lambda,V^\star}\propto e^{\ip{V^\star}{S^\to_\Lambda}}\arrowN$, with value matrix $V^\star$ being the corresponding Lagrange multiplier (for an interior constraint, \Cref{thm:shifted-iproj}-\ref{item:shifted:thm:b}). 
In particular query--key attention and value matrices are untied.
\item \label{item:meta:ord:c} \emph{Transformer as last-token prediction score.} For $n\ge2$ with $\nu(n)>0$, conditionally on $N=n$ and on the first $n-1$ tokens $X_{1:n-1}=x_{1:n-1}$, 
under $\mrmP^\to_{\Lambda,V^\star}$, the last token $X_n$ has, exactly, the von Mises--Fisher law with density proportional to $\exp(\frac1{n-1}\,y^\top V^\star \Attn_\Lambda((x_{n-1},n-1);\arrowmu))$ with respect to $\uvol$.
For $\Lambda=Q^\top K$, its score on $\mbbS^d$ is 
\begin{equation*}
\nabla_{\mbbS^d}\log p(y)=\frac1{n-1}\,P^\perp_y\,V^\star \Attn_\Lambda((x_{n-1},n-1);\arrowmu)=\frac1{n-1}\,P^\perp_y\sum_{j=1}^{n}\frac{e^{\ip{Qx_{n-1}}{Kx_j}+M_{n-1,j}}}{\sum_{\ell=1}^{n}e^{\ip{Qx_{n-1}}{Kx_\ell}+M_{n-1,\ell}}}\,V^\star x_j\,,
\end{equation*}
that is $\frac1{n-1}$ times the tangential part of one causally masked attention head, with query matrix $Q$, key matrix $K$ and value matrix $V^\star$, evaluated at the last position of the context.
\end{enumerate}
\end{metathm}

This Meta-Theorem leads us to the exact transformer architecture. Indeed, the last token's log-density is the smooth function $y\mapsto\frac1{n-1}y^\top V^\star \Attn_\Lambda((x_{n-1},n-1);\arrowmu)$ on $\mbbS^d$, with gradient $\frac1{n-1}P^\perp_y\big(V^\star \Attn_\Lambda((x_{n-1},n-1);\arrowmu)\big)$. Therefore the last token's law is the invariant law of the Langevin stochastic differential equation
\begin{equation}\label{eq:SDE:one:head}
\De Y_t=\frac1{n-1}P^\perp_{Y_t}\big(V^\star \Attn_\Lambda((x_{n-1},n-1);\arrowmu)\big)\,\De t+\sqrt2\,\De W^{\mbbS^d}_t ,
\end{equation}
with $W^{\mbbS^d}$ a Brownian motion on $\mbbS^d$ with generator $\frac12\Delta_{\mbbS^d}$, which is a reversible and ergodic SDE \cite{bakry2014analysis}. 
Therefore, in order to sample from the law of the last token, we are led to consider the projected Euler--Maruyama scheme with step $\tau$,
\begin{equation}\label{eq:scheme:one:head}
Y_{k+1}=\mathrm{LN}\Big(Y_k+\frac\tau{n-1}P^\perp_{Y_k}\big(V^\star \Attn_\Lambda((x_{n-1},n-1);\arrowmu)\big)+\sqrt{2\tau}\,P^\perp_{Y_k}\xi_k\Big),
\end{equation}
with $\mathrm{LN}$ the LayerNorm normalization or the plain normalization $\mathrm{LN}(w)=w/\norm w$ onto the sphere,
and $\xi_k\overset{iid}\sim\mathcal N(0,\Id)$. This recovers almost exactly one step of a single-head transformer block: a residual update by the output of a causally masked attention head, followed by a normalization onto the sphere, with injected noise.
The identification concerns the form of one update only. The drift in \eqref{eq:scheme:one:head} is
computed once, from the context $x_{1:n-1}$, and does not depend on $Y_k$. Depth is therefore not explained by this construction (see the conclusion).

The only ingredient of a transformer block that is not present in this scheme is the position-wise multi-layer perceptron (MLP) that follows the attention step, while the only ingredient of the scheme that is not present in a deterministic transformer block is the noise $\sqrt{2\tau}\,P^\perp_{Y_k}\xi_k$. 
We argue, at a heuristic level, that the two play the same role. First, in the splitting interpretation of \cite{lu2019understanding}, a transformer layer is a Lie--Trotter splitting step of an interacting particle system: the attention sub-layer is the interaction term, 
while the position-wise MLP is the term acting on each token separately. In the scheme above the interaction is the attention drift, and the term acting on each token separately is the noise. Second, Agazzi et al.~\cite{agazzi2026stochastic} prove that in deep and wide transformers whose MLP blocks have random weights, as at initialization, 
the MLP blocks produce in the scaling limit a transport-type noise on the sphere, independent of the configuration of the tokens and common to all of them, so that the token dynamics converge to a stochastic interacting particle system on $\mbbS^d$ whose density solves an SPDE with a Laplace--Beltrami diffusion term. 
In our sampling problem only the candidate token moves, the context being fixed, so that the distinction between common and independent noise is immaterial: the MLP noise acts on the candidate as a Brownian perturbation, as in the Langevin scheme. 
Related stochastic versions of self-attention dynamics, in which the noise is added by hand \cite{rigollet2025meanfield,shalova2024solutions} or arises from random value matrices \cite{fedorov2026clustering}, display the same mechanism: noise prevents the collapse of the dynamics onto a single cluster, which the deterministic dynamics would reach. 
In the language of the scheme, without noise each step is a projected gradient ascent step of $y\mapsto y^\top V^\star\Attn_\Lambda((x_{n-1},n-1);\arrowmu)$, which converges to the mode of the von Mises--Fisher law instead of sampling from it. We stress that this identification is heuristic: in a trained transformer the MLP weights are not random, and the results of \cite{agazzi2026stochastic} concern the random-weight scaling regime.

\begin{figure}[t]
  \centering
  \includegraphics[width=0.85\textwidth]{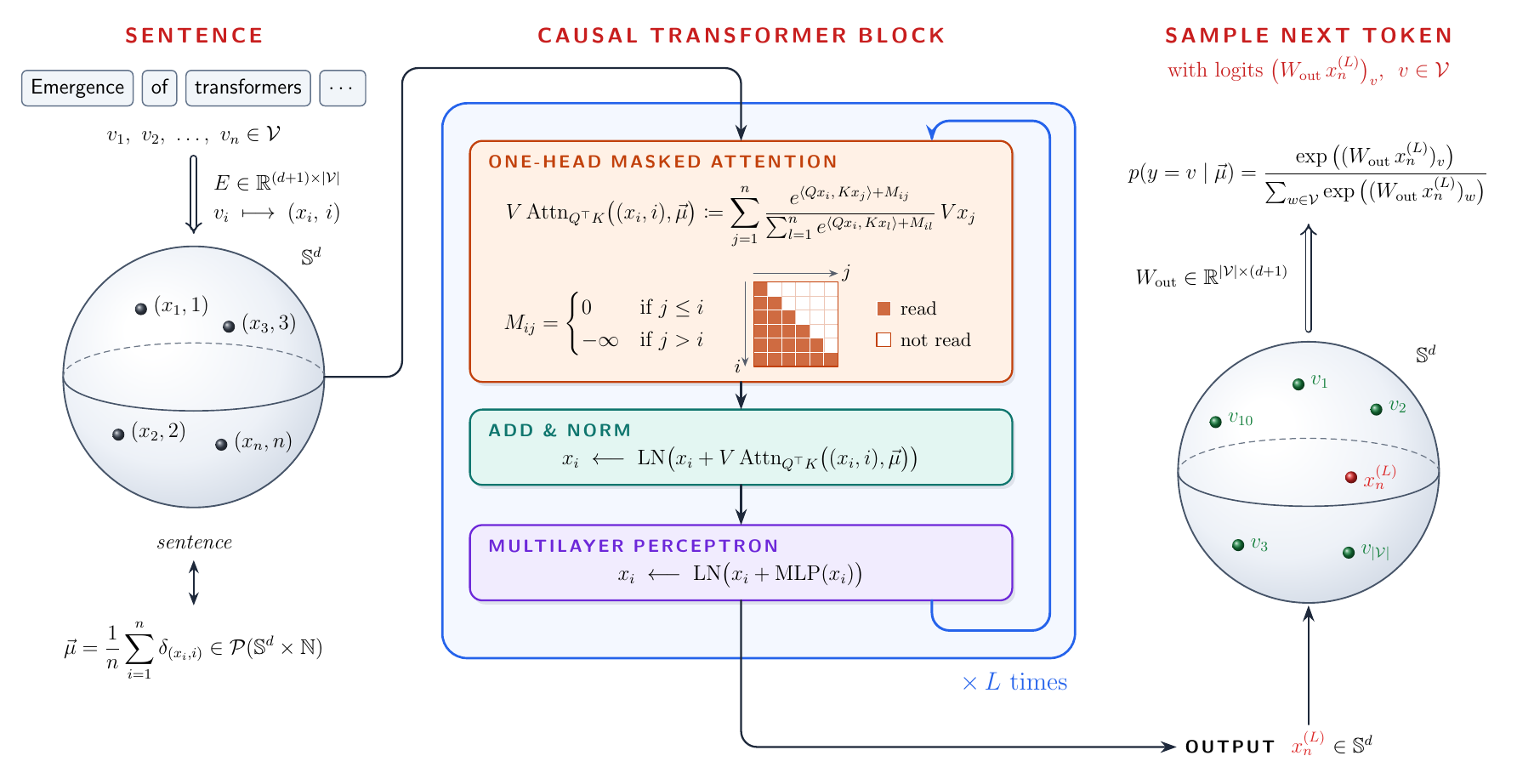}
  \caption{Causal version of the pipeline of \Cref{fig:pipeline}. Two things change. First, each token is recorded together with its position, so that the sentence becomes the ordered measure $\arrowmu=\frac1n\sum_{i}\delta_{(x_i,i)}\in\mcP(\mbbS^d\times\N)$. 
  Second, the attention sub-layer is causally masked: the additive mask $M$ sets to zero the weight of every key coming after the query ($j>i$), so that each token only reads itself and the tokens before it. }
  \label{fig:pipeline:masked}
\end{figure}

\subsection{Emergence of multi-head architectures: more choice rules in the statistics}\label{sec:multihead}
A query need not choose its key under a single rule. Suppose that it chooses under $H$ different rules, each satisfying Luce's axiom with its own choice function and its own normalization, for instance because different rules capture different relations between tokens. Truncating each choice function at its lowest relational order, as above, gives $H$ choice rules with pairwise distinct attention matrices $\Lambda_h=Q_h^\top K_h$, $h=1,\dots,H$, and the natural statistic is the vector of the corresponding cross-moments.
Below we distinguish how considering multiple choice functions in the statistical constraint affects our model.

\medskip

\noindent\textbf{Unordered model.} With $\Phi=(S_{\Lambda_1},\dots,S_{\Lambda_H})$, the minimizer of \eqref{eq:entropic-intro} is
\[
\mrmP^\star\propto\exp\Big(\sum_{h=1}^H\ip{V^\star_h}{S_{\Lambda_h}(\mu)}\Big)\,\mbfN=\exp\Big(\sum_{h=1}^H\int x^\top V^\star_h\,\Attn_{\Lambda_h}(x;\mu)\,\mu(\De x)\Big)\,\mbfN,
\]
with one value matrix $V^\star_h$ per head, each being the Lagrange multiplier of the corresponding cross-moment. Existence and uniqueness of the multipliers follow from \Cref{thm:iproj}, since the vector statistic is non-degenerate as soon as $\nu$ charges some length $n\ge2H+1$ (\Cref{prop:independence} with $A_0=0$). Since first variations are linear in the statistic, \Cref{lem:score} shows that the score of the last token is the sum over $h$ of the single-head scores of \Cref{thm:last}, whose query part is
\[
\sum_{h=1}^HV^\star_h\,\Attn_{\Lambda_h}(y;\mu_y)=\sum_{h=1}^H\sum_{j=1}^n\frac{e^{\ip{Q_hy}{K_hx_j}}}{\sum_{\ell=1}^ne^{\ip{Q_hy}{K_hx_\ell}}}\,V^\star_hx_j ,
\]
that is, multi-head attention: concatenating the $H$ head outputs amounts to summing over the heads, $V^\star_h$ being the value projection of head $h$. As for one head, each head also contributes a key-side term and a correction.

\medskip 

\begin{figure}[t]
  \centering
  \includegraphics[width=0.7\textwidth]{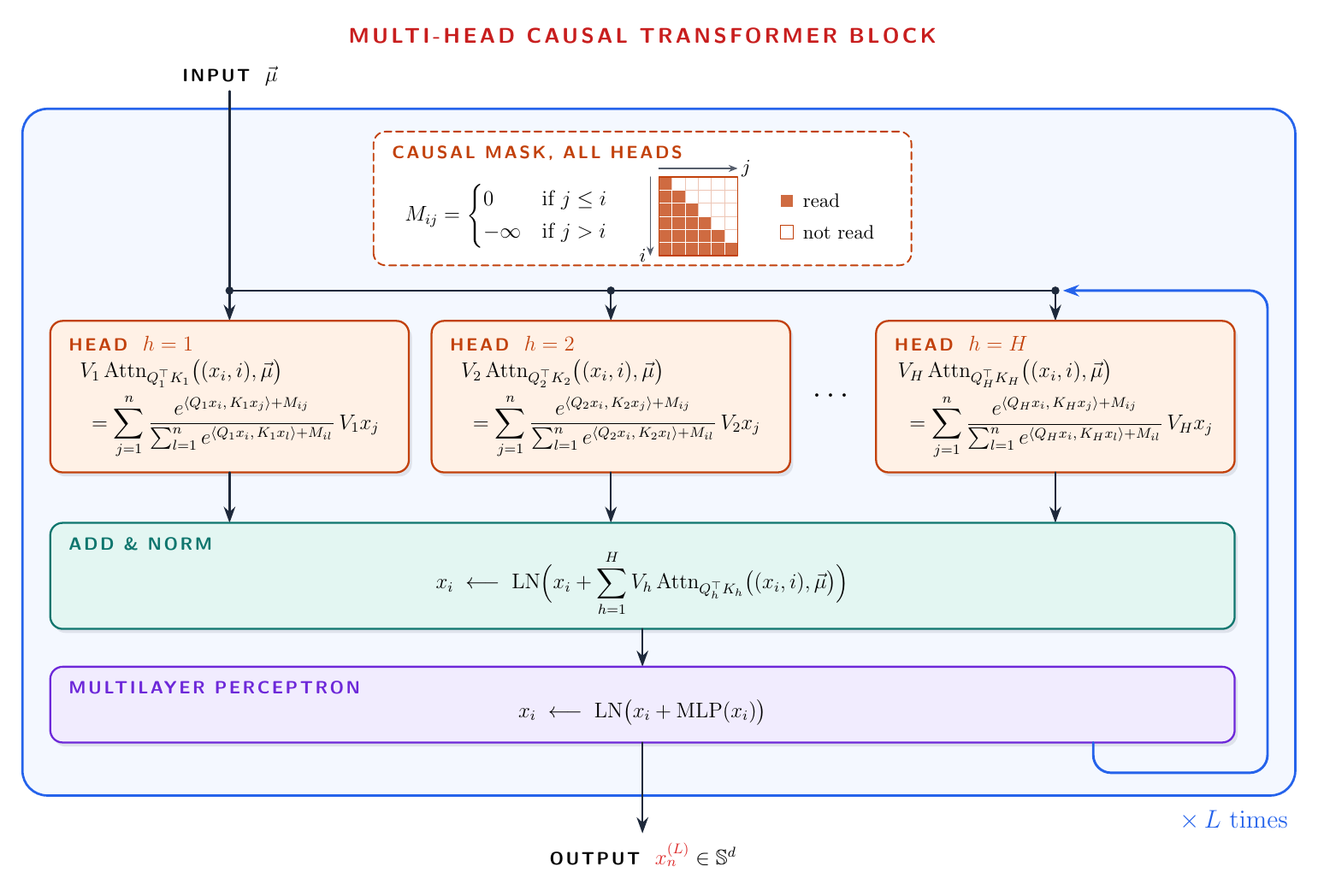}
  \caption{Multi-head version of the causal transformer block of \Cref{fig:pipeline:masked}. The single masked head is replaced by $H$ heads, each with its own query, key and value matrices $Q_h,K_h,V_h$, all sharing the same causal mask $M$. 
  The head outputs are summed before the residual connection and the normalization. In our model each head corresponds to a different choice rule, with attention matrix $\Lambda_h=Q_h^\top K_h$, and $V_h$ is the Lagrange multiplier of the corresponding cross-moment.}
  \label{fig:multihead}
\end{figure}

\noindent\textbf{Causal model.} With the masked choice rules and the shifted cross-moments, the tilt is $\exp\big(\sum_h\ip{V^\star_h}{S^\to_{\Lambda_h}}\big)\arrowN$. 
Its readout is still linear in the last token, with a coefficient that depends on the context only, and the proof of \Cref{thm:shifted-last} applies verbatim: 
given $N=n$ and $X_{1:n-1}=x_{1:n-1}$, the last token has the von Mises--Fisher law with density proportional to
\[
\exp\Big(\frac1{n-1}\,y^\top\sum_{h=1}^HV^\star_h\,\Attn_{\Lambda_h}((x_{n-1},n-1);\arrowmu)\Big),
\]
whose mean direction is the output, at the last position of the context, of a causally masked multi-head attention layer, without key-side or correction terms. 
Multi-head attention is thus the statistical counterpart of a query having several independent ways of choosing its keys and the corresponding sampling scheme reads as
\begin{equation*}
Y_{k+1}=\mathrm{LN}\Big(Y_k+\frac\tau{n-1}P^\perp_{Y_k}\big(\sum_{h=1}^H V^\star_h \Attn_{\Lambda_h}((x_{n-1},n-1);\arrowmu)\big)+\sqrt{2\tau}\,P^\perp_{Y_k}\xi_k\Big),
\end{equation*}
which corresponds to a residual update by the output of a causally masked multi-head attention architecture, followed by a normalization onto the sphere, with noise injection.

\subsection{Unembedding as a projection onto the finite vocabulary}\label{sec:unembedding}
In practice vocabularies are finite, and the transformer architecture includes an embedding of the vocabulary $\mcV=\{v_1,v_2,\dots,v_{|\mcV|}\}$ into a lower-dimensional space and a final unembedding operator, from which the logits of the law of the next token are computed. For ease of exposition, let us consider a finite vocabulary $\mcV\subset\mbbS^d$ and the identity embedding. According to our model, in order to sample the last token we are led to simulate the Langevin SDE~\eqref{eq:SDE:one:head}, for instance via the scheme~\eqref{eq:scheme:one:head}, which mimics the evolution of the last token in a transformer. Since $\mcV$ is a discrete subset of $\mbbS^d$, the output $\widehat Y\in\mbbS^d$ of the scheme is in general not an element of the vocabulary. To sample a token we therefore look for the elements of $\mcV$ that are closest to $\widehat Y$, and we measure closeness through the inner products $\ip{v}{\widehat Y}$, $v\in\mcV$: if $\ip{v}{\widehat Y}\approx1$, the next token is most probably $v$, whereas if $\ip{v}{\widehat Y}\le0$ it is unlikely to be $v$.

In light of these considerations, given $\widehat Y$, we define the law of the last token $Y^\mcV$ on the vocabulary by
\begin{equation}\label{eq:my:logits}
p^\mcV\big(Y^\mcV=v\,\big|\,\widehat Y\big)=\frac{\exp\big(\sigma(\ip{v}{\widehat Y})\big)}{\sum_{w\in\mcV}\exp\big(\sigma(\ip{w}{\widehat Y})\big)},\qquad v\in\mcV,\qquad \sigma(t)=\tan\Big(\frac\pi4(t-1)\Big),
\end{equation}
that is, the logit of $v$ is a non-decreasing, non-positive function of the inner product $\ip{v}{\widehat Y}$, which equals $0$ when $\widehat Y=v$ (exact recovery of a token) and $-\infty$ in the antipodal case $\widehat Y=-v$, so that antipodal tokens are never sampled. The choice of $\sigma$ is arbitrary: any non-decreasing function $\sigma\colon[-1,1]\to[-\infty,0]$ with $\sigma(1)=0$ would do, possibly with a temperature parameter, $\sigma_T(t)=\sigma(t)/T$.

This construction recovers the unembedding step of transformers. In practice, the output $\widehat Y$ of the last layer is multiplied by an unembedding matrix $W_{\mathrm{out}}\in\R^{|\mcV|\times(d+1)}$, and the law of the next token is the softmax of the resulting logits,
\[
p^\mcV\big(Y^\mcV=v\,\big|\,\widehat Y\big)=\frac{\exp\big((W_{\mathrm{out}}\widehat Y)_v\big)}{\sum_{w\in\mcV}\exp\big((W_{\mathrm{out}}\widehat Y)_w\big)},
\]
where $(W_{\mathrm{out}}\widehat Y)_v$ is the entry of $W_{\mathrm{out}}\widehat Y\in\R^{|\mcV|}$ indexed by $v$. In practice $W_{\mathrm{out}}$ is learned from data. Our interpretation suggests instead building $W_{\mathrm{out}}$ by stacking the tokens of the vocabulary as rows,
\[
W_{\mathrm{out}}=\left(\begin{array}{c}
v_1^\top\\ 
v_2^\top\\ 
\vdots\\ 
v_{|\mcV|}^\top
\end{array}\right)\in\R^{|\mcV|\times(d+1)},\qquad\text{so that}\qquad W_{\mathrm{out}}\widehat Y=\left(\begin{array}{c}
\ip{v_1}{\widehat Y}\\ 
\ip{v_2}{\widehat Y}\\ 
\vdots\\ 
\ip{v_{|\mcV|}}{\widehat Y}
\end{array}\right)\in\R^{|\mcV|}.
\]
With this notation, \eqref{eq:my:logits} reads
\[
p^\mcV\big(Y^\mcV=\cdot\,\big|\,\widehat Y\big)=\mathrm{softmax}\Big(\sigma\big(W_{\mathrm{out}}\widehat Y\big)\Big),
\]
with $\sigma$ applied componentwise. For the linear choice $\sigma(t)=(t-1)/T$ this is exactly the softmax of the logits $W_{\mathrm{out}}\widehat Y/T$, since the constant $-1/T$ cancels in the normalization: the standard unembedding with temperature $T$. 
The unembedding matrix is then the matrix of the vocabulary, which is fixed and does not need to be trained, and it has a clear geometric meaning: $W_{\mathrm{out}}$ projects the output of the scheme, a point of the sphere that is in general not a token, onto the finite vocabulary, 
by scoring each token by its alignment with $\widehat Y$, and the next token is sampled with higher odds for the tokens closest to $\widehat Y$. The non-linearity $\sigma$ only reshapes these scores, and it could be trained while keeping $W_{\mathrm{out}}$ fixed, 
since the latter encodes the proximity of the output to the admissible tokens. This reading of the unembedding as a projection onto the vocabulary is also the one underlying the ``logit lens'' and related interpretability methods, which decode intermediate representations of a transformer by multiplying them by the unembedding matrix \cite{dar2023analyzing}. 
An exact version of this construction holds when the reference law of tokens is itself discrete; we discuss it in \Cref{app:vocab}.

\medskip

Finally, let us explain how the embedding can be included in this discussion. To lower the dimension of the tokens, the first step of a transformer is an embedding, which maps the vocabulary, represented by the one-hot vectors $e_1,\dots,e_{|\mcV|}\in\R^{|\mcV|}$, into a lower-dimensional space through a learned matrix $E\in\R^{(d+1)\times|\mcV|}$, followed by a normalization onto $\mbbS^d$. The embedded vocabulary is then $\{v_i=Ee_i/\abs{Ee_i}:\ i\le|\mcV|\}\subset\mbbS^d$, and our construction gives the unembedding matrix whose rows are the normalized columns of $E$,
\[
W_{\mathrm{out}}=\mathrm{diag}\big(\abs{Ee_1},\dots,\abs{Ee_{|\mcV|}}\big)^{-1}E^\top .
\]
Hence the unembedding is still learned, but it is not an independent parameter: up to the normalization of the rows, it is the transpose of the embedding matrix, which is learned once. 
This is the weight tying of input and output embeddings, used in the original transformer \cite{vaswani2017attention} and introduced in \cite{press2017using,inan2017tying}, where it is motivated empirically and, in \cite{inan2017tying}, through a loss framework; in our framework it is a consequence of interpreting the unembedding as the projection of the output of the sampling scheme onto the embedded vocabulary.
Untying embedding and unembedding gives more flexibility to large models \cite{chung2021rethinking} and is associated with more isotropic representations \cite{machina2024anisotropy}, while tying biases token embeddings toward the output space \cite{lopardo2026weight}. In our setting this can be circumvented without untying: $W_{\mathrm{out}}$ only encodes the proximity of $\widehat Y$ to the tokens, while the output-specific parameters are carried by the trainable non-linearity $\sigma$, applied to $W_{\mathrm{out}}\widehat Y$.

\bigskip

\subsection{Comparison with existing contributions}\label{sec:intro:literature}
A vast body of literature has addressed why transformers work well. Most contributions aim at explaining this by studying how sentences evolve when fed into a deep transformer architecture. This line of research treats tokens as interacting particles and 
layers of a deep architecture as splitting steps of a numerical ODE solver for a convection--diffusion equation \cite{lu2019understanding}. The same approach has been lifted to the space of probability measures, where one could treat the mean-field limit of self-attention as a continuity equation on $\mcP(\mbbS^d)$ \cite{sander2022sinkformers,geshkovski2025perspective}.
This last approach has led to many fruitful discoveries such as the emergence of clustering \cite{geshkovski2023emergence,chen2025quantitative}, of long metastable phases \cite{geshkovski2024dynamic},  also under causal masking \cite{karagodin2024clustering} (see also \cite{geshkovski2026measure} for expressivity).
These works take the attention field as given and study the flow it generates, and gradient-flow structures are obtained for tied and symmetric matrices or for a reweighted metric \cite[Section~3]{geshkovski2025perspective}. In our work instead, we consider every possible untied pair $(\Lambda,V)$ and argue that the self-attention field is the query part of a Wasserstein gradient of the attention readout (\Cref{rem:particles}), that
the unnormalized field is the score of a kernel-sum statistic (\Cref{rem:usa}), and that in the causal model one masked head recovers exactly the score of the last token (\Cref{thm:shifted-last}). Similarly to \cite{rigollet2025meanfield,shalova2024solutions,fedorov2026clustering,agazzi2026stochastic}, we also consider stochastic attention dynamics, whose noise appears here in the Langevin sampler of the last token.

Finally, the biggest difference between these works and ours concerns the architecture parameters: there the attention is a fixed given input (often symmetric or the identity) and the value matrix is often tied to the query--key matrix; here both matrices are obtained as optimal parameters
of an entropic projection of a structureless null model on $\mcP(\mcP(\mbbS^d))$.

\medskip

Ours is not the first contribution establishing a connection between entropic transport problems and the softmax attention mechanism (\Cref{prop:gibbs-attn}).
For a single query, this one-sided entropic transport reading of attention is developed in \cite{litman2025scaled}, which also interprets the backward pass as an advantage-based policy gradient. There the score and the reference measure are given; here the bilinear score follows from Luce's axiom and the harmonic truncation (\Cref{prop:harmonic}), and $\Lambda$ is the Lagrange multiplier of the query--key cross-moment (\Cref{prop:condmaxent}). 
The same entropic transport problem considered in \cite{litman2025scaled}, with both marginals prescribed instead, corresponds to the
Schr\"odinger problem \cite{Schr,Schr32,LeoSch,nutz2021entropic}. This choice leads to doubly stochastic attention matrices, also known in this setting as Sinkformers \cite{sander2022sinkformers} (\Cref{rem:sinkhorn}). 
The same idea also appears in \cite{fraiman2026expressive}, where a text is a distribution of tokens and attention is a probabilistic relation between them: softmax normalization produces conditional relations, Sinkhorn normalization produces joint relations with prescribed marginals.
There the normalization is a design choice based on the Schr\"odinger problem and the Schr\"odinger/entropic plan adds to the softmax weights a key bias, the Schr\"odinger/entropic potential, which depends on the whole sentence. 
This is precisely what Luce's axiom \cite[Axiom 1]{luce1959individual} excludes here in our derivation of the attention mechanism as the lowest harmonic order relational rule. In particular, the connection with one-sided entropic transport problems here is a consequence of our derivation rather than the starting point for it.
 
\medskip

Different papers have addressed the derivation of the attention mechanism. In addition to the entropic transport approach, other regularizations can be considered which yield sparse variants (e.g., \cite{martins2016softmax}). 
Our derivation is not based on a different regularization but rather on Luce's axiom, the classical foundation of the softmax in discrete choice \cite{luce1959individual,mcfadden1974,yellott1977relationship,luce1977choice20years}.
A derivation close to ours is the one considered in modern Hopfield networks and energy transformers, where attention is identified as a descent step for an energy with values tied to keys \cite{ramsauer2021hopfield}.  This corresponds to our free-energy statistic (\Cref{rem:free-energy}), whereas our cross-moment model unties query--key and value matrices. 

A complementary statistical line of work fixes the attention architecture and a data model, and studies what it can represent or what risk minimization recovers: the relevant token in single-location regression~\cite{marion2025attention}, the centroids of a Gaussian mixture~\cite{maulensoto2025clustering}, 
and the leading principal direction of the token covariance~\cite{maulensoto2026pca}. Viewing prompts as measures, Boursier and Boyer~\cite{boursier2026softmax} show that, for i.i.d.\ centered Gaussian tokens, softmax attention converges in the infinite-prompt limit to a linear operator on the token measure; as noted in~\cite{gabet2026softmax}, 
this removes query-dependent selection, which Gaussian-mixture tokens restore. An analogous dichotomy appears in \Cref{prop:additive}, although through a different mechanism: there, moment statistics only pool the context whatever the token law, and competition between keys requires relational structure, which in the present work is derived from Luce's axiom rather than from the data distribution. 
Bohbot et al.~\cite{bohbot2026token} also model prompts as empirical measures of tokens and quantify how fast attention computed on $n$ tokens converges to its infinite-token limit (see also the concentration bounds in~\cite{boursier2026softmax}). 

\medskip
Finally, the link between language models, conditional maximum entropy and exponential families is well established, and dates back to before the word \emph{large} appeared in front of \emph{language models}. 
In \cite{berger1996maximum,rosenfeld1996maximum} conditional maximum-entropy models pair features of the history with the next word, whereas in \cite{rosenfeld2001whole} whole-sentence exponential models are introduced as tilts of a baseline language model by sentence features. 
The model we consider here has this form as well, with a structureless baseline and a relational, matrix-valued feature, and the selection of $\Lambda$ follows the minimax-entropy principle \cite{zhu1997minimax}. 
Gibbs measures and entropy also appear in statistical-mechanics analyses of transformers
(e.g.\ \cite{tiberi2024dissecting,cui2025phase,troiani2025fundamental,boncoraglio2025bayes}).
There the Gibbs measure is placed on the weights, as a posterior or a Boltzmann measure of the
training loss, for a fixed model of the data, whereas here the Gibbs measure lives on the space of sentences.

\subsection{Conclusion and future perspectives}
In this paper we have shown that the attention head of the transformer architecture is the score of
the law of the last token in a simple statistical model of language. The setting is natural in
light of two structural features of transformers, normalization and permutation equivariance:
sentences are empirical measures of tokens on the sphere, a language is a law on $\mcP(\mbbS^d)$, 
and our prior belief is that sentences carry no structure at all. The only principle we use
is the classical maximum-entropy, or minimal-information, principle: among all laws of
sentences that reproduce an observed sentence-level statistic, we select the one closest to the
structureless null model in relative entropy. Within this modeling setting, the softmax attention head, 
its untied query--key and value matrices, the causal mask and the multi-head sum are derived from our modeling choices rather than architectural inputs.
Softmax attention is the lowest-order choice rule compatible with Luce's axiom, whereas polynomial statistics, which do not let keys compete, only pool the context. Its query--key matrix is the Lagrange multiplier of a conditional maximum-entropy problem inside each sentence, while the value matrix is the Lagrange multiplier of the attention cross-moment over the language. In particular, query--key and value matrices are untied, as in practice.
In the unordered model the score of the last token contains a transformer head, together with a key-side term and a correction. 
For ordered sentences, causal masking is placed in the reference (uniform) choice rule and propagates to the attention rule through the finiteness of the relative entropy, and the next-token cross-moment removes both extra terms.
The last token then has exactly a von Mises--Fisher law, whose score is one causally masked attention head. 
A projected Euler discretization of its Langevin dynamics has the form of a residual attention update followed by normalization, with the injected noise playing, heuristically, the role of the MLP. 
Moreover, several constraining statistics give a multi-head attention model, and the projection onto a finite vocabulary gives a weight-tied unembedding.
Finally, constraining the attention free energy instead of the cross-moment gives a model of the same kind, in which values are necessarily tied to keys (\Cref{rem:free-energy}): the untied architecture of transformers is singled out by the cross-moment statistic.

\medskip

We believe our contribution opens the way to several possible research directions. In particular, we are currently investigating the use of different priors, the analysis of higher-order attention models, where we truncate the choice rule function at higher harmonic orders, and the inclusion of feature embeddings within the choice rule (cf.~\Cref{prop:condmaxent}).  
From a statistical perspective, we have characterized an optimal choice rule $\Lambda$ by the minimax-entropy principle (\Cref{rem:training}), but its statistical analysis (consistency and rates when trained over $J$ observed sentences) remains open. 
Lastly, in this work we have considered the law of the last token and we have seen how a one-layer transformer architecture emerges in its sampling scheme. However, in practice many different layers of this architecture are stacked on top of each other and all tokens undergo the same numerical steps. 
It is unclear how this fits the current setup, and we are currently investigating how this deep architecture could be explained from our entropic projection point of view. Understanding this in particular would lead us to consider evolution equations on $\mcP(\mbbS^d)$ and $\mcP(\mcP(\mbbS^d))$.

\subsection{Organization and notation}
In \Cref{sec:preliminaries} we solve the entropic problem \eqref{eq:entropic-intro} for a general statistic (\Cref{thm:iproj}), and express the law of the last token and its score through the first variation of the statistic (\Cref{lem:score}).
In \Cref{sec:why} we show that polynomial statistics only pool the context, derive the lowest-order choice rule from Luce's axiom and study the model with the attention cross-moment at a fixed attention matrix. We show existence and uniqueness of the value matrix, and compute the first variation of the readout, 
and the exact law and score of the last token. We conclude with remarks on training and on the free-energy statistic.
In \Cref{sec:shifted} we treat ordered sentences, the causal choice rule and the shifted cross-moment, and prove that the last token has a von Mises--Fisher law whose score is one causal attention head.
The appendices contain the proof of the non-degeneracy of attention cross-moments and the emergence of the unembedding matrix for discrete vocabularies.

\subsubsection*{Notation}
We write $\mbbM_{d+1}=\R^{(d+1)\times(d+1)}$ for the space of square matrices of size $d+1$, with the inner product $\ip{A}{B}=\tr(A^\top B)$.
Let $\norm{\cdot}_\op$ and $\norm{\cdot}_\Fr$ be the operator and Frobenius norms, so that $x^\top\Lambda y=\ip{\Lambda}{xy^\top}$ and $\abs{x^\top\Lambda y}\le\norm{\Lambda}_\op\le\norm{\Lambda}_\Fr$ for $x,y\in\mbbS^d$.
We assume $d\ge1$, and $P^\perp_x=\Id-xx^\top$ is the orthogonal projection onto the tangent space $T_x\mbbS^d$.
$\mcP(\mcX)$ carries the weak topology, $\mu(f)=\int f\,\De\mu$, and $\rel{\mrmP}{\mrmQ}=\int\log\frac{\De\mrmP}{\De\mrmQ}\,\De\mrmP$ if $\mrmP\ll\mrmQ$, and $+\infty$ otherwise. We write $x_{1:n}=(x_1,\dots,x_n)$ and $(x_{1:n},y)\in(\mbbS^d)^{n+1}$ for the vector obtained by appending $y$. Finally, $\mbbE_{\mbfN}$ and $\mbbE_\mrmP$ denote expectations of functions of the sentence $\mu\in\mcP(\mbbS^d)$.

\subsection*{Statement on use of AI}  The author used Claude (Anthropic; models Sonnet~5.5 and Opus~5.5) for language editing, for
rendering \Cref{fig:pipeline,fig:pipeline:masked,fig:multihead} from the author's sketches, for
bibliographic exploration, and for discussing and checking mathematical arguments. 
The author takes full responsibility for the mathematical arguments, their correctness, and the final content of the manuscript.

\section{Preliminaries}\label{sec:preliminaries}

In this section the statistic $\Phi$ is general. We solve the entropic problem \eqref{eq:entropic-intro}, compute the exact law of the last token of a sentence under the solution, and express its score through the first variation of $\Phi$.

\subsection{Sentences and the entropic projection}

Let us start this preliminary section with a few basic remarks on our setup, on the null reference prior and on entropic minimization problems. Firstly, since $\mbbS^d$ is compact, recall that $\mcP(\mbbS^d)$ is compact for the weak topology, and so is $\mcP(\mcP(\mbbS^d))$.
Notice also that, since $\uvol$ is diffuse, under $\mbfN$ the atoms of $\mu$ are almost surely pairwise distinct, so $N$ is almost surely the number of atoms of $\mu$ and is a function of $\mu$.

For $m\ge1$ and $x_{1:m}\in(\mbbS^d)^m$ let us consider the function $L_m(x_{1:m})=\frac1m\sum_{i\le m}\delta_{x_i}$, and let
\[
\mcE_m=\big\{L_m(x_{1:m}):\ x_{1:m}\in(\mbbS^d)^m\big\}\subset\mcP(\mbbS^d),\qquad \mcS_{\mbfN}=\overline{\bigcup_{m\in\supp(\nu)}\mcE_m}.
\]
Since $L_m\colon(\mbbS^d)^m\to\mcP(\mbbS^d)$ is continuous and $(\mbbS^d)^m$ is a compact space, each $\mcE_m$ is compact too. Moreover, the set $\mcS_{\mbfN}$  corresponds
to the support of the null prior $\mbfN$. Indeed for any sentence $\mu\sim\mbfN$, conditionally on having length $N=m$, we have $\mu\in\mcE_m$ (since $\uvol^{\otimes m}$ has full support on $(\mbbS^d)^m$) and the support of the mixture over $m\in\supp(\nu)$ is the closure of the union of the supports. 
Finally, if $\supp(\nu)$ is infinite, then $\mbfN$ has full support, i.e. $\mcS_{\mbfN}=\mcP(\mbbS^d)$. This directly follows from Varadarajan's theorem \cite{varadarajan1958convergence}: 
for any $\rho\in\mcP(\mbbS^d)$ it is enough to consider a collection of i.i.d.\ random tokens $X_1,X_2,\dots$ with common law $\rho$: then $L_m(X_{1:m})$ weakly converges almost surely toward $\rho$. In particular it means that there exists a realization $(x_i)_{i\ge1}\subset\mbbS^d$ of these random variables such that
$L_m(x_{1:m})$ weakly converges toward $\rho$. Since $\supp(\nu)$ is infinite it contains a diverging sequence $m_k\uparrow \infty$ along which we then have $L_{m_k}(x_{1:m_k})\in\mcE_{m_k}$ weakly converging to $\rho$, which implies $\rho\in \overline{\bigcup_{m\in\supp(\nu)}\mcE_m}=\mcS_{\mbfN}$.

\bigskip

Next, we recall a standard result for constrained $\KL$-minimization problems, such as ours~\eqref{eq:entropic-intro}. 
In view of that, let $\Phi\colon\mcP(\mbbS^d)\to\R^k$ be a bounded and weakly continuous map and let us consider 
\[
\Gamma_\Phi(\eta)=\log\mbbE_{\mbfN}\big[e^{\ip{\eta}{\Phi}}\big],\qquad \mrmP_\eta=e^{\ip{\eta}{\Phi}-\Gamma_\Phi(\eta)}\,\mbfN,\qquad \mcK_\Phi=\overline{\mathrm{conv}}\,\big(\Phi\big(\mcS_{\mbfN}\big)\big)\subset\R^k ,
\]
where $\overline{\mathrm{conv}}\,\big(\Phi\big(\mcS_{\mbfN}\big)\big)$ denotes the closure of the convex hull of $\Phi\big(\mcS_{\mbfN}\big)$.

\begin{theorem}[Entropic projection theorem]\label{thm:iproj}
Let $\Phi\colon\mcP(\mbbS^d)\to\mbbR^k$ be bounded, weakly continuous and assume that $\ip{u}{\Phi}$ is not constant on $\mcS_{\mbfN}$ for every $u\in\R^k\setminus\{0\}$. 
Then, the log-partition function $\Gamma_\Phi$ is real-analytic and strictly convex on $\R^k$, and $\nabla\Gamma_\Phi$ is a diffeomorphism of $\R^k$ onto the interior of $\mcK_\Phi$. 
For $c\in\mathrm{int}(\mcK_\Phi)$ the problem
\begin{equation}\label{eq:problem:entropic}
\min\big\{\rel{\mrmP}{\mbfN}\;:\;\mrmP\in\mcP(\mcP(\mbbS^d)),\ \mbbE_\mrmP[\Phi]=c\big\}
\end{equation}
has the unique solution $\mrmP_{\eta_c}\propto\exp(\langle \eta_c,\,\Phi\rangle)\,\mbfN$, where $\eta_c$ is the Lagrange multiplier such that $\nabla\Gamma_\Phi(\eta_c)=c$. Every admissible $\mrmP$ satisfies the Pythagorean identity
\begin{equation}\label{eq:pythagoras}
\rel{\mrmP}{\mbfN}=\rel{\mrmP}{\mrmP_{\eta_c}}+I_\Phi(c),\qquad I_\Phi(c):=\rel{\mrmP_{\eta_c}}{\mbfN}=\sup_{\eta\in\R^k}\big\{\ip{\eta}{c}-\Gamma_\Phi(\eta)\big\}.
\end{equation}
For $c\notin\mcK_\Phi$ no $\mrmP$ with $\rel{\mrmP}{\mbfN}<\infty$ is admissible.
\end{theorem}

The characterization of minimizers for $\KL$ divergence as exponential tilting of the reference measure is a well-known fact in optimization analysis under 
linear constraints. We refer the reader to \cite{follmer:stoch:finance,brown1986,barndorff-nielsen1978,csiszar1975,csiszar2005closures}. 
For the sake of completeness we provide below a proof of this result in our setting, based on \cite{brown1986}.

\begin{proof} We divide the proof into steps.

\medskip

\noindent\emph{Reduction to an exponential family on $\R^k$.} All the quantities in the statement depend on $\mu\in\mcP(\mbbS^d)$ only through $\Phi(\mu)$.
Let $\nu_\Phi\coloneqq\Phi_\#\mbfN\in\mcP(\R^k)$ and $M\coloneqq\sup\abs{\Phi}<\infty$, so that $\nu_\Phi$ is concentrated on the ball $\{\abs t\le M\}$ and
\[
\Gamma_\Phi(\eta)=\log \mcZ(\eta),\qquad \mcZ(\eta)\coloneqq\int_{\R^k}e^{\ip{\eta}{t}}\,\nu_\Phi(\De t).
\]
Thus $\{e^{\ip{\eta}{t}-\Gamma_\Phi(\eta)}\nu_\Phi\}_{\eta\in\R^k}$ is a standard exponential family on $\R^k$ in the sense of \cite[Definition 1.1]{brown1986}, with base measure $\nu_\Phi$ and sufficient statistic $t$, and $\mrmP_\eta$ is its lift to sentences: $\Phi_\#\mrmP_\eta=e^{\ip{\eta}{t}-\Gamma_\Phi(\eta)}\nu_\Phi$.

\medskip

\noindent\emph{Analyticity.} For $\eta\in\mathbb C^k$ we have $\abs{e^{\ip{\eta}{t}}}\le e^{M\abs{\mathrm{Re}\,\eta}}$ on the support of $\nu_\Phi$, 
so the integral defining $\mcZ$ converges for every $\eta\in\mathbb C^k$ and, by dominated convergence, may be differentiated under the integral sign in each complex variable:
 $\mcZ$ is an entire function, the Laplace transform of a compactly supported measure (cf. \cite[Theorem 7.23]{rudin1991FA}). On $\R^k$ it is real and $\mcZ(\eta)\ge e^{-M\abs\eta}>0$, 
 so $\Gamma_\Phi=\log \mcZ$ is real-analytic on $\R^k$ too, with $\abs{\Gamma_\Phi(\eta)}\le M\abs\eta$. Differentiating under the integral sign gives
\[
\nabla\Gamma_\Phi(\eta)=\mbbE_{\mrmP_\eta}[\Phi],\qquad\nabla^2\Gamma_\Phi(\eta)=\Cov_{\mrmP_\eta}(\Phi)\succeq0 .
\]

\medskip

\noindent\emph{Minimality and strict convexity.} We claim $\Cov_{\mrmP_\eta}(\Phi)\succ0$. 
Since $\mrmP_\eta$ and $\mbfN$ are equivalent (the density $e^{\ip{\eta}{\Phi}-\Gamma_\Phi(\eta)}$ is bounded away from $0$ and $\infty$), $\mrmP_\eta$ has support $\mcS_{\mbfN}$. 
If $u^\top\Cov_{\mrmP_\eta}(\Phi)u=0$ for some $u\ne0$, the continuous function $\ip{u}{\Phi}$ is $\mrmP_\eta$-almost surely constant, 
hence constant on $\mcS_{\mbfN}$, which is excluded by assumption. This is equivalent to saying that the exponential family $\{e^{\ip{\eta}{t}-\Gamma_\Phi(\eta)}\nu_\Phi\}_{\eta\in\R^k}$ is minimal.
Consequently, $\Gamma_\Phi$ is strictly convex.

\medskip

\noindent\emph{Diffeomorphism.} The parameter space $\{\eta:\ \mcZ(\eta)<\infty\}$ is the whole of $\R^k$, which is open: the family is regular, hence steep \cite[Definition 3.2]{brown1986}. 
For a minimal and steep standard exponential family, the mean value map $\eta\mapsto\nabla\Gamma_\Phi(\eta)$ is a bijection of the interior of the natural parameter space onto the interior of the 
convex support of the base measure \cite[Theorem~3.6]{brown1986}. 
Here the convex support is $\overline{\mathrm{conv}}(\supp(\nu_\Phi))=\mcK_\Phi$: indeed $\Phi$ is continuous and $\mcS_{\mbfN}=\supp(\mbfN)$ is compact, 
so that $\Phi(\mcS_{\mbfN})$ is compact as well and $\supp(\nu_\Phi)=\Phi(\mcS_{\mbfN})$.
Hence $\nabla\Gamma_\Phi:\R^k\to\inter(\mcK_\Phi)$ is a bijection. It is real-analytic and its Jacobian $\nabla^2\Gamma_\Phi$ is invertible everywhere, so by the inverse function theorem its inverse is smooth: $\nabla\Gamma_\Phi$ is a diffeomorphism.

\medskip

\noindent\emph{Existence and uniqueness.}  Since $\nabla\Gamma_\Phi:\R^k\to\inter(\mcK_\Phi)$ is a bijection, for any $c\in\inter(\mcK_\Phi)$ there exists a unique $\eta_c\in\mbbR^k$ such that $\nabla\Gamma_\Phi(\eta_c)=c$. As a consequence of this, since $\nabla\Gamma_\Phi(\eta)=\mbbE_{\mrmP_\eta}[\Phi]$, we have shown that
$\mrmP_{\eta_c}$ is a probability measure satisfying the constraint $\mbbE_{\mrmP_{\eta_c}}[\Phi]=c$. Moreover, by definition we have
\[\rel{\mrmP_{\eta_c}}{\mbfN}=\mbbE_{\mrmP_{\eta_c}}[\langle \eta_c,\Phi\rangle-\Gamma_\Phi(\eta_c)]=\langle \eta_c,c\rangle-\Gamma_\Phi(\eta_c)<+\infty\,,\]
which is finite. Therefore, the set of admissible solutions to~\eqref{eq:problem:entropic}
\[\mathcal{A}_c\coloneqq \left\{\mrmP\in\mcP(\mcP(\mbbS^d))\,\colon \, \mbbE_{\mrmP}[\Phi]=c\right\}\]
contains $\mrmP_{\eta_c}$, which has finite relative entropy. The set $\mcA_c$ is convex, by the linearity of the map $\mrmP\mapsto\mbbE_{\mrmP}[\Phi]$, and weakly closed, since this map is weakly continuous, $\Phi$ being bounded and continuous. 
Since the sublevel sets of $\rel{\cdot}{\mbfN}$ are weakly compact \cite[Lemma~1.4.3]{DupuisEllis}, the set $\mcA_c\cap\{\mrmP:\ \rel{\mrmP}{\mbfN}\leq\rel{\mrmP_{\eta_c}}{\mbfN}\}$ is non-empty and weakly compact. 
This is enough for arguing that there exists a solution to~\eqref{eq:problem:entropic}: the relative entropy functional $\mrmP\mapsto\rel{\mrmP}{\mbfN}$ is weakly lower semicontinuous, hence it attains its minimum over this compact set, which is its minimum over $\mcA_c$, and this minimum is finite. 
Moreover, this minimum is unique because $\rel{\cdot}{\mbfN}$ is strictly convex on its
domain $\{\mrmP : \rel{\mrmP}{\mbfN}<\infty\}$ (cf.\ \cite[Lemma~1.4.3]{DupuisEllis}),
and $\mcA_c$ is convex (a strictly convex functional has at most one
minimizer over a convex set, giving uniqueness).

\medskip

\noindent\emph{Characterization of the minimizer.} We aim at showing that $\mrmP_{\eta_c}$ is the unique minimizer. Indeed, for any admissible $\mrmP\in\mcA_c$
we have
\begin{equation*}
  \begin{aligned}
    \rel{\mrmP}{\mbfN}-\rel{\mrmP_{\eta_c}}{\mbfN}=&\,
    \rel{\mrmP}{\mbfN}-\langle\eta_c,\,c\rangle+\log\mbbE_{\mbfN}\big[e^{\langle\eta_c,\,\Phi\rangle}\big]\\
        =&\,\rel{\mrmP}{\mbfN}-\mbbE_\mrmP[\langle\eta_c,\,\Phi\rangle]+\log\mbbE_{\mbfN}\big[e^{\langle\eta_c,\,\Phi\rangle}\big]\\
        =&\,\rel{\mrmP}{\mrmP_{\eta_c}}\geq 0\,,
  \end{aligned}
\end{equation*}
with equality if and only if $\mrmP=\mrmP_{\eta_c}$. This shows that the minimizer over $\mcA_c$ is necessarily $\mrmP_{\eta_c}$, whereas the validity of \eqref{eq:pythagoras} 
follows from the above computation combined with the Donsker--Varadhan variational formula \cite[Lemma 1.4.3]{DupuisEllis}
\begin{equation*}
  \begin{aligned}
    \rel{\mrmP_{\eta_c}}{\mbfN}=\sup_{\phi\in\mcC(\mcP(\mbbS^d))}\big\{\mbbE_{\mrmP_{\eta_c}}[\phi]-\log\mbbE_\mbfN[e^\phi]\big\}\geq \sup_{\eta\in\R^k}\big\{\mbbE_{\mrmP_{\eta_c}}[\ip{\eta}{\Phi}]-\Gamma_\Phi(\eta)\big\}\\
    =\sup_{\eta\in\R^k}\big\{\ip{\eta}{c}-\Gamma_\Phi(\eta)\big\}\geq \langle \eta_c,c\rangle-\Gamma_\Phi(\eta_c)=\rel{\mrmP_{\eta_c}}{\mbfN}\,.
  \end{aligned}
\end{equation*}

\medskip

\noindent\emph{Constraints outside $\mcK_\Phi$.} If $\rel{\mrmP}{\mbfN}<\infty$, then $\mrmP\ll\mbfN$, so $\mu\in\mcS_{\mbfN}$ and $\Phi(\mu)\in\Phi(\mcS_{\mbfN})$ 
for $\mrmP$-almost every $\mu$. Hence $\mbbE_\mrmP[\Phi]$ lies in the closed convex set $\mcK_\Phi$, and no such $\mrmP$ is admissible when $c\notin\mcK_\Phi$.

\end{proof}

\subsection{The law of the last token for a general statistic $\Phi$}\label{sec:last-general}
We conclude this preliminary section by computing the law of the last token under the unique minimizer $\mrmP^\Phi$ of \eqref{eq:problem:entropic}, when considering a general statistic $\Phi$ and observed value $c\in\inter(\mcK_\Phi)$. 
According to \Cref{thm:iproj}, we can write $\mrmP^\Phi\propto\exp(\langle V^\Phi,\,\Phi\rangle)\,\mbfN$ (for consistency with the Introduction, let $V^\Phi=\eta_c$ be the Lagrange multiplier appearing in \Cref{thm:iproj}).
Clearly, $\mrmP^\Phi$ is equivalent to $\mbfN$; hence, under $\mrmP^\Phi$, the sentence $\mu$ almost surely has $N$ pairwise distinct atoms. As in the sampling procedure of the Introduction, the tokens of a sentence are its atoms 
read in a uniformly random order: given $\mu$, $(X_1,\dots,X_N)$ is a uniformly random ordering of the atoms of $\mu$.

Before resuming our discussion, let us briefly comment on the conditional predictive law of the last token $X_n$ given $X_{1:n-1}$, which exists by disintegration 
as a probability kernel in $x_{1:n-1}$ \cite[Theorem 8.37]{klenke}. Hereafter, given $N=n$, the conditional law of $X_n$ given $X_{1:n-1}$ is understood as the regular conditional distribution version given by \eqref{eq:last-token}, which is defined for every context $x_{1:n-1}$. In particular, when $\Phi$ is weakly continuous, this is the unique continuous version.

\begin{proposition}[The tokens and the last token]\label{prop:last-token}
Let $\mcZ=\mbbE_{\mbfN}[e^{\langle V^\Phi,\Phi\rangle}]$.
\begin{enumerate}[(a)]
\item \label{item:last:gen:a} $\mrmP^\Phi\big(N=m,\ X_{1:m}\in\De x_{1:m}\big)=\mcZ^{-1}\,\nu(m)\,e^{\langle V^\Phi,\Phi(L_m(x_{1:m}))\rangle}\,\uvol^{\otimes m}(\De x_{1:m})$, and $\mu=L_N(X_{1:N})$.
\item \label{item:last:gen:b} Let $n\ge1$ with $\nu(n)>0$. Given $N=n$ and $X_{1:n-1}=x_{1:n-1}$, the last token $X_n$ has density with respect to $\uvol$ 
\begin{equation}\label{eq:last-token}
p(y)=\frac{e^{\langle V^\Phi,\Phi(\mu_y)\rangle}}{\int e^{\langle V^\Phi,\Phi(\mu_{y'})\rangle}\,\uvol(\De y')},\qquad \mu_y=\frac1n\Big(\sum_{i<n}\delta_{x_i}+\delta_y\Big).
\end{equation}
\end{enumerate}
\end{proposition}

\begin{proof} Claim (a) directly follows from the definition of $\mbfN$ and $\mrmP^\Phi$.
Claim (b) follows from an application of Bayes' formula to (a) with $m=n$: the conditional density of $X_n$ given $X_{1:n-1}=x_{1:n-1}$ 
and $N=n$ is proportional to $y\mapsto e^{\langle V^\Phi,\Phi(L_n(x_{1:n-1},y))\rangle}$, and $L_n(x_{1:n-1},y)=\mu_y$.
\end{proof}

Next, we are interested in computing the score associated to the predictive law \eqref{eq:last-token}. Indeed, having access to this score allows one to generate last-token samples via a Langevin sampler. More precisely, if $\log p(y)\in\mcC^1(\mbbS^d)$ with Lipschitz gradient (as will be the case in our model, see \Cref{prop:first-variation}) then we can use (a discretization of) the
Langevin dynamics $\De Y_t=\nabla_{\mbbS^d}\log p(Y_t)\,\De t+\sqrt2\,\De W^{\mbbS^d}_t$ on $\mbbS^d$, with $W$ a Brownian motion on $\mbbS^d$ with generator $\frac12\Delta_{\mbbS^d}$, for sampling from $p\,\uvol$. Indeed it is a reversible and ergodic SDE, with invariant law $p\,\uvol$ \cite{bakry2014analysis}, i.e. its law converges to the law of the last token.

\begin{lemma}[Score of the last token]\label{lem:score}
Let $\Phi=(\varphi_1,\dots,\varphi_k):\mcP(\mbbS^d)\to\R^k$ and $V\in\R^k$, and assume that each component $\varphi_\ell$ is weakly continuous and satisfies the following two conditions:
\begin{enumerate}[(i)]
\item (Gâteaux differentiability) \label{ass:lemma:item} There is a bounded measurable function $(\mu,z)\mapsto\delta\varphi_\ell(\mu)(z)$ on $\mcP(\mbbS^d)\times\mbbS^d$ such that, for all $\mu,\varrho\in\mcP(\mbbS^d)$, the map $t\mapsto\varphi_\ell(\mu+t(\varrho-\mu))$ is differentiable on $(0,1)$ with
\[
\frac{\De}{\De t}\varphi_\ell\big(\mu+t(\varrho-\mu)\big)=\int\delta\varphi_\ell\big(\mu+t(\varrho-\mu)\big)\,\De(\varrho-\mu).
\]
The function $\delta\varphi_\ell(\mu)$ denotes the first variation of $\varphi_\ell$ at $\mu$.
\item (Continuous gradient of the first variation) \label{ass:lemma:item:ii} For every $\mu$, the function $\delta\varphi_\ell(\mu)$ is the restriction to $\mbbS^d$ of a $\mcC^1$ function defined on a neighborhood of $\mbbS^d$ in $\R^{d+1}$, 
whose gradient $(\mu,z)\mapsto\nabla_z\delta\varphi_\ell(\mu)(z)$ is jointly continuous on $\mcP(\mbbS^d)\times\mbbS^d$, where $\mcP(\mbbS^d)$ carries the weak topology.
\end{enumerate}
Write $\delta\Phi=(\delta\varphi_1,\dots,\delta\varphi_k)$. Let $n\ge1$, $x_{1:n-1}\in(\mbbS^d)^{n-1}$ and $\mu_y=\frac1n\big(\sum_{i<n}\delta_{x_i}+\delta_y\big)$ for $y\in\mbbS^d$.
Then $y\mapsto\ip{V}{\Phi(\mu_y)}$ belongs to $\mcC^1(\mbbS^d)$, and its Riemannian gradient is
\begin{equation}\label{eq:score-general}
\nabla_{\mbbS^d}\,\ip{V}{\Phi(\mu_y)}=\frac1n\,P^\perp_y\,\nabla_z\ip{V}{\delta\Phi(\mu_y)(z)}\Big|_{z=y},\qquad P^\perp_y=\Id-yy^\top .
\end{equation}
 In particular, if $\nu(n)>0$, the score of the last-token density $p$ of \Cref{prop:last-token}-\ref{item:last:gen:b} is
\[\nabla_{\mbbS^d}\log p(y)=\frac1n\,P^\perp_y\,\nabla_z\ip{V^\Phi}{\delta\Phi(\mu_y)(z)}\Big|_{z=y}.\] 
\end{lemma}

Here $\nabla_z$ is the gradient in the variable $z$ with the measure $\mu_y$ frozen; only its tangential part $P^\perp_y\nabla_z$ is intrinsic, as it does not depend on the $\mcC^1$ extension chosen in (ii). 
In the rest of the paper whenever we say that $\Phi\colon\mcP(\mbbS^d)\to \mbbR^k$ is a $\mcC^1$ function, we tacitly mean that it satisfies the assumptions of \Cref{lem:score}.

\begin{proof}
  Firstly, let $d_{\mathrm{BL}}(\mu,\mu')=\sup\big\{\abs{\int f\,\De(\mu-\mu')}:\ \norm f_\infty\le1,\ \mathrm{Lip}(f)\le1\big\}$ be the bounded-Lipschitz distance, which metrizes the weak topology on $\mcP(\mbbS^d)$.
 By linearity it suffices to prove the claim for a single component $\varphi=\varphi_\ell$, $\ell\in\{1,\dots,k\}$. 
Let $y\in\mbbS^d$, $v\in T_y\mbbS^d$, and let $\gamma:(-1,1)\to\mbbS^d$ be a $\mcC^1$ curve with $\gamma(0)=y$ and $\gamma'(0)=v$. For $s\ne0$, consider the signed measure $\mu_{\gamma(s)}-\mu_y=\frac1n(\delta_{\gamma(s)}-\delta_y)$, 
and the segment $\mu^s_t=\mu_y+t(\mu_{\gamma(s)}-\mu_y)\in\mcP(\mbbS^d)$, for $t\in[0,1]$.

Now, consider $g(t)=\varphi(\mu^s_t)$. Since $t\mapsto\mu^s_t$ is weakly continuous on $[0,1]$ and $\varphi$ is weakly continuous, $g$ is continuous on $[0,1]$, and by \ref{ass:lemma:item} it is differentiable on $(0,1)$. By the mean value theorem there is $t_s\in(0,1)$ with $g(1)-g(0)=g'(t_s)$. Hence, by \ref{ass:lemma:item} and $\mu_{\gamma(s)}-\mu_y=\frac1n(\delta_{\gamma(s)}-\delta_y)$,
\begin{equation*}
  \begin{aligned}
\frac{\varphi(\mu_{\gamma(s)})-\varphi(\mu_y)}{s}&=\frac1s\int\delta\varphi(\mu^s_{t_s})\,\De(\mu_{\gamma(s)}-\mu_y)
=\frac1n\,\frac1s\Big(\delta\varphi(\mu^s_{t_s})\big(\gamma(s)\big)-\delta\varphi(\mu^s_{t_s})(y)\Big)\\
&=\frac1n\,\frac1s\int_0^s\ip{\nabla_z\delta\varphi(\mu^s_{t_s})\big(\gamma(r)\big)}{\gamma'(r)}\,\De r\,,
  \end{aligned}
\end{equation*}
where the last step follows from the chain rule along $\gamma$ ($\delta\varphi(\mu^s_{t_s})$ is $\mcC^1$ near $\mbbS^d$). We claim that 
\begin{equation}\label{eq:limit:lemma:variation}
\lim_{s\to0}\frac{\varphi(\mu_{\gamma(s)})-\varphi(\mu_y)}s=\frac1n\ip{\nabla_z\delta\varphi(\mu_y)(y)}{v}=\ip{\tfrac1nP^\perp_y\nabla_z\delta\varphi(\mu_y)(y)}{v},
\end{equation}
where the last equality holds because $v\in T_y\mbbS^d$, so that $\ip wv=\ip{P^\perp_yw}v$ for every $w\in\R^{d+1}$. In order to show the validity of the limit in \eqref{eq:limit:lemma:variation}, it is enough to argue that uniformly in $t\in[0,1]$
\[
\sup_{t\in[0,1]}d_{\mathrm{BL}}(\mu^s_t,\mu_y)\le\tfrac1n\abs{\gamma(s)-y}\xrightarrow[s\to0]{}0\,,
\]
and combine it with the fact that $(\mu,z)\mapsto\nabla_z\delta\varphi(\mu)(z)$ is continuous on the compact metric space $\mcP(\mbbS^d)\times\mbbS^d$, hence bounded, and uniformly continuous, with a modulus of continuity $\varpi$. Indeed, for $t\in[0,1]$ and $\abs r\le\abs s$ it then follows that
\begin{equation*}
  \begin{aligned}
    \Big|\ip{\nabla_z\delta\varphi(\mu^s_{t_s})\big(\gamma(r)\big)}{\gamma'(r)}-\ip{\nabla_z\delta\varphi(\mu_y)(y)}{v}\Big|\le\varpi\Big(\tfrac1n\abs{\gamma(s)-y}+\sup_{\abs r\le\abs s}\abs{\gamma(r)-y}\Big)\sup_{\abs r\le\abs s}\abs{\gamma'(r)} \\
    +\|\nabla_z\delta\varphi(\mu)(z)\|_{\mrmL^\infty(\mcP(\mbbS^d)\times\mbbS^d)}\sup_{\abs r\le\abs s}\abs{\gamma'(r)-v},
  \end{aligned}
\end{equation*}
and the right-hand side tends to $0$ as $s\to0$, by the continuity of $\gamma$ and $\gamma'$, which implies \eqref{eq:limit:lemma:variation}. This shows the validity of the directional version of \eqref{eq:score-general} (along a direction $v\in T_y\mbbS^d$) for one fixed $\varphi=\varphi_\ell$.

\medskip

Set $g_\ell(y)=\frac1nP^\perp_y\nabla_z\delta\varphi_\ell(\mu_y)(y)$ and $G(y)=\sum_\ell V_\ell\,g_\ell(y)=\frac1nP^\perp_y\nabla_z\ip{V}{\delta\Phi(\mu_y)(z)}\big|_{z=y}$. 
By linearity, the derivative of $y\mapsto\ip{V}{\Phi(\mu_y)}$ along every $\mcC^1$ curve through $y$ with velocity $v$ equals $\ip{G(y)}{v}$. Then, since $y\mapsto(\mu_y,y)$ is continuous and $\nabla_z\delta\varphi_\ell$ jointly continuous, we conclude that
$y\mapsto\ip{V}{\Phi(\mu_y)}\in\mcC^1(\mbbS^d)$, and its Riemannian gradient is $G$.
Finally, since $\log p(y)=\ip{V^\Phi}{\Phi(\mu_y)}+\mathrm{const}$ by \eqref{eq:last-token}, the formula for the score of $p$ follows by taking $V=V^\Phi$.
\end{proof}

\section{The choice of the statistics: emergence of attention choice rules}\label{sec:why}

The aim of this section is to prove the claims of \Cref{sec:intro:choice}, by relying on \Cref{lem:score} where we have shown that the statistic $\Phi$ acts on the last 
token through the gradient of its first variation at the completed sentence. 
We first show  that polynomial (in particular pair-additive) statistics only let a new token feel the sentence through fixed, unnormalized weights, and through its mean in the case of the covariance.
This leads us to discard such statistics since linguistic structure should be relational, that is, the role of a token should be determined by which other tokens of the sentence it relates to the most.
We then build in \Cref{sec:choice} a statistic that takes into account this linguistic structure by introducing the concept of \emph{choice rule} for the sentence $\mu$ as a choice mechanism
between keys and queries within the same sentence $\mu$. Finally, we derive the attention choice rule from Luce's choice axiom and a harmonic truncation, characterize it by entropy (\Cref{sec:entropy-rule}), and define the statistic $\Phi=S_\Lambda$ as its attention cross-moment. 
For presentation's purposes we postpone the non-degeneracy of $S_\Lambda$, needed in the $\KL$ minimization (cf. \Cref{thm:iproj}), to \Cref{app:non-deg}.
Then, in \Cref{sec:cross:read:proj} we solve the entropic problem  for a fixed attention matrix $\Lambda$ \eqref{eq:entropic-intro} for $\Phi=S_\Lambda$, compute the first variation of the attention readout and the exact law and score of the last token. 
We conclude with a brief comment on training of the attention and value matrices $\Lambda$ and $V$, and on the possible use of the free energy as a different statistic (\Cref{sec:training}).

\bigskip

\begin{proposition}[Polynomial and moment statistics in the measure]\label{prop:additive}
The following statements hold true.

\begin{enumerate}[(a)]
\item \label{item:a} Fix a symmetric function $f\in \mcC((\mbbS^d)^r)$, with $r\ge1$. The corresponding polynomial statistic $\Phi(\mu):=\int f\,\De\mu^{\otimes r}$ has first variation $\delta\Phi(\mu)(z)=r\int f(z,x_2,\dots,x_r)\,\mu^{\otimes(r-1)}(\De x_{2:r})$. If $f$ is $\mcC^1$ on a neighborhood of $(\mbbS^d)^r$ in $(\R^{d+1})^r$, then $\Phi$ is $\mcC^1$ (in the sense of \Cref{lem:score}). 
The score of the associated last-token law is an average, over the $(r-1)$-tuples of tokens of the completed sentence, of $r\nabla_zf(z,x_2,\dots,x_r)|_{z=y}$. Each tuple enters with the same weight $n^{-(r-1)}$, which depends only on the length of the sentence, 
and its contribution does not depend on the other tuples. There is no query-dependent normalization, as in the transformer architecture \eqref{eq:intro:transformer}, and hence no competition between keys.
\item \label{item:b} For $r=2$ and $h\in \mcC(\mbbS^d\times\mbbS^d)$, not necessarily symmetric, $\Phi_h(\mu)=\iint h\,\De\mu\,\De\mu$ has first variation $\delta\Phi_h(\mu)(z)=\int\big(h(z,x)+h(x,z)\big)\mu(\De x)$.
\item \label{item:c} For $h(x,y)=x^\top Ay$ (moment statistics), $\Phi_A(\mu)=m_\mu^\top Am_\mu$ with $m_\mu=\int x\,\mu(\De x)$, and $\delta\Phi_A(\mu)(z)=z^\top(A+A^\top)m_\mu$. The force $\nabla_z\delta\Phi_A(\mu)(z)=(A+A^\top)m_\mu$ is the same for every candidate, and the context enters only through its mean: the query attends to the empirical mean of the sentence, not to single keys.
\item \label{item:d} (Kernel sums) Let $\Lambda\in\Mat$ and $h(x,y)=k_\Lambda(x,y)=e^{x^\top\Lambda y}$, so that
\[
\Phi_{k_\Lambda}(\mu)=\iint e^{x^\top\Lambda y}\,\mu(\De x)\,\mu(\De y),\qquad
\delta\Phi_{k_\Lambda}(\mu)(z)=\int\big(e^{z^\top\Lambda x}+e^{x^\top\Lambda z}\big)\,\mu(\De x),
\]
and
\[
\nabla_z\delta\Phi_{k_\Lambda}(\mu)(z)=\int\big(e^{z^\top\Lambda x}\,\Lambda x+e^{x^\top\Lambda z}\,\Lambda^\top x\big)\,\mu(\De x).
\]
At the completed sentence $\mu_y=\frac1n\sum_{j=1}^n\delta_{x_j}$, with $x_n=y$, the score of the last token under the tilt $e^{\Phi_{k_\Lambda}}\mbfN$ (Lagrange multiplier equal to $1$) is therefore
\[
\nabla_{\mbbS^d}\log p(y)=\frac1{n^2}\,P^\perp_y\sum_{j=1}^n\Big(e^{y^\top\Lambda x_j}\,\Lambda x_j+e^{x_j^\top\Lambda y}\,\Lambda^\top x_j\Big).
\]
Every key $x_j$ acts with a weight that does not depend on the other keys, $e^{y^\top\Lambda x_j}$ (as a key of the query $y$) and $e^{x_j^\top\Lambda y}$ (as a query reading $y$). 
The weights are the unnormalized version of~\eqref{eq:intro:transformer}. Without normalization the weights are not a probability, so that adding a key never decreases the contribution of the others, and the force grows with the number of keys similar to $y$. 
Moreover the vector read from a key is $\Lambda x_j$, so that values are tied to the query--key matrix.
\end{enumerate}
\end{proposition}

\begin{remark}[Connection with unnormalized self-attention]\label{rem:usa}
Item~\ref{item:d} is the statistical counterpart of the unnormalized self-attention dynamics of \cite[Section~1.2.3]{geshkovski2025perspective}. 
For $\Lambda=\beta\Id$, the kernel sum $\Phi_{k_{\beta\Id}}(\mu)=\iint e^{\beta\ip{x}{x'}}\,\mu(\De x)\,\mu(\De x')$ equals $2\beta\,\mrmE_\beta[\mu]$, where $\mrmE_\beta$ is the interaction energy of \cite[Eq.~(1.13)]{geshkovski2025perspective}. 
There the authors consider the unnormalized dynamics \textup{(USA)}, $\dot x_i=P^\perp_{x_i}\big(\frac1n\sum_je^{\beta\ip{x_i}{x_j}}x_j\big)$, which is the Wasserstein gradient flow of $\mrmE_\beta$ \cite[Lemma~1.8]{geshkovski2025perspective}. 
Accordingly, for the tilt $\mrmP\propto e^{\lambda\Phi_{k_{\beta\Id}}}\mbfN$, \Cref{lem:score} and \ref{item:d} show that the score of the last token is $\frac{2\lambda\beta}{n}$ times the \textup{(USA)} vector field evaluated at the candidate $y$, over the completed sentence $\mu_y$. 
The softmax normalization of attention divides this field by the scalar partition function (cf.~\eqref{eq:intro:transformer}), and does not change its direction: in \cite[Section 1.2.4]{geshkovski2025perspective} the normalized dynamics is recovered as a gradient flow of the same energy for a metric reweighted by the partition functions. 
In our setting, we will show how the normalization enters instead through the choice rule of \Cref{sec:choice}, at the level of the statistic.
\end{remark}

\begin{proof}
\ref{item:a} By symmetry and multilinearity, $t\mapsto\Phi(\mu+t(\varrho-\mu))=\sum_{k=0}^r\binom rk t^k\int f\,\De(\varrho-\mu)^{\otimes k}\otimes\mu^{\otimes(r-k)}$ is a polynomial. 
Since $f$ is symmetric,  we have $\frac{\De}{\De t}\Phi(\mu_t)=r\int f\,\De(\varrho-\mu)\otimes\mu_t^{\otimes(r-1)}$, with $\mu_t=\mu+t(\varrho-\mu)$. Therefore, $\delta\Phi(\mu)(z)=r\int f(z,\cdot)\De\mu^{\otimes (r-1)}$.
The function $\delta\Phi$ is bounded by $r\norm f_\infty$. If $f\in\mcC^1$, then the gradient of the first variation $r\int\nabla_zf(z,\cdot)\,\De\mu^{\otimes(r-1)}$ is bounded and jointly continuous in $(\mu,z)$. 

 Item \ref{item:b} can be obtained from \ref{item:a} applied to the symmetrization of $h$. Items \ref{item:c} and \ref{item:d} follow from direct computations.
\end{proof}

\begin{example}[Covariance]\label{ex:covariance}
Let $\Phi(\mu)=\Cov(\mu)=\int xx^\top\De\mu-m_\mu m_\mu^\top$ and $V\in\Mat$. By \Cref{prop:additive}, applied to $h(x,y)=x^\top Vx$, which depends on one variable, and to $h(x,y)=x^\top Vy$,
\[
\delta\ip{V}{\Cov}(\mu)(z)=z^\top Vz-z^\top(V+V^\top)m_\mu+\mathrm{const},\qquad\nabla_z\delta\ip{V}{\Cov}(\mu)(z)=(V+V^\top)(z-m_\mu).
\]
At the completed sentence $\mu_y$ of \eqref{eq:last-token}, $y-m_{\mu_y}=\frac{n-1}n(y-\bar x)$ with $\bar x=\frac1{n-1}\sum_{i<n}x_i$. Hence
\[
\nabla_z\delta\ip{V}{\Cov}(\mu_y)(z)\big|_{z=y}=\frac{n-1}n\,(V+V^\top)\big(y-\bar x\big),
\]
which is the formula of the introduction. Its tangential part, divided by $n$, is the score of the last token (\Cref{lem:score}). The context acts only through its mean $\bar x$.
\end{example}

\subsection{Choosing keys: the lowest-order relational choice rule}\label{sec:choice}
As we have already mentioned in the introduction, since linguistic structure is relational we introduce the concept of \emph{choice rule} for the sentence $\mu$, which is a coupling in
\[
\Pi_1(\mu)=\big\{\pi\in\mcP(\mbbS^d\times\mbbS^d):\ \pi(\cdot\times\mbbS^d)=\mu\big\},
\]
the first coordinate being the query and the second the key; we write $\pi(\De x,\De y)=\mu(\De x)\,\pi_x(\De y)$. Among all possible choice rules, the weakest form of competition we can 
consider in our model is Luce's choice axiom \cite{luce1959individual}, where
the odds of choosing the key $y$ rather than $y'$ depend on the query and on the two keys, and not on the other tokens of the sentence. Under this assumption we are guaranteed the existence of a 
\emph{choice function} $h:\mbbS^d\times\mbbS^d\to\R$, the same for all sentences, such that for any sentence $\mu$ the associated choice rule is
\begin{equation}\label{eq:score:rule:again}
\pi^h_\mu(\De x,\De y)=\mu(\De x)\,\frac{e^{h(x,y)}\,\mu(\De y)}{\int e^{h(x,y')}\,\mu(\De y')}\,.
\end{equation}
The choice function encodes relational preferences and syntactical relations between tokens, and in particular different choice functions might focus on different aspects of our language.
Notice that the rule depends on $h$ only through the log-odds $h(x,y)-h(x,y')$, so the choice function is defined up to the gauge shifts $h\mapsto h+a(x)$. 

In what follows we let $\msY_\ell$ denote the set of spherical harmonics of degree $\ell$ on $\mbbS^d$, so that $\msY_0$ are the constants and $\msY_1=\{x\mapsto v^\top x\}$. 
Finally, for any choice function $h\in\mrmL^2(\uvol\otimes\uvol)$ we consider its harmonic expansion $h=\sum_{\ell,\ell'\ge0}h_{\ell\ell'}$ with $h_{\ell\ell'}\in\msY_\ell\otimes\msY_{\ell'}$.

\begin{proposition}[The lowest-order relational interaction]\label{prop:harmonic}
Let $h,h'\in \mcC(\mbbS^d\times\mbbS^d)$.
\begin{enumerate}[(a)]
\item \label{item:lowest:gauge} $\pi^h_\mu=\pi^{h'}_\mu$ for every $\mu\in\mcP(\mbbS^d)$ if and only if $h-h'$ depends on the query only, that is, $h-h'\in\bigoplus_{\ell\ge0}\msY_\ell\otimes\msY_0$: the gauge removes exactly the components $h_{\ell0}$.
\item \label{item:lowest:key:bias} The components $h_{0\ell'}$, $\ell'\ge1$, depend on the key only: they shift the log-odds by the same amount for every query, a key bias without relational content.
\item \label{item:lowest:reletional} The relational part $h_{\mathrm{rel}}=\sum_{\ell,\ell'\ge1}h_{\ell\ell'}$ is gauge invariant, and its component of lowest total degree is $h_{11}\in\msY_1\otimes\msY_1=\{(x,y)\mapsto \langle x,\,\Lambda y\rangle:\ \Lambda\in\Mat\}$.
\end{enumerate}
\end{proposition}

\begin{proof}
Let us first show item~\ref{item:lowest:gauge}. Clearly, if $h-h'=a(x)$, the factor $e^{a(x)}$ cancels in \eqref{eq:score:rule:again}. 
Conversely, let $x\ne y$ and $\mu=\frac12(\delta_x+\delta_y)$. The query $x$ chooses the key $z\in\{x,y\}$ with probability $e^{h(x,z)}/(e^{h(x,x)}+e^{h(x,y)})$, and equality of 
the two rules gives $h(x,z)-h'(x,z)=\log\frac{e^{h(x,x)}+e^{h(x,y)}}{e^{h'(x,x)}+e^{h'(x,y)}}$ for both $z=x$ and $z=y$, hence $g(x,y)=g(x,x)$ for $g=h-h'$. 
Since $y\ne x$ is arbitrary, $g(x,\cdot)$ is constant for every $x$, that is, $g(x,y)=a(x)$ with $a(x)=g(x,x)$.
Finally, a function of the query alone has only components with $\ell'=0$, and conversely. 

Item~\ref{item:lowest:key:bias} is immediate.

Item~\ref{item:lowest:reletional} can be proven as follows. Since the gauge changes only components with $\ell'=0$, the relational part of the choice function $h_{\mathrm{rel}}=\sum_{\ell,\ell'\ge1}h_{\ell\ell'}$ is gauge invariant. 
Since the space $\msY_1$ consists of the restrictions of linear forms, we conclude that $\msY_1\otimes\msY_1$ is the space of bilinear forms $\{(x,y)\mapsto \langle x,\,\Lambda y\rangle :\ \Lambda\in\Mat\}$. 
Therefore, there exists a matrix $\Lambda\in\mbbM_{d+1}$ such that the lowest-order term in the harmonic expansion equals $h_{11}(x,y)=\langle x, \Lambda y\rangle\,\in\msY_1\otimes\msY_1$.

\end{proof}

Truncating the relational part of the choice function at its lowest order, and setting the key bias to zero, gives our \emph{lowest-order relational choice rule} 
$\pi^\Lambda_\mu$ of \eqref{eq:rule-intro}, with $k_\Lambda(x,y)=e^{\langle x,\Lambda y\rangle}$, $\Lambda\in\Mat$:
\begin{equation}\label{eq:rule-again}
\pi^\Lambda_\mu(\De x,\De y)=\mu(\De x)\,\frac{k_\Lambda(x,y)\,\mu(\De y)}{Z^\Lambda_\mu(x)},\qquad Z^\Lambda_\mu(x)=\int k_\Lambda(x,y)\,\mu(\De y)\,.
\end{equation}
The query $x$ chooses its key by softmax attention with 
query--key matrix $\Lambda$, and $\Lambda=0$ is the uniform choice. We stress that this truncation is a modeling choice. 
The component $h_{11}$ is the relational component of lowest degree, not in general the dominant one, and we plan to address this point of view in future work.

\subsection{Entropy: the lowest-order rule as an exponential family}\label{sec:entropy-rule}
For a fixed sentence, the rules \eqref{eq:rule-again} form an exponential family through the uniform choice,
\begin{equation}\label{eq:exp:fam:choice}
\frac{\De\pi^\Lambda_\mu}{\De(\mu\otimes\mu)}(x,y)=\exp\big(\ip{\Lambda}{xy^\top}-\log Z^\Lambda_\mu(x)\big),\qquad\Lambda\in\Mat,
\end{equation}
with natural parameter $\Lambda$ and sufficient statistic $xy^\top$. Its log-partition function, integrated over the queries, is the \emph{attention free energy}
\begin{equation}\label{eq:F-intro}
F_\Lambda(\mu)=\int\log Z^\Lambda_\mu(x)\,\mu(\De x).
\end{equation}
For notational convenience we write $Z_\mu=Z^\Lambda_\mu$. Finally, recall from~\eqref{eq:attn-intro} our definition of attention vector field and attention cross-moment associated to $\Lambda$:
\[
\Attn_\Lambda(x;\mu)\coloneqq\mbbE_{\pi^\Lambda_\mu}[Y|X=x]=\frac{\int y\,k_\Lambda(x,y)\,\mu(\De y)}{Z^\Lambda_\mu(x)}\,,\quad\,S_\Lambda(\mu)\coloneqq \mbbE_{\pi^\Lambda_\mu}[XY^\top]=\int xy^\top\De\pi^\Lambda_\mu\,.
\]
Note that $Z_\mu(x)$ and $\Attn_\Lambda(x;\mu)$ are defined for every $x\in\R^{d+1}$, not only on the support of $\mu$, and that $e^{-\norm\Lambda_\op}\le Z_\mu\le e^{\norm\Lambda_\op}$ on $\mbbS^d$.

\begin{proposition}[Attention as semi-relaxed entropic self-transport]\label{prop:gibbs-attn}
For every $\mu\in\mcP(\mbbS^d)$ and $\Lambda\in\Mat$, consider the entropic optimization problem 
\begin{equation}\label{eq:F-variational}
F_\Lambda(\mu)=\max_{\pi\in\Pi_1(\mu)}\Big\{\int x^\top\Lambda y\,\pi(\De x,\De y)-\rel{\pi}{\mu\otimes\mu}\Big\}\,.
\end{equation}
Then, $\pi^\Lambda_\mu$ corresponds to the unique maximizer of this problem. Moreover:
\begin{enumerate}[(a)]
\item \label{item:attention:EOT:a} $\mcH_\Lambda(\mu):=\rel{\pi^\Lambda_\mu}{\mu\otimes\mu}=\ip{\Lambda}{S_\Lambda(\mu)}-F_\Lambda(\mu)$, with $S_\Lambda(\mu)=\int xy^\top\De\pi^\Lambda_\mu$ the attention cross-moment;
\item \label{item:attention:EOT:b} $\Lambda\mapsto F_\Lambda(\mu)$ is convex and smooth, with $\nabla_\Lambda F_\Lambda(\mu)=S_\Lambda(\mu)$;
\item \label{item:attention:EOT:c} $\nabla_x\log Z^\Lambda_\mu(x)=\Lambda\Attn_\Lambda(x;\mu)$, whose tangential part $P^\perp_x\Lambda\Attn_\Lambda(x;\mu)$ is the self-attention vector field on $\mbbS^d$ with value matrix $\Lambda$;
\item \label{item:attention:EOT:d} the Jacobian of $x\mapsto\Attn_\Lambda(x;\mu)$ is $\Cov_{\pi^\Lambda_\mu(\cdot|x)}(Y)\Lambda^\top$, 
where, with a slight abuse of notation, for any $x\in\mbbS^d$ we write $\pi^\Lambda_\mu(\cdot|x)(\De y)=k_\Lambda(x,y)\,\mu(\De y)/Z_\mu(x)$ for the law of the key chosen by the query $x$, and
 $\Cov_{\pi^\Lambda_\mu(\cdot|x)}(Y)$ is its covariance.
\end{enumerate}
\end{proposition}

\begin{proof}
  Let us first notice that for any $\pi\in\Pi_1(\mu)$ we have 
  \begin{equation*}
   \int x^\top\Lambda y\,\pi(\De x,\De y)-\rel{\pi}{\mu\otimes\mu}=\int \log Z_\mu(x)\,\pi(\De x,\De y)-\rel{\pi}{\pi^\Lambda_\mu}=F_\Lambda(\mu)-\rel{\pi}{\pi^\Lambda_\mu}.
  \end{equation*}
Since $\rel{\pi}{\pi^\Lambda_\mu}\ge0$, with equality if and only if $\pi=\pi^\Lambda_\mu$, this shows that the supremum in \eqref{eq:F-variational} equals $F_\Lambda(\mu)$ and is attained only at $\pi^\Lambda_\mu$. 
Taking $\pi=\pi^\Lambda_\mu$ gives $\ip{\Lambda}{S_\Lambda(\mu)}-\mcH_\Lambda(\mu)=F_\Lambda(\mu)$ and hence Item~\ref{item:attention:EOT:a}, because $\int x^\top\Lambda y\,\De\pi^\Lambda_\mu=\ip{\Lambda}{\int xy^\top\De\pi^\Lambda_\mu}=\ip{\Lambda}{S_\Lambda(\mu)}$. 

Item~\ref{item:attention:EOT:b} can be proven as follows.  By \eqref{eq:F-variational}, $\Lambda\mapsto F_\Lambda(\mu)$ is a supremum of affine functions over a set independent of $\Lambda$, hence convex. 
For smoothness, note that $\Lambda\mapsto k_\Lambda(x,y)$ is smooth with $\nabla_\Lambda k_\Lambda(x,y)=xy^\top k_\Lambda(x,y)$, and all derivatives are bounded uniformly in $(x,y)\in\mbbS^d\times\mbbS^d$ for $\Lambda$ in a bounded set. 
Differentiating under the integral sign, and using $Z_\mu\ge e^{-\norm\Lambda_\op}$,
\[
\nabla_\Lambda\log Z_\mu(x)=\frac{\int xy^\top k_\Lambda(x,y)\,\mu(\De y)}{Z_\mu(x)}=x\,\Attn_\Lambda(x;\mu)^\top,
\qquad
\nabla_\Lambda F_\Lambda(\mu)=\int x\,\Attn_\Lambda(x;\mu)^\top\mu(\De x)=S_\Lambda(\mu).
\]

Item~\ref{item:attention:EOT:c} follows in a similar way: $\nabla_xk_\Lambda(x,y)=k_\Lambda(x,y)\,\Lambda y$ and hence
\[
\nabla_x\log Z^\Lambda_\mu(x)=\frac{\int\Lambda y\,k_\Lambda(x,y)\,\mu(\De y)}{Z_\mu(x)}=\Lambda\Attn_\Lambda(x;\mu),
\]
whose tangential part $P^\perp_x\Lambda\Attn_\Lambda(x;\mu)$ is the self-attention vector field on $\mbbS^d$ with value matrix $\Lambda$.

Item~\ref{item:attention:EOT:d} is a direct consequence of the properties of exponential families. 
Indeed, the derivative of the mean of the exponential family $\mu_\theta(\De y)\propto e^{\ip\theta y}\mu(\De y)$ with respect to its natural parameter $\theta$ is its covariance.

Therefore, if we consider the exponential family $\mu_{\Lambda^\top x}$, we have $\Attn_\Lambda(x;\mu)=\int y\,\mu_{\Lambda^\top x}(\De y)$, 
$\Cov_{\pi^\Lambda_\mu(\cdot|x)}(Y)=\Cov_{\mu_{\Lambda^\top x}}(Y)$ and, by the chain rule, the Jacobian of $x\mapsto\Attn_\Lambda(x;\mu)$ is then $\Cov_{\pi^\Lambda_\mu(\cdot|x)}(Y)\Lambda^\top$.

\end{proof}

Items~\ref{item:attention:EOT:a} and \ref{item:attention:EOT:b} identify the attention cross-moment $S_\Lambda(\mu)$ as the mean parameter of the family, and $\mcH_\Lambda=\ip{\Lambda}{S_\Lambda}-F_\Lambda$ as the Legendre transform of the free energy evaluated at $S_\Lambda(\mu)$. 
The constrained counterpart of \eqref{eq:F-variational} is a conditional maximum-entropy problem. We state it for general bounded query and key features $\psi(x)\in\R^p$ and $\chi(y)\in\R^q$, which shows that our model can be further generalized to include feature embeddings. For the sake of clarity, in this paper we work with $\psi=\chi=\mathrm{id}$, which gives the rule \eqref{eq:rule-intro}.

\begin{proposition}[Conditional maximum entropy]\label{prop:condmaxent}
Let $\psi,\chi$ be bounded and measurable, $\Lambda\in\R^{p\times q}$, and
\[
\pi^{\Lambda}_\mu(\De x,\De y)=\mu(\De x)\,\frac{e^{\psi(x)^\top\Lambda\chi(y)}\,\mu(\De y)}{\int e^{\psi(x)^\top\Lambda\chi(y')}\mu(\De y')},\qquad C_\Lambda=\int\psi(x)\chi(y)^\top\,\pi^\Lambda_\mu(\De x,\De y).
\]
Then $\pi^\Lambda_\mu$ is the unique minimizer of $\rel{\pi}{\mu\otimes\mu}$ over $\{\pi\in\Pi_1(\mu):\int\psi\chi^\top\De\pi=C_\Lambda\}$.
\end{proposition}

\begin{proof}
By reasoning as in the previous proof, for $\pi\in\Pi_1(\mu)$ with finite entropy we can show that 
\[
\rel{\pi}{\mu\otimes\mu}=\rel{\pi}{\pi^\Lambda_\mu}+\ip{\Lambda}{\textstyle\int\psi\chi^\top\De\pi}-\int\log\Big(\int e^{\psi(x)^\top\Lambda\chi(y)}\mu(\De y)\Big)\mu(\De x).
\]
On the constraint set the last two terms do not depend on $\pi$, which means that minimizing $\rel{\pi}{\mu\otimes\mu}$ on the constrained set is 
equivalent to minimizing $\rel{\pi}{\pi^\Lambda_\mu}$ there, which admits the choice rule $\pi^\Lambda_\mu$ as unique minimizer.
\end{proof}

This result describes the choice rule $\pi^\Lambda_\mu$ as a conditional entropic projection \cite{csiszar1975}: the least informative rule with its own query--key cross-moment. If $\mu=\frac1n\sum_{i\le n}\delta_{x_i}$ with distinct tokens and $\pi\in\Pi_1(\mu)$ chooses the key $x_j$ for the query $x_i$ with probability $p_{ij}$, then
\[
\rel{\pi}{\mu\otimes\mu}=\log n+\frac1n\sum_{i,j}p_{ij}\log p_{ij}=\log n-H(Y\,|\,X),
\]
where $H(Y\,|\,X)$ is the Shannon entropy of the key given the (uniformly distributed) query. Minimizing the relative entropy under a constraint 
$\int\psi\chi^\top\De\pi=C$ is therefore the conditional maximum-entropy problem of \cite[\S3]{berger1996maximum}, with context law $\mu$ and features
 $\psi\chi^\top$. By Lagrangian duality, if $C$ lies in the relative interior of the set of attainable cross-moments $\{\int\psi\chi^\top\De\pi:\ \pi\in\Pi_1(\mu)\}$,
the constrained minimizer is $\pi^\Lambda_\mu$ for any maximizer $\Lambda$ of the concave function $\Lambda\mapsto\ip{\Lambda}{C}-F_\Lambda(\mu)$, 
whose maximizers are characterized by $\int\psi\chi^\top\De\pi^\Lambda_\mu=C$ \cite{csiszar1975,berger1996maximum}. 
The maximizer need not be unique, but the minimizer $\pi^\Lambda_\mu$ is, by \Cref{prop:condmaxent}. In words, the lowest-order relational choice rule is the least 
informative rule that is informed of the query--key cross-moment.

\begin{remark}[One marginal versus two: the difference with Sinkformers]\label{rem:sinkhorn}
Restricting \eqref{eq:F-variational} to couplings with both marginals equal to $\mu$ gives the entropic optimal transport (Schr\"odinger) problem with reward $x^\top\Lambda y$ 
\cite{Schr,Schr32,LeoSch,nutz2021entropic}, whose optimizer is the doubly stochastic attention of Sinkformers \cite{sander2022sinkformers}. 
Softmax attention is the \emph{semi-relaxed} problem: the query marginal is fixed, while the key marginal is $K^{*}_\mu(y)\,\mu(\De y)$, with
\begin{equation}\label{eq:indegree}
K^{*}_\mu(y)=\int \frac{k_\Lambda(x,y)}{Z_\mu(x)}\,\mu(\De x),
\end{equation}
the attention that the sentence pays to $y$.
\end{remark}

\subsection{Attention cross-moment and readouts in the minimization problem}\label{sec:cross:read:proj}

The choice rule enters the language model through the mean parameter of the family~\eqref{eq:exp:fam:choice}, the attention cross-moment $S_\Lambda(\mu)=\nabla_\Lambda F_\Lambda(\mu)$ of \eqref{eq:S-intro}. 
It is sufficient for the rule: by \Cref{prop:condmaxent}, $\pi^\Lambda_\mu$ is the only least informative rule with cross-moment $S_\Lambda(\mu)$, so $\mu$ and $S_\Lambda(\mu)$ determine the plan. It also collects every lowest-order relational readout of the choice: for $g(x,y)=x^\top Vy\in\msY_1\otimes\msY_1$ one has $\mbbE_{\pi^\Lambda_\mu}[g]=\ip{V}{S_\Lambda(\mu)}$.

\begin{definition}[Attention readout]\label{def:readout}
For $\Lambda,V\in\Mat$, the \emph{attention readout} of a sentence $\mu\in\mcP(\mbbS^d)$ is
\begin{equation}\label{eq:readout}
\Psi_{\Lambda,V}(\mu)=\ip{V}{S_\Lambda(\mu)}=\mbbE_{\pi^\Lambda_\mu}\big[X^\top VY\big]=\int x^\top V\Attn_\Lambda(x;\mu)\,\mu(\De x).
\end{equation}
\end{definition}

The attention readout naturally appears in the Gibbs tilts of \Cref{thm:iproj}, with $\Phi=S_\Lambda$. In particular, their laws are $\mrmP_{\Lambda,V}\propto e^{\Psi_{\Lambda,V}}\mbfN$, the value matrix $V$ being the multiplier (cf. \Cref{thm:fixed}). 
In $\Psi_{\Lambda,V}$ the attention plan is chosen with $\Lambda$ and its payoff is evaluated with $V$.

In the next lemma we collect some useful properties of the attention cross-moment and readout.

\begin{lemma}\label{lem:S-properties}
Let $\Lambda,V\in\Mat$ and $\mu\in\mcP(\mbbS^d)$.
\begin{enumerate}[(a)]
\item  \label{item:S-prop:a} $\norm{S_\Lambda(\mu)}_\op\le1$ and $\abs{\Psi_{\Lambda,V}(\mu)}\le\norm V_\op$. Moreover $S_\Lambda(\delta_x)=xx^\top$ and $\Psi_{\Lambda,V}(\delta_x)=x^\top Vx$.
\item  \label{item:S-prop:b} $\mu\mapsto S_\Lambda(\mu)$ is weakly continuous. More precisely, if $\mu_k\to\mu$ weakly, then $Z_{\mu_k}$, $\Attn_\Lambda(\cdot;\mu_k)$ and $\Cov_{\pi^\Lambda_{\mu_k}(\cdot|x)}(Y)$ converge uniformly on $\mbbS^d$ to $Z_\mu$, $\Attn_\Lambda(\cdot;\mu)$ and $\Cov_{\pi^\Lambda_\mu(\cdot|x)}(Y)$.
\end{enumerate}
\end{lemma}

\begin{proof}
Statement~\ref{item:S-prop:a} is immediate. Statement~\ref{item:S-prop:b} follows from $\{k_\Lambda(x,\cdot)\}_{x\in\mbbS^d}$ being uniformly bounded and equicontinuous, and from $Z_\mu\ge e^{-\norm\Lambda_\op}$.
\end{proof}

\bigskip

To apply \Cref{thm:iproj} to the statistic $S_\Lambda$ we need it to be non-degenerate on the support of the null model, i.e.\ $\ip{A}{S_\Lambda}$ must not be constant on $\mcS_{\mbfN}$ unless $A=0$. This holds for every non-zero choice matrix.

\begin{proposition}[Non-degeneracy]\label{prop:nondeg-main}
Let $\Lambda\ne0$ and assume that $\nu(m)>0$ for some $m\ge3$. If $\ip{A}{S_\Lambda}$ is constant on $\mcS_{\mbfN}$, then $A=0$.
\end{proposition}

The proof, in \Cref{app:non-deg}, restricts $\ip{A}{S_\Lambda}$ to two-point sentences $(1-t)\delta_x+t\delta_y$. 
The softmax normalization produces factors $e^{x^\top\Lambda(y-x)}$ which, along great circles, are exponentials of trigonometric polynomials and cannot be compensated by polynomial terms. 
No symmetry of $\Lambda$ is needed, whereas for $\Lambda=0$, i.e.\ for the uniform choice $\mu\otimes\mu$, $\ip{A}{S_0}$ is constant for every antisymmetric $A$. 
The bound $m\ge3$ comes from the rational dependence of $S_\Lambda$ on the weights of a sentence.

\bigskip

\Cref{prop:nondeg-main} allows us to consider for any non-zero choice matrix $\Lambda\in\mbbM_{d+1}$  the minimization problem \eqref{eq:entropic-intro} with constraining statistic $\Phi=S_\Lambda$, that is, the problem of finding 
\[
R_S(\Lambda)=\inf\big\{\rel{\mrmP}{\mbfN}:\ \mbbE_\mrmP [S_\Lambda]=\mbbE_{\mbfP^\star}[S_\Lambda]\big\}
\]
which corresponds to the amount of information that must be added to the null model to reproduce the attention cross-moment of the language.
$\mbfP^\star\in\mcP(\mcP(\mbbS^d))$ is the language law which is generally unknown, and in practice $\mbbE_{\mbfP^\star}[S_\Lambda]$ is replaced by its empirical counterpart over a corpus $\widehat{S_\Lambda}$.

We are now ready to specialize the abstract results from \Cref{sec:preliminaries} to our language model setting. In this setting the exponential family associated to the minimization problem is the exponential family with sufficient statistic $S_\Lambda$
and parametrized by value matrices $V\in\mbbM_{d+1}$
\[
\mrmP_{\Lambda,V}=e^{\Psi_{\Lambda,V}-\Gamma(\Lambda,V)}\,\mbfN,\quad\text{with }\Gamma(\Lambda,V)=\log\mbbE_{\mbfN}\big[e^{\Psi_{\Lambda,V}}\big]
\]
Further recall the definition of the closed convex hull
$\mcK_\Lambda=\overline{\mathrm{conv}}\,S_\Lambda\big(\mcS_{\mbfN}\big)\subset\Mat$, associated to the cross-moment statistic, where $\mcS_\mbfN$ is the support of the prior model $\mbfN$.

\begin{theorem}[Entropic minimization of language laws with cross-moment statistics]\label{thm:fixed}
Assume that $\nu(m)>0$ for some $m\ge3$ and let $\Lambda\ne0$.
\begin{enumerate}[(a)]
\item\label{item:fixed:a} $V\mapsto\mcM_\Lambda(V):=\mbbE_{\mrmP_{\Lambda,V}}S_\Lambda$ is a real-analytic diffeomorphism of $\Mat$ onto the interior of $\mcK_\Lambda$. Its differential at $V$ is the covariance operator $\Cov_{\mrmP_{\Lambda,V}}(S_\Lambda)$, which is positive definite on $\Mat$.
\item\label{item:fixed:b} If $\mbbE_{\mbfP^\star}S_\Lambda\in\mathrm{int}\,\mcK_\Lambda$, the problem defining $R_S(\Lambda)$ has the unique solution $\mrmP_{\Lambda,V^\star(\Lambda)}$, where $V^\star(\Lambda)=\mcM_\Lambda^{-1}(\mbbE_{\mbfP^\star}S_\Lambda)$ is the unique value matrix with $\mbbE_{\mrmP_{\Lambda,V^\star}}S_\Lambda=\mbbE_{\mbfP^\star}S_\Lambda$, and
\[
R_S(\Lambda)=\sup_{V\in\Mat}\Big\{\ip{V}{\mbbE_{\mbfP^\star}S_\Lambda}-\Gamma(\Lambda,V)\Big\}.
\]
Every admissible $\mrmP$ satisfies $\rel{\mrmP}{\mbfN}=\rel{\mrmP}{\mrmP_{\Lambda,V^\star}}+R_S(\Lambda)$.
\item\label{item:fixed:c} (Every value matrix is realized) For every $V\in\Mat$, the language $\mbfP^\star=\mrmP_{\Lambda,V}$ has value matrix $V^\star(\Lambda)=V$. In particular, the value matrix is not determined by the query--key matrix: query--key and value matrices are untied.
\end{enumerate}
\end{theorem}

\begin{proof}
The first two claims follow by applying \Cref{thm:iproj} to $\Phi=S_\Lambda\in\Mat\cong\R^{(d+1)^2}$, which is bounded and weakly continuous by \Cref{lem:S-properties}, and $\nu$-non-degenerate by \Cref{prop:nondeg-main}. 
As regards \Cref{item:fixed:c}, for $\mbfP^\star=\mrmP_{\Lambda,V}$ we have $\mbbE_{\mbfP^\star}S_\Lambda=\mcM_\Lambda(V)\in\mathrm{int}\,\mcK_\Lambda$, so \ref{item:fixed:b} applies and $V^\star(\Lambda)=\mcM_\Lambda^{-1}(\mcM_\Lambda(V))=V$.
\end{proof}

\begin{remark}[Interior constraints for a corpus]\label{rem:interior}
In practice the constraint of \Cref{thm:fixed}-\ref{item:fixed:b} is the empirical cross-moment $\widehat{S_\Lambda}=\frac1J\sum_{j=1}^JS_\Lambda(\mu_j)$ of a corpus of sentences $\mu_1,\dots,\mu_J\in\mcS_{\mbfN}$, 
and it is natural to ask when it lies in $\inter\mcK_\Lambda$. A simple sufficient condition is that the centered matrices $C_j=S_\Lambda(\mu_j)-\widehat{S_\Lambda}$, $j\le J$, span $\Mat$. Since $\sum_jC_j=0$, these $J$ matrices span a space of dimension at most $J-1$, so this requires $J\ge(d+1)^2+1$ sentences, in general position.

To see why the condition suffices, assume by contradiction that $\widehat{S_\Lambda}$ is not an interior point of the closed convex set $\mcK_\Lambda$. By the supporting hyperplane theorem, there is a direction $A\in\Mat\setminus\{0\}$ along which $\widehat{S_\Lambda}$ is extremal in $\mcK_\Lambda$, that is, $\ip{A}{S}\le\ip{A}{\widehat{S_\Lambda}}$ for every $S\in\mcK_\Lambda$. Each $S_\Lambda(\mu_j)$ belongs to $\mcK_\Lambda$, so the numbers $\ip{A}{\widehat{S_\Lambda}}-\ip{A}{S_\Lambda(\mu_j)}=-\ip{A}{C_j}$ are all non-negative, and their average is zero. Hence they all vanish: $A$ is orthogonal to every $C_j$, which contradicts the spanning assumption. 
\end{remark}

\begin{remark}[Two multipliers of one moment]\label{rem:two-multipliers}
  In an attention head the query, key, value and output projections enter only through the two products $Q^\top K$ and $W_OW_V$, which play independent roles: the first determines the attention pattern, the second what is transmitted along it. In our model these are exactly the two parameters, $\Lambda=Q^\top K$ and $V$, and they arise as the multipliers of two different variational problems.
Indeed, for a fixed sentence, \Cref{prop:condmaxent} identifies $\Lambda$ as the Lagrange multiplier of the constraint $\mbbE_\pi[XY^\top]=C$ on choice rules, 
the constraint being met with $C=S_\Lambda(\mu)$. Over the language, $V^\star$ is the Lagrange multiplier of the same moment averaged over sentences. 
The two multipliers are determined by different problems, one inside each sentence and one across sentences, and in the cross-moment model nothing relates them.
\end{remark}

\begin{remark}[Tied values are the exception]\label{rem:tied}
Under the assumptions of \Cref{thm:fixed}, the data cross-moments $c\in\mathrm{int}\,\mcK_\Lambda$ for which the value matrix is tied to the attention matrix, $V^\star(\Lambda)\in\R\Lambda$, 
form the curve $\mcM_\Lambda(\R\Lambda)$. Since $\mcM_\Lambda$ is a real-analytic diffeomorphism onto the open set $\mathrm{int}\,\mcK_\Lambda$ and $\R\Lambda$ is a closed embedded line, 
this curve is a closed one-dimensional embedded real-analytic submanifold of $\mathrm{int}\,\mcK_\Lambda$; as $(d+1)^2>1$, it has empty interior and zero Lebesgue measure in $\Mat\cong\R^{(d+1)^2}$. 
At a fixed attention matrix the value matrix is thus a free parameter, whatever matrix makes the model reproduce the cross-moment of the language, and tied values are the exception. This is in contrast with the free-energy statistic of \Cref{rem:free-energy}, for which values are always tied.
\end{remark}

We now compute the first variation of the attention readout and its gradient, which by \Cref{lem:score} is the force that a new token undergoes, conditioned on the presence of context in our model.

\begin{proposition}[First variation of the attention readout]\label{prop:first-variation}
Let $\Lambda,V\in\Mat$. Then $\Psi_{\Lambda,V}$ satisfies the hypotheses of \Cref{lem:score}: it is weakly continuous, 
and the differentiability conditions \ref{ass:lemma:item}--\ref{ass:lemma:item:ii} hold with
\begin{equation}\label{eq:first-variation-psi}
\delta\Psi_{\Lambda,V}(\mu)(z)=z^\top V\Attn_\Lambda(z;\mu)+\int \frac{k_\Lambda(x,z)}{Z_\mu(x)}\,x^\top V\big(z-\Attn_\Lambda(x;\mu)\big)\,\mu(\De x),\qquad \mu\big(\delta\Psi_{\Lambda,V}(\mu)\big)=\Psi_{\Lambda,V}(\mu),
\end{equation}
$\abs{\delta\Psi_{\Lambda,V}(\mu)(z)}\le\norm V_\op\big(1+2e^{2\norm\Lambda_\op}\big)$ for $z\in\mbbS^d$, and, for $z\in\R^{d+1}$,
\begin{align}\label{eq:force-psi}
  &\,\nabla_z\,\delta\Psi_{\Lambda,V}(\mu)(z)=V \Attn_\Lambda(z;\mu)+V^\top\!\!\int x\,\frac{k_\Lambda(x,z)}{Z_\mu(x)}\,\mu(\De x)+\rho_\mu(z),\\
  &\, \text{with }\rho_\mu(z)=\Lambda\,\Cov_{\pi^\Lambda_\mu(\cdot|z)}(Y)\,V^\top z+\Lambda^\top\!\!\int \frac{k_\Lambda(x,z)}{Z_\mu(x)}\,\big[x^\top V\big(z-\Attn_\Lambda(x;\mu)\big)\big]\,x\,\mu(\De x),\notag
  \end{align}
where we recall our abuse of notation for $\pi^\Lambda_\mu(\cdot|z)(\De y)=k_\Lambda(z,y)\,\mu(\De y)/Z_\mu(z)$, which indicates the law of the key chosen by any query $z\in\mbbS^d$.
\end{proposition}

\begin{proof}
The weak continuity of the readout $\Psi_{\Lambda,V}$ directly follows from \Cref{lem:S-properties}-\ref{item:S-prop:b}.
For notational convenience let $N_\mu(x)\coloneqq \int y k_\Lambda(x,y)\mu(\De y)$ so that $\Attn_\Lambda(\cdot;\mu)=N_\mu/Z_\mu$. 
Fix $\mu,\varrho\in\mcP(\mbbS^d)$, consider the line $\mu_t=\mu+t(\varrho-\mu)$ for $t\in[0,1]$ and let $N_t=N_{\mu_t}$ and $Z_t=Z_{\mu_t}$. 
These two are affine functions in $t$, with $\dot N_t(x)=\int y\,k_\Lambda(x,y)\,(\varrho-\mu)(\De y)$, $\dot Z_t(x)=\int k_\Lambda(x,y)\,(\varrho-\mu)(\De y)$ and $Z_t\ge e^{-\norm\Lambda_\op}$ on $\mbbS^d$. Hence $\Attn_\Lambda(\cdot;\mu_t)=N_t/Z_t$ is smooth in $t$, with
\[
\partial_t\Attn_\Lambda(x;\mu_t)=\frac{\dot N_t(x)-\Attn_\Lambda(x;\mu_t)\dot Z_t(x)}{Z_t(x)}=\int \frac{k_\Lambda(x,y)}{Z_t(x)}\,\big(y-\Attn_\Lambda(x;\mu_t)\big)\,(\varrho-\mu)(\De y).
\]
Since $\mu_t$ is affine in $t$ and $\Attn_\Lambda(\cdot;\mu_t)$, $\partial_t\Attn_\Lambda(\cdot;\mu_t)$ are bounded on $\mbbS^d$ uniformly in $t$, 
we may differentiate $\Psi_{\Lambda,V}(\mu_t)=\int x^\top V\Attn_\Lambda(x;\mu_t)\,\mu_t(\De x)$ under the integral sign, which gives the first variation expression in~\eqref{eq:first-variation-psi}.
The identity $\mu(\delta\Psi_{\Lambda,V}(\mu))=\Psi_{\Lambda,V}(\mu)$ follows from $\int k_\Lambda(x,z)\,\mu(\De z)=Z_\mu(x)$ and $\int k_\Lambda(x,z)\,z\,\mu(\De z)=Z_\mu(x)\Attn_\Lambda(x;\mu)$.

On $\mbbS^d$, $\abs{z^\top V\Attn_\Lambda(z;\mu)}\le\norm V_\op$, $k_\Lambda(x,z)/Z_\mu(x)\le e^{\norm\Lambda_\op}/e^{-\norm\Lambda_\op}=e^{2\norm\Lambda_\op}$ and $\abs{z-\Attn_\Lambda(x;\mu)}\le2$, which gives the bound on $\delta\Psi_{\Lambda,V}$.

The formula \eqref{eq:first-variation-psi} defines a smooth function of $z\in\R^{d+1}$, since $Z_\mu(z)>0$. 
By \Cref{prop:gibbs-attn}-\ref{item:attention:EOT:d}, the Jacobian of $z\mapsto\Attn_\Lambda(z;\mu)$ is $\Cov_{\pi^\Lambda_\mu(\cdot|z)}(Y)\,\Lambda^\top$, and the covariance is symmetric, so that
\[
\nabla_z\big(z^\top V\Attn_\Lambda(z;\mu)\big)=V\Attn_\Lambda(z;\mu)+\Lambda\,\Cov_{\pi^\Lambda_\mu(\cdot|z)}(Y)\,V^\top z .
\]
Moreover $\nabla_z\big(k_\Lambda(x,z)/Z_\mu(x)\big)=\big(k_\Lambda(x,z)/Z_\mu(x)\big)\Lambda^\top x$, with $\mu$ and $x$ fixed. This gives \eqref{eq:force-psi}, the gradient of the extension of $\delta\Psi_{\Lambda,V}(\mu)$ to $\R^{d+1}$ given by \eqref{eq:first-variation-psi}.

It remains to check that $(\mu,z)\mapsto\nabla_z\delta\Psi_{\Lambda,V}(\mu)(z)$ is jointly continuous on $\mcP(\mbbS^d)\times\mbbS^d$. By \Cref{lem:S-properties}-\ref{item:S-prop:b}, the functions $Z_\mu$, $\Attn_\Lambda(\cdot;\mu)$ and $\Cov_{\pi^\Lambda_\mu(\cdot|\,\cdot)}(Y)$ 
depend continuously on $(\mu,z)$, and the integrals in \eqref{eq:force-psi} are of the form $\int g_\mu(x,z)\,\mu(\De x)$ with $g_\mu$ continuous and depending continuously on $\mu$, uniformly in $(x,z)$; hence they are jointly continuous as well.
In particular, using $k_\Lambda(x,z)/Z_\mu(x)\le e^{2\norm\Lambda_\op}$ and $\norm{\Cov_{\pi^\Lambda_\mu(\cdot|z)}(Y)}_\op\le1$, one gets $\abs{\nabla_z\delta\Psi_{\Lambda,V}(\mu)(z)}\le\norm V_\op\big(1+\norm\Lambda_\op\big)\big(1+2e^{2\norm\Lambda_\op}\big)$ on $\mbbS^d$.

\end{proof}

Let us interpret these formulas. The first variation \eqref{eq:first-variation-psi} has two terms. 
The first, $z^\top V\Attn_\Lambda(z;\mu)$, is the alignment, through $V$, between $z$ and what it reads as a query. The second,
\[
\int\frac{k_\Lambda(x,z)}{Z_\mu(x)}\,x^\top V\big(z-\Attn_\Lambda(x;\mu)\big)\,\mu(\De x),
\]
measures the usefulness of $z$ as a key: each token $x$ that attends to $z$, with weight $k_\Lambda(x,z)/Z_\mu(x)$, compares its alignment with $z$ to its alignment with what it currently reads.

Correspondingly, the force \eqref{eq:force-psi} on $z$ has three terms.
\begin{enumerate}[(i)]
\item The query term $V\Attn_\Lambda(z;\mu)$: an average of the keys with softmax weights, which compete through the normalization $Z_\mu(z)$, transported by the value matrix $V$. This is the transformer head \eqref{eq:intro:transformer}.
\item The key-side term $V^\top\!\int x\,\frac{k_\Lambda(x,z)}{Z_\mu(x)}\,\mu(\De x)$: the tokens that attend to $z$, weighted by the attention they pay to it and read through $V^\top$.
\item The correction $\rho_\mu(z)\in\mathrm{range}\,\Lambda+\mathrm{range}\,\Lambda^\top$, which records that moving $z$ changes not only what is read but also where tokens look. Its query part $\Lambda\,\Cov_{\pi^\Lambda_\mu(\cdot|z)}(Y)\,V^\top z$ is the gradient of the alignment through the attention weights of $z$; it vanishes when $z$ attends to a single key, since the covariance is then zero. Its key part is the gradient of the usefulness through the weights $k_\Lambda(x,z)/Z_\mu(x)$ with which the other tokens read $z$.
\end{enumerate}

\begin{remark}[The correction does not vanish]\label{rem:rho}
For $\mu=\delta_x$ one has $\Cov_{\pi_\mu^\Lambda(\cdot|x)}=0$ and $\Attn_\Lambda(x;\mu)=x$, so $\rho_{\delta_x}(z)=e^{x^\top\Lambda(z-x)}\,x^\top V(z-x)\,\Lambda^\top x$.
If $\rho\equiv0$, then $x^\top V(z-x)=0$ for every $z$ whenever $\Lambda^\top x\ne0$, that is, $V^\top x=0$ on a dense set of $x$ when $\Lambda\ne0$. 
So $\rho$, as a function of $(\mu,z)$, vanishes identically only if $\Lambda=0$ or $V=0$. 
In the free-energy model (\Cref{rem:free-energy}), by contrast, the plan is optimal for the payoff, and there is no such correction.
\end{remark}

We now combine \Cref{prop:last-token}, \Cref{lem:score}, and \Cref{prop:first-variation} in order to get the last token law and its score under any Gibbs tilt $\mrmP_{\Lambda,V}$. For any fixed $\Lambda$, this in particular applies to $V^\star=V^\star(\Lambda)$ as in \Cref{thm:fixed}, and hence to our optimizer $\mrmP_{\Lambda,V^\star}$ for \eqref{eq:entropic-intro}, which corresponds to Item~\ref{item:meta:sym:iii} of the Meta-Theorem for the unordered model.

\begin{theorem}[Last token of the attention model]\label{thm:last}
Let $\Lambda,V\in\Mat$, $n\ge1$ with $\nu(n)>0$, $x_{1:n-1}\in(\mbbS^d)^{n-1}$, and $\mu_y=\frac1n(\sum_{i<n}\delta_{x_i}+\delta_y)$.
\begin{enumerate}[(a)]
\item \label{item:last:att:a} (Exact law) Under $\mrmP_{\Lambda,V}$, given $N=n$ and $X_{1:n-1}=x_{1:n-1}$, the last token has density with respect to $\uvol$
\[
p(y)=\frac{\exp\big(\Psi_{\Lambda,V}(\mu_y)\big)}{\int\exp\big(\Psi_{\Lambda,V}(\mu_{y'})\big)\,\uvol(\De y')},\qquad y\in\mbbS^d .
\]
\item \label{item:last:att:b} (Score) $\log p$ is smooth on $\mbbS^d$, and its Riemannian gradient is
\begin{equation}\label{eq:score-attention}
\nabla_{\mbbS^d}\log p(y)=\frac1n\,P^\perp_y\Big(V\Attn_\Lambda(y;\mu_y)+V^\top\!\!\int x\,\frac{e^{\langle x,\Lambda y\rangle}}{\int e^{\langle x,\Lambda z'\rangle}\mu_y(\De z')}\,\mu_y(\De x)+\rho_{\mu_y}(y)\Big),
\end{equation}
with $\rho$ as in \eqref{eq:force-psi}.
\item \label{item:last:att:c} (Transformer reading) For $\Lambda=Q^\top K$, the first term in \eqref{eq:score-attention} is the transformer map \eqref{eq:intro:transformer} with query matrix $Q$, key matrix $K$ and value matrix $V$, applied to the candidate $y$ as a query over the completed sentence:
\[
V\Attn_\Lambda(y;\mu_y)=\sum_{j=1}^n\frac{e^{\ip{Qy}{Kx_j}}}{\sum_{\ell=1}^ne^{\ip{Qy}{Kx_\ell}}}\,Vx_j,\qquad x_n=y .
\]
\end{enumerate}
\end{theorem}

\begin{proof}
Item~\ref{item:last:att:a} is \Cref{prop:last-token}-\ref{item:last:gen:b} with $\Phi=S_{\Lambda}$, which is bounded and weakly continuous by \Cref{lem:S-properties}. 

Claim \ref{item:last:att:b} can be proven as follows. The smoothness of $\log p$ directly follows from the one of $y\mapsto\Psi_{\Lambda,V}(\mu_y)=\frac1n\sum_{i\le n}x_i^\top V\Attn_\Lambda(x_i;\mu_y)$, with $x_n=y$, 
which is a finite combination of smooth functions of $y$ (because all the denominators are at least $e^{-\norm\Lambda_\op}$). 
Then, by \Cref{prop:first-variation}, $\Psi_{\Lambda,V}$ is $\mcC^1$ in the sense of \Cref{lem:score} (differentiability conditions \ref{ass:lemma:item}--\ref{ass:lemma:item:ii}),
which gives $\nabla_{\mbbS^d}\log p(y)=\frac1nP^\perp_y\nabla_z\delta\Psi_{\Lambda,V}(\mu_y)(z)|_{z=y}$. 
Combining this last identity with \eqref{eq:force-psi} at $\mu=\mu_y$ and $z=y$ gives~\eqref{eq:score-attention}.

Item~\ref{item:last:att:c} is a straightforward computation.

\end{proof}

\begin{remark}[All tokens at once]\label{rem:particles}
\Cref{lem:score} applies to any token of a sentence, not only to the last one. Let $\mu=\frac1n\sum_{j\le n}\delta_{z_j}$ and fix $i\le n$. Taking the other tokens $(z_j)_{j\ne i}$ as context and $y=z_i$, the completed sentence $\mu_y$ of \Cref{lem:score} is $\mu$ itself, so that the Riemannian gradient of $z_i\mapsto\Psi_{\Lambda,V}(\mu)$ is
\[
\nabla_{z_i}\Psi_{\Lambda,V}(\mu)=\frac1n\,P^\perp_{z_i}\,\nabla_z\delta\Psi_{\Lambda,V}(\mu)(z)\big|_{z=z_i}.
\]
Hence gradient ascent of $\Psi_{\Lambda,V}$ on $(\mbbS^d)^n$, accelerated by the factor $n$, moves every token $z_i$ along the vector field $z\mapsto P^\perp_z\nabla_z\delta\Psi_{\Lambda,V}(\mu)(z)$ of \eqref{eq:force-psi}, which is the Wasserstein gradient of $\Psi_{\Lambda,V}$ on $\mcP(\mbbS^d)$. In the measure-to-measure picture of transformers \cite{sander2022sinkformers,geshkovski2023emergence,geshkovski2025perspective,geshkovski2026measure}, a layer instead moves each token along the self-attention field $P^\perp_zV\Attn_\Lambda(z;\mu)$. By \eqref{eq:force-psi}, this is only the query term of the Wasserstein gradient of the attention readout: the key-side term and the correction $\rho_\mu$ are what make the field an exact gradient.
\end{remark}

\subsection{Concluding remarks on training and free energy}\label{sec:training}

\begin{remark}[Training]\label{rem:training}
In this paper the attention matrix $\Lambda$ is fixed, while the value matrix is not tuned independently of the data: it is the Lagrange multiplier $V^\star(\Lambda)$ of \Cref{thm:fixed}, determined by the constraint $\mbbE_{\mrmP}S_\Lambda=\mbbE_{\mbfP^\star}S_\Lambda$. It is also the maximum-likelihood value matrix: for the expected log-likelihood ratio of the model against the null model,
\[
\mcL(\Lambda,V)=\mbbE_{\mbfP^\star}\Big[\log\frac{\De\mrmP_{\Lambda,V}}{\De\mbfN}\Big]=\mbbE_{\mbfP^\star}\big[\Psi_{\Lambda,V}\big]-\log\mbbE_{\mbfN}\big[e^{\Psi_{\Lambda,V}}\big],
\]
the map $V\mapsto\mcL(\Lambda,V)$ is concave, its maximizer is $V^\star(\Lambda)$, and its maximum is $R_S(\Lambda)$, by \Cref{thm:fixed} and convex duality. Only $\Lambda$ remains to be learned, and the natural criterion is
\[
\Lambda^\star\in\argmax_\Lambda\ \mcL\big(\Lambda,V^\star(\Lambda)\big)=\argmax_\Lambda\ \inf\big\{\rel{\mrmP}{\mbfN}:\ \mbbE_\mrmP S_\Lambda=\mbbE_{\mbfP^\star}S_\Lambda\big\},
\]
over a compact set of admissible attention matrices: $\Lambda^\star$ selects the attention cross-moment least compatible with the null model, and $\mrmP_{\Lambda^\star,V^\star(\Lambda^\star)}$ is the least informative model reproducing it. 
This is the minimax-entropy principle of \cite{zhu1997minimax}. 
The maximization is over a compact capacity set $\mfC\subset\Mat$, as in practice (norm or rank constraints on $\Lambda=Q^\top K$). A maximizer then exists as soon as $\Lambda\mapsto\mcL(\Lambda,V^\star(\Lambda))$ is continuous on $\mfC$, which is not automatic since $V^\star(\Lambda)$ is only implicitly defined. This holds, e.g., if $0\notin\mfC$ and the constraint is interior for every $\Lambda\in\mfC$, by the implicit function theorem applied to $\nabla_V\Gamma(\Lambda,V)=\mbbE_{\mbfP^\star}S_\Lambda$. 
A complete analysis of this training problem is beyond the scope of the present work, and we plan to address it in future work.
\end{remark}

\begin{remark}[The free-energy statistic: tied values]\label{rem:free-energy}
By \Cref{prop:gibbs-attn}, the attention cross-moment is the gradient of the attention free energy $F_\Lambda(\mu)=\int\log Z_\mu\,\De\mu$, i.e.\ $S_\Lambda=\nabla_\Lambda F_\Lambda$. 
One may constrain the scalar $F_\Lambda$ instead of $S_\Lambda$. If $F_\Lambda$ is not constant on $\mcS_{\mbfN}$ and the constraint is interior, 
\Cref{thm:iproj} gives the minimizer $\mrmP^F_{\lambda^\star,\Lambda}\propto e^{\lambda^\star F_\Lambda}\mbfN$, with a scalar Lagrange multiplier $\lambda^\star$. Three things change with respect to the cross-moment model.
\begin{enumerate}[(i)]
\item \emph{Values are tied.} By \Cref{prop:gibbs-attn}-\ref{item:attention:EOT:a}, $\lambda F_\Lambda=\Psi_{\Lambda,\lambda\Lambda}-\lambda\mcH_\Lambda$: the plan is read with the value matrix $\lambda\Lambda$, whereas in the cross-moment model $V^\star$ is free (\Cref{thm:fixed}-\ref{item:fixed:c}).
\item \emph{There is no correction term.} A direct computation gives $\delta F_\Lambda(\mu)(z)=\log Z_\mu(z)+\int\frac{k_\Lambda(x,z)}{Z_\mu(x)}\,\mu(\De x)$, hence
\[
\nabla_z\delta F_\Lambda(\mu)(z)=\Lambda\Attn_\Lambda(z;\mu)+\Lambda^\top\!\!\int x\,\frac{k_\Lambda(x,z)}{Z_\mu(x)}\,\mu(\De x),
\]
which is \eqref{eq:force-psi} with $V=\Lambda$ and without $\rho_\mu$.
\item \emph{The value read from a key is the key.} For $\Lambda=Q^\top K$, the query term is $Q^\top\sum_j\mathrm{softmax}_j(\ip{Qy}{Kx_j})\,Kx_j$.
\end{enumerate}
\end{remark}

\section{The causal model: masking and next-token readout}\label{sec:shifted}

By \Cref{thm:last}, the score of the last token of the model $\mrmP_{\Lambda,V^\star}$ is $\frac1n P^\perp_y$ applied to the sum of three terms: the attention head $V^\star\Attn_\Lambda(y;\mu_y)$ of the candidate $y$, a key-side term and a correction $\rho_{\mu_y}$. 
These terms appear in the bidirectional model when differentiating the first variation \eqref{eq:first-variation-psi} of the readout, which consists of two parts, the payoff of $z$ as a query and its effect on the readouts of the other tokens:
\begin{equation*}
\delta\Psi_{\Lambda,V}(\mu)(z)=z^\top V\Attn_\Lambda(z;\mu)+\int \frac{k_\Lambda(x,z)}{Z_\mu(x)}\,x^\top V\big(z-\Attn_\Lambda(x;\mu)\big)\,\mu(\De x)\,.
\end{equation*}
The gradient of the first part is $V\Attn_\Lambda(z;\mu)$ plus the query part of $\rho_\mu$, which comes from the dependence on $z$ of the weights with which $z$ reads. The gradient of the second part is the key-side term plus the key part of $\rho_\mu$, which comes from the dependence on $z$ of the weights with which the other tokens read $z$. Hence:
\begin{enumerate}[(i)]
\item \label{item:causal:i} the key-side term and the key part of $\rho_\mu$ vanish if no query reads the candidate;
\item \label{item:causal:ii} the query part of $\rho_\mu$ vanishes if the candidate is not a query.
\end{enumerate}

A decoder transformer uses only an attention head, and in this section we explain why the other two terms are absent there.
More precisely, we are going to change only the statistic: the choice rule is the causally masked rule \eqref{eq:causal-rule-intro}, and the readout pairs what a token reads with the token that follows it. 
Indeed, causal masking gives property~\ref{item:causal:i} for the last token, except for the self-term in which the last token reads itself: no query comes after it. 
A decoder also satisfies property~\ref{item:causal:ii}: the logit of a candidate next token is the inner product of the candidate with the output of the head whose query is the last token read.
The candidate is compared with this output and never acts as a query. In particular the head term itself is evaluated at the query $x_{n-1}$, not at the candidate. The statistic below encodes this.

\medskip

The rest of the section is organized as follows. 
In \Cref{sec:shifted-causal} we set up ordered sentences and derive the causal choice rule, 
in \Cref{sec:shifted-def} we define the shifted cross-moment, solve the corresponding entropic projection, 
and show that the law of the last token is exactly a softmax over the sphere whose logit is one causal attention head, 
so that its score has no key-side term and no correction. This proves the Meta-Theorem for the masked model.

\subsection{Ordered sentences, the null model and the causal choice rule}\label{sec:shifted-causal}
As explained in \Cref{sub:intro:ordered} we are interested in keeping track of the positions of tokens in sentences. Since our previous model was set up 
in the space of laws of random measures $\mcP(\mcP(\mbbS^d))$, one way of keeping track of positions there is by lifting positions as well and considering ordered sentences as 
\[
\arrowmu=\iota(n,x_{1:n})\coloneqq\frac1n\sum_{i=1}^n\delta_{(x_i,i)}\in\mcP(\mbbS^d\times\N)\,\quad\text{ for }n\geq 1\,.
\]
The map $\iota$ is injective on $\mcO\coloneqq\bigsqcup_{n\ge1}(\mbbS^d)^n$, even when tokens are repeated, because the positions are recorded: $n$ is the number of atoms of $\arrowmu$ and $x_i$ is the token at position $i$.
It is continuous on each $(\mbbS^d)^n$.
The null model $\arrowN=\iota_\#\big(\sum_n\nu(n)\,\delta_n\otimes\uvol^{\otimes n}\big)$ is the law of $\arrowmu$ when $N\sim\nu$ and the tokens are read in their order of generation; it is concentrated on $\iota(\mcO)$, and so is every $\mrmP$ with $\rel{\mrmP}{\arrowN}<\infty$. 
The image of $\arrowmu$ under $(x,i)\mapsto x$ is the unordered sentence $\mu$, and the image of $\arrowN$ is then $\mbfN$.

\medskip

In this ordered setting, a choice rule for $\arrowmu$ is a coupling in the set $\Pi_1(\arrowmu)$ of probability measures on $(\mbbS^d\times\N)^2$ with first marginal $\arrowmu$, the query being written first. 
In a causally masked rule the query $(x_i,i)$ only chooses among the keys $(x_j,j)$ with $j\le i$. Then, the least informative choice rule we can consider here is the masked uniform choice 
\[
\mrmR^{\mathrm M}_{\arrowmu}(\De x\,\De i,\De y\,\De j)=\arrowmu(\De x\,\De i)\,\frac{\mathds1_{j\le i}\,\arrowmu(\De y\,\De j)}{\arrowmu(\mbbS^d\times\{1,\dots,i\})}=\sum_{i=1}^n\sum_{j=1}^n\,\frac1n\,\frac{e^{M_{ij}}}{\sum_{\ell=1}^ne^{M_{i\ell}}}\;\delta_{(x_i,i)}\otimes\delta_{(x_j,j)}\,,
\]
where $M\in\{0,-\infty\}^{n\times n}$ is the causal additive mask defined as $M_{ij}= (-\infty)\,\mathds1_{i<j}$, with the convention $e^{-\infty}=0$.
This is the masked rule without relational content: the query $(x_i,i)$ chooses its key uniformly among the $i$ tokens that are not later than itself. 
In the unordered model this corresponds to the independent coupling $\mu\otimes\mu$.

\begin{remark}[Masked Luce axiom and the lowest-order masked rule]\label{rem:masked-luce}
As in the Introduction, we postulate Luce's axiom among the competitors that are not later than the query, with a choice function $h\in \mcC(\mbbS^d\times\mbbS^d)$ that is the same for all sentences and does not depend on positions: 
the query $(x_i,i)$ chooses the key $(x_j,j)$ with probability $\mathds1_{j\le i}\,e^{h(x_i,x_j)}/\sum_{\ell\le i}e^{h(x_i,x_\ell)}$. This is the masked version of the ratio form \eqref{eq:score-rule}. 
\Cref{prop:harmonic} holds verbatim for masked rules. 
The only point to check is the converse in \ref{item:lowest:gauge}: if $h$ and $h'$ give the same masked rule on every ordered sentence, consider the ordered sentence $(y,x)$ of length $2$ with $y\ne x$; 
the query $(x,2)$ chooses between $y$ and $x$, and equality of the two rules gives $g(x,y)=g(x,x)$ for $g=h-h'$ and every $y\ne x$, hence $g(x,\cdot)$ is constant for every $x$. 

Truncating the relational part of $h$ at its lowest order and setting the key bias to zero gives $h(x,y)=x^\top\Lambda y$, 
that is, the causally masked rule $\pi^{\Lambda,\mathrm M}_{\arrowmu}$ of \eqref{eq:causal-rule-intro}
\[\pi^{\Lambda,\mathrm M}_{\arrowmu}=\sum_{i=1}^n\sum_{j=1}^n\,\frac1n\,\frac{e^{\ip{x_i}{\Lambda x_j}+M_{ij}}}{\sum_{\ell=1}^ne^{\ip{x_i}{\Lambda x_\ell}+M_{i\ell}}}\;\delta_{(x_i,i)}\otimes\delta_{(x_j,j)}\,.\]
\end{remark}

In coordinates, the query $(x_i,i)$ of $\pi^{\Lambda,\mathrm M}_{\arrowmu}$ chooses the key $(x_j,j)$ with probability $n\Pi^{\mathrm M}_{ij}$, and reads
\begin{equation}\label{eq:shifted-weights}
n\Pi^{\mathrm M}_{ij}=\frac{\mathds1_{j\le i}\,e^{x_i^\top\Lambda x_j}}{\sum_{\ell\le i}e^{x_i^\top\Lambda x_\ell}}=\frac{e^{\ip{x_i}{\Lambda x_j}+M_{ij}}}{\sum_{\ell=1}^ne^{\ip{x_i}{\Lambda x_\ell}+M_{i\ell}}},\qquad\Attn_\Lambda((x_i,i);\arrowmu)=\sum_{j\le i}n\Pi^{\mathrm M}_{ij}\,x_j .
\end{equation}
Note that $n\Pi^{\mathrm M}_{ij}$ and $\Attn_\Lambda((x_i,i);\arrowmu)$ depend on the sentence only through $x_{1:i}$, and not on its length $n$.

\begin{proposition}[The causal choice rule]\label{prop:causal-rule}
Let $\arrowmu=\iota(n,x_{1:n})$ and $C^{\mathrm M}=\int xy^\top\De\pi^{\Lambda,\mathrm M}_{\arrowmu}$.
\begin{enumerate}[(a)]
\item \label{item:causal-rule:a} Every $\pi\in\Pi_1(\arrowmu)$ with $\rel{\pi}{\mrmR^{\mathrm M}_{\arrowmu}}<\infty$ gives zero mass to $\{j>i\}$.
\item \label{item:causal-rule:b} $\pi^{\Lambda,\mathrm M}_{\arrowmu}$ is the unique minimizer of $\rel{\pi}{\mrmR^{\mathrm M}_{\arrowmu}}$ over $\{\pi\in\Pi_1(\arrowmu):\int xy^\top\De\pi=C^{\mathrm M}\}$.
\item \label{item:causal-rule:c} The query $(x_i,i)$ chooses the key $(x_j,j)$ with probability $n\Pi^{\mathrm M}_{ij}$ of \eqref{eq:shifted-weights}, and $C^{\mathrm M}=\frac1n\sum_ix_i\AttM{i}^\top$.
\end{enumerate}
\end{proposition}

\begin{proof}
Write $q=(x,i)$ for a query, $\pi(\De q,\De k)=\arrowmu(\De q)\,\pi_q(\De k)$, and $\mrmR_q$ for the conditional law of the key under $\mrmR^{\mathrm M}_{\arrowmu}$. 
Since $\pi$ and $\mrmR^{\mathrm M}_{\arrowmu}$ have the same first marginal, the chain rule \cite[Lemma 1.4.3]{DupuisEllis} gives $\rel{\pi}{\mrmR^{\mathrm M}_{\arrowmu}}=\int\arrowmu(\De q)\,\rel{\pi_q}{\mrmR_q}$. 
If this is finite, then $\pi_q\ll\mrmR_q$ for $\arrowmu$-almost every $q$. In particular, since $\mrmR_q$ is carried by $\{j\le i\}$, the same must hold for $\pi_q$ $\arrowmu$-almost every $q$, and therefore $\pi$ must give zero mass to $\{j>i\}$. This shows \ref{item:causal-rule:a}.

Item~\ref{item:causal-rule:b} can be proven as we did for the unordered setting in~\Cref{prop:condmaxent}. Indeed, for any $\pi\in\Pi_1(\arrowmu)$ with $\rel{\pi}{\mrmR^{\mathrm M}_{\arrowmu}}<\infty$ we have
\begin{equation*}
\begin{aligned}
\rel{\pi}{\mrmR^{\mathrm M}_{\arrowmu}}=\rel{\pi}{\pi^{\Lambda,\mathrm M}_{\arrowmu}}+\Big\langle\Lambda,\int xy^\top\De\pi\Big\rangle-\int\log\mbbE_{\mrmR_q}\big[e^{x^\top\Lambda y}\big]\,\arrowmu(\De q).
\end{aligned}
\end{equation*}
On the constraint set the last two terms do not depend on $\pi$, and $\rel{\pi}{\pi^{\Lambda,\mathrm M}_{\arrowmu}}=0$ if and only if $\pi=\pi^{\Lambda,\mathrm M}_{\arrowmu}$. 

Claim~\ref{item:causal-rule:c} is a straightforward computation.
\end{proof}

The masked entries of the coupling are therefore not set to zero by hand in the rule: the mask is placed in the reference coupling $\mrmR^{\mathrm M}_{\arrowmu}$, and the finiteness of the relative entropy imposes their presence in the choice rule as well. 
Therefore the masking in the choice rule \eqref{eq:causal-rule-intro} is not imposed on the rule by hand, but is rather a consequence of the masked reference rule $\mrmR^{\mathrm M}_{\arrowmu}$.
As in the unordered case, $\Lambda$ plays the role of the multiplier of the constraint, the constraint level $C^{\mathrm M}$ being computed from $\Lambda$ itself. In the rest of the section we only use two facts:
\begin{equation}\label{eq:shifted-facts}
\AttM{i}\ \text{depends only on}\ x_{1:i},\qquad\abs{\AttM{i}}\le1\,.
\end{equation}
In particular $\AttM{i}$ does not depend on the length $n$ of the sentence, nor on the tokens after position~$i$.

\subsection{The shifted cross-moment and the next-token model}\label{sec:shifted-def}
In the previous section we have shown how the masking takes care of property~\ref{item:causal:i}. In this section we show how a small modification of the cross-moment statistic yields property~\ref{item:causal:ii}.

\begin{lemma}[Shifted cross-moment]\label{lem:shifted-forms}
Let $n\ge2$, $\arrowmu=\iota(n,x_{1:n})$, $X=(x_1,\dots,x_n)\in\R^{(d+1)\times n}$, $\Pi^{\mathrm M}=(\Pi^{\mathrm M}_{ij})_{i,j\le n}$ as in \eqref{eq:causal-rule-intro}, and $J\in\mbbM_n$ the lower shift matrix, $J_{ij}=\mathds1_{\{i=j+1\}}$. Let $\arrowmu_{1:n-1}=\frac1{n-1}\sum_{i<n}\delta_{(x_i,i)}$ be the context and $s(x,i)=(x_{i+1},i+1)$, $i<n$, the successor map on its atoms. Then
\begin{equation}\label{eq:shifted-coupling}
S^\to_\Lambda(\arrowmu)\coloneqq\frac1{n-1}\sum_{i=1}^{n-1}x_{i+1}\AttM{i}^\top=\frac n{n-1}\,XJ\Pi^{\mathrm M}X^\top=\int xy^\top\,\De\pi^{\Lambda,\to}_{\arrowmu},
\end{equation}
where $\pi^{\Lambda,\to}_{\arrowmu}=(\iota_{\arrowmu}\times\iota_{\arrowmu})_\#\big(\tfrac n{n-1}J\Pi^{\mathrm M}\big)=(s\times\mathrm{id})_\#\,\pi^{\Lambda,\mathrm M}_{\arrowmu_{1:n-1}}$, with $\iota_{\arrowmu}(i)=(x_i,i)$, is a probability measure on $(\mbbS^d\times\N)^2$. 
Its first marginal is the uniform law on $\{(x_i,i):\ 2\le i\le n\}$, and, given that the first coordinate is $(x_{i+1},i+1)$, the second one is $(x_j,j)$ with probability $n\Pi^{\mathrm M}_{ij}$, $j\le i$: the key is chosen by the masked choice rule of the preceding token $(x_i,i)$, and paired with its successor $x_{i+1}$.
\end{lemma}

\begin{proof}
$(J\Pi^{\mathrm M})_{ij}=\Pi^{\mathrm M}_{i-1,j}$ for $i\ge2$, and the first row of $J\Pi^{\mathrm M}$ vanishes. Hence $XJ\Pi^{\mathrm M}X^\top=\sum_{i=2}^n\sum_j\Pi^{\mathrm M}_{i-1,j}\,x_ix_j^\top=\frac1n\sum_{i=1}^{n-1}x_{i+1}\AttM{i}^\top$, 
which gives the second equality. Each row of $\Pi^{\mathrm M}$ sums to $\frac1n$, so $\frac n{n-1}J\Pi^{\mathrm M}$ is a non-negative matrix with total mass $1$, 
whose row $i+1$ is the masked choice $(n\Pi^{\mathrm M}_{ij})_{j=1,\dots,n}$ of the query $i$, multiplied by $\frac1{n-1}$.  Its pushforward under $\iota_{\arrowmu}\times\iota_{\arrowmu}$ is therefore a probability measure with the stated marginal and conditional law,
 and $\int xy^\top\De\pi^{\Lambda,\to}_{\arrowmu}=\frac n{n-1}\sum_{i,j}(J\Pi^{\mathrm M})_{ij}x_ix_j^\top$, which is the third equality.

 Finally, consider the context $\arrowmu_{1:n-1}=\frac1{n-1}\sum_{i<n}\delta_{(x_i,i)}$. A query $(x_i,i)$ with $i\le n-1$ only chooses among the keys at positions $j\le i$, which all belong to the context, so its masked choice probabilities are the same in $\arrowmu$ and in $\arrowmu_{1:n-1}$, namely $n\Pi^{\mathrm M}_{ij}$. 
 Hence $\pi^{\Lambda,\mathrm M}_{\arrowmu_{1:n-1}}$ gives mass $\frac{n\Pi^{\mathrm M}_{ij}}{n-1}$ to $\{((x_i,i),(x_j,j))\,\colon\,j\leq i<n\}$. The map $(s\times\mathrm{id})$ replaces each query by its successor and leaves the key unchanged, so 
 the pushforward under $(s\times \id)_\#\pi^{\Lambda,\mathrm M}_{\arrowmu_{1:n-1}}$ gives mass $\frac{n\Pi^{\mathrm M}_{ij}}{n-1}=\frac n{n-1}(J\Pi^{\mathrm M})_{i+1,j}$ to $((x_{i+1},i+1),(x_j,j))$, which is the mass given by $(\iota_{\arrowmu}\times\iota_{\arrowmu})_\#\big(\frac n{n-1}J\Pi^{\mathrm M}\big)$. 

\end{proof}

In our model we will therefore consider the shifted cross-moment statistic. This leads to the next-token attention readout.

\begin{definition}[Next-token attention readout]\label{def:shifted}
For $\Lambda,V\in\Mat$ and an ordered sentence $\arrowmu=\iota(n,x_{1:n})$ let $\Psi^\to_{\Lambda,V}(\arrowmu)=\ip{V}{S^\to_\Lambda(\arrowmu)}=\frac1{n-1}\sum_{i=1}^{n-1}x_{i+1}^\top V\AttM{i}$ if $n\ge2$, and $S^\to_\Lambda(\arrowmu)=0$, $\Psi^\to_{\Lambda,V}(\arrowmu)=0$ if $n=1$. The \emph{next-token model} is
\[
\mrmP^{\to}_{\Lambda,V}=\frac{e^{\Psi^\to_{\Lambda,V}}\,\arrowN}{\mbbE_{\arrowN}e^{\Psi^\to_{\Lambda,V}}},\qquad\text{that is,}\qquad\mrmP^{\to}_{\Lambda,V}\big(N=n,\ X_{1:n}\in\De x_{1:n}\big)=\frac{\nu(n)\,e^{\Psi^\to_{\Lambda,V}(x_{1:n})}\,\uvol^{\otimes n}(\De x_{1:n})}{\sum_{n'}\nu(n')\int e^{\Psi^\to_{\Lambda,V}}\,\De\uvol^{\otimes n'}} .
\]
\end{definition}
\noindent
We sometimes write $\Psi^\to$ for $\Psi^\to_{\Lambda,V}$. By \eqref{eq:shifted-facts}, $\abs{x_{i+1}^\top V\AttM{i}}\leq \norm V_\op$, $\abs{\Psi^\to_{\Lambda,V}}\leq \norm V_\op$ and $\norm{S^\to_\Lambda}_\op\le1$.

The masked choice rule is unchanged and the decoder property~\ref{item:causal:ii} changes only the readout, which differs from the causal query--key cross-moment $C^{\mathrm M}=\frac1n\sum_ix_i\AttM{i}^\top$ of \Cref{prop:causal-rule}-\ref{item:causal-rule:c}:
what the query $x_i$ reads is no longer paired with $x_i$ itself but with the next token $x_{i+1}$. The normalization $\frac1{n-1}$ makes $S^\to_\Lambda$ the cross-moment of the probability coupling $\pi^{\Lambda,\to}_{\arrowmu}$ (\Cref{lem:shifted-forms}): the choice rule is run on the context, and each query is replaced by its successor. 

\medskip

We are now ready to prove Item~\ref{item:meta:ord:b} from the Meta-Theorem in the ordered setting.

\begin{theorem}[Entropic minimization of language laws with shifted cross-moment statistics]\label{thm:shifted-iproj}
Let $\Lambda\in\Mat$ be arbitrary, including $\Lambda=0$.
\begin{enumerate}[(a)]
\item \label{item:shifted:thm:a} (Non-degeneracy) Let $n\ge2$. If $\ip{V}{S^\to_\Lambda}$ is constant on $(\mbbS^d)^n$ for some $V\in\Mat$, then $V=0$.
\item \label{item:shifted:thm:b} Assume that $\nu(n)>0$ for some $n\ge2$, and let $\mcK^\to_\Lambda$ be the closed convex hull of $S^\to_\Lambda(\supp\arrowN)$. Then $V\mapsto\mcM^\to_\Lambda(V)\coloneqq\mbbE_{\mrmP^\to_{\Lambda,V}}S^\to_\Lambda$ is a real-analytic diffeomorphism of $\Mat$ onto $\mathrm{int}\,\mcK^\to_\Lambda$. If $\mbbE_{\mbfP^\star}S^\to_\Lambda\in\mathrm{int}\,\mcK^\to_\Lambda$, the problem
\[
\min\big\{\rel{\mrmP}{\arrowN}:\ \mrmP\in\mcP(\mcP(\mbbS^d\times\N)),\ \mbbE_\mrmP S^\to_\Lambda=\mbbE_{\mbfP^\star}S^\to_\Lambda\big\}
\]
has the unique solution $\mrmP^\to_{\Lambda,V^\star}$, with $V^\star=V^\star(\Lambda)=(\mcM^\to_\Lambda)^{-1}(\mbbE_{\mbfP^\star}S^\to_\Lambda)$, and the Pythagorean identity 
\eqref{eq:pythagoras} holds with $\arrowN$, $S^\to_\Lambda$ and $\mrmP^\to_{\Lambda,V^\star}$ in place of $\mbfN$, $\Phi$ and $\mrmP_{\eta_c}$.
\item \label{item:shifted:thm:c} (Every value matrix is realized) Under the assumption of~\ref{item:shifted:thm:b}, for every $V\in\Mat$ the language $\mbfP^\star=\mrmP^\to_{\Lambda,V}$ has value matrix $V^\star(\Lambda)=V$.
\end{enumerate}
\end{theorem}

\begin{proof}
In $\ip{V}{S^\to_\Lambda}=\frac1{n-1}\sum_{i<n}x_{i+1}^\top V\AttM{i}$, the token $x_n$ appears only in the term $i=n-1$, because $\AttM{i}$ depends on $x_{1:i}$ only. 
If we assume it to be constant and evaluate it on $(x_1,\dots,x_n)$ and on $(x_1,\dots,-x_n)$ we immediately deduce $x_{n}^\top V\AttM{n-1}=0$ for any $x_n\in\mbbS^d$. Therefore $V\AttM{n-1}=0$ for any $(x_1,\dots,x_{n-1})$.
In the constant configuration $x_1=\dots=x_{n-1}=x$ every key equals $x$, so $\AttM{n-1}=x$, and we get $Vx=0$ for every $x\in\mbbS^d$, so $V=0$. This proves~\ref{item:shifted:thm:a}.

\medskip

Claim~\ref{item:shifted:thm:b} can be proven as \Cref{thm:iproj}, 
whose proof only uses boundedness and measurability of the statistic, the
fact that it is not a.s.\ constant in any direction under the null law, and the identification of the
convex support of its image law. In particular it does not rely on the compactness of $\mcP(\mcP(\mbbS^d))$.
Since $\iota$ is an injective Borel map on the Polish space $\mcO$, it is a Borel isomorphism onto its (Borel) image by the Lusin--Souslin theorem, and it preserves relative entropies. 
We can therefore work indifferently with laws of $(N,X_{1:N})$ on $\mcO=\bigsqcup_{n\ge1}(\mbbS^d)^n$ here, instead of with laws on $\mcP(\mbbS^d\times\N)$ concentrated on $\iota(\mcO)$.
Notice that under $\arrowN$, the law of $(N,X_{1:N})$ is $\sum_n\nu(n)\,\delta_n\otimes\uvol^{\otimes n}$. We show the applicability of \Cref{thm:iproj} in this setting.
\begin{enumerate}
  \item Boundedness of the shifted cross-moment $S^\to_\Lambda$ is trivial since it takes values in the compact set $\{S\in\Mat:\ \norm S_\Fr\le1\}$. 
  \item Non-degeneracy follows from~\ref{item:shifted:thm:a}. Indeed, if we suppose that  $\ip{V}{S^\to_\Lambda}$ is $\arrowN$-almost surely constant then for any fixed $n\ge2$ with $\nu(n)>0$, $\ip{V}{S^\to_\Lambda}$ is $\uvol^{\otimes n}$-almost surely constant on $(\mbbS^d)^n$ and by continuity constant everywhere in $(\mbbS^d)^n$. By~\ref{item:shifted:thm:a} we would then deduce $V=0$.
  \item We need to show that $\overline{\mathrm{conv}}(\supp((S^\to_\Lambda)_\#\arrowN))=\overline{\mathrm{conv}}\,\big(\bigcup_{n\in\supp(\nu)}S^\to_\Lambda\big((\mbbS^d)^n\big)\big)=\mcK^\to_\Lambda$. The image law of $S^\to_\Lambda$ under the prior $\arrowN$ is the mixture $\sum_n\nu(n)\,(S^\to_\Lambda)_\#\uvol^{\otimes n}$. 
  Since $(S^\to_\Lambda)_\#\uvol^{\otimes n}$ has support $S^\to_\Lambda((\mbbS^d)^n)$, the support of the mixture is the closure $\overline{\cup_{n\in\supp(\nu)}S^\to_\Lambda((\mbbS^d)^n)}$ of the union of these supports, and its closed convex hull is exactly $\mcK^\to_\Lambda$.
\end{enumerate}
Hence the proof of \Cref{thm:iproj} applies verbatim, which proves~\ref{item:shifted:thm:b}.

\medskip

Finally, Item~\ref{item:shifted:thm:c} can be proven as in \Cref{thm:fixed}-\ref{item:fixed:c}: for $\mbfP^\star=\mrmP^\to_{\Lambda,V}$ the constraint $\mcM^\to_\Lambda(V)$ lies in $\mathrm{int}\,\mcK^\to_\Lambda$, and $V^\star(\Lambda)=(\mcM^\to_\Lambda)^{-1}(\mcM^\to_\Lambda(V))=V$.

\end{proof}

As in \Cref{rem:interior}, the interior assumption in~\ref{item:shifted:thm:b} holds for the empirical shifted cross-moment $\widehat{S^\to_\Lambda}=\frac1J\sum_{j=1}^JS^\to_\Lambda(\arrowmu_j)$ of a corpus as soon as the centered matrices $S^\to_\Lambda(\arrowmu_j)-\widehat{S^\to_\Lambda}$, $j\le J$, span $\Mat$.

\bigskip

We can now conclude this section with the proof of Item~\ref{item:meta:ord:c} from the Meta-Theorem in the ordered setting. 
The following result applies to any Gibbs tilt $\mrmP^\to_{\Lambda,V}$. For any fixed $\Lambda$, this in particular applies to $V^\star=V^\star(\Lambda)$ as in \Cref{thm:shifted-iproj}, and hence to our optimizer $\mrmP^\to_{\Lambda,V^\star}$ for \eqref{eq:meta:min:ord}, which corresponds to Item~\ref{item:meta:ord:c} of the Meta-Theorem for the ordered model.

\begin{theorem}[Last token: one decoder head, exactly]\label{thm:shifted-last}
Let $n\ge2$ with $\nu(n)>0$, $\Lambda,V\in\Mat$, $x_{1:n-1}\in(\mbbS^d)^{n-1}$, and let $\AttM{n-1}$ 
be the causal attention output \eqref{eq:shifted-weights} of the last context token, which is a function of $x_{1:n-1}$ only.
\begin{enumerate}[(a)]
\item \label{item:thm:shifted:last:a} (Exact law) Under $\mrmP^{\to}_{\Lambda,V}$, given $N=n$ and $X_{1:n-1}=x_{1:n-1}$, the last token has density with respect to $\uvol$
\begin{equation}\label{eq:shifted-last}
p(y)=\frac{\exp\big(\frac1{n-1}\,y^\top V\AttM{n-1}\big)}{\int\exp\big(\frac1{n-1}\,y'^\top V\AttM{n-1}\big)\,\uvol(\De y')},\qquad y\in\mbbS^d .
\end{equation}
If $V\AttM{n-1}\ne0$, this is the von Mises--Fisher law with mean direction $V\AttM{n-1}$ (normalized) and concentration $\abs{V\AttM{n-1}}/(n-1)$.
\item \label{item:thm:shifted:last:b} (Score) The Riemannian gradient of $\log p$ on $\mbbS^d$ is
\begin{equation}\label{eq:shifted-score}
\nabla_{\mbbS^d}\log p(y)=\frac1{n-1}\,P^\perp_y\,V\AttM{n-1}=\frac1{n-1}\,P^\perp_y\sum_{j=1}^{n-1}\frac{e^{x_{n-1}^\top\Lambda x_j+M_{n-1,j}}}{\sum_{\ell=1}^{n-1}e^{x_{n-1}^\top\Lambda x_\ell+ M_{n-1,\ell}}}\,Vx_j\,,
\end{equation}
with the mask $M\in\{0,-\infty\}^{n\times n}$ defined as $M_{ij}= (-\infty)\,\mathds1_{i<j}$, with the convention $e^{-\infty}=0$.
\item \label{item:thm:shifted:last:c} (Transformer reading) For $\Lambda=Q^\top K$, the logit $y^\top V\AttM{n-1}$ in \eqref{eq:shifted-last} is the inner product of the candidate $y$ with the output, at position $n-1$ (the last position of the context), of the causally masked version of the head \eqref{eq:intro:transformer}, 
with query matrix $Q$, key matrix $K$ and value matrix $V$.
\end{enumerate}
\end{theorem}

\begin{proof}
 By \Cref{def:shifted}, given $N=n$ the ordered tokens have density proportional to $e^{\Psi^\to_{\Lambda,V}}$ with respect to $\uvol^{\otimes n}$, and $\nu(n)>0$. Conditioning on $X_{1:n-1}=x_{1:n-1}$ (Bayes' formula, as in \Cref{prop:last-token}-\ref{item:last:gen:b} ) gives the density proportional to $y\mapsto e^{\Psi^\to_{\Lambda,V}(x_{1:n-1},y)}$. In \Cref{def:shifted} with $x_n=y$, the terms $i\le n-2$ involve $x_{i+1}$ with $i+1\le n-1$ and $\AttM{i}$ with $i\le n-2$, which depend only on $x_{1:n-1}$ by \eqref{eq:shifted-facts}; they cancel in the normalization. 
 The remaining term $i=n-1$ is $\frac1{n-1}y^\top V\AttM{n-1}$, and $\AttM{n-1}$ depends only on $x_{1:n-1}$. The last token appears neither among the keys of a query, since every key of $\AttM{i}$ has index at most $i\le n-1$, nor as a query, since the queries are $x_1,\dots,x_{n-1}$. 
 This proves~\ref{item:thm:shifted:last:a}.

 Claims~\ref{item:thm:shifted:last:b} and \ref{item:thm:shifted:last:c} are straightforward computations.
\end{proof}

Notice that the readout depends on the candidate $y$ only through the linear term $\frac1{n-1}y^\top V\AttM{n-1}$, so the score is the tangential part of a vector that does not depend on $y$.
The candidate is neither a key nor a query of the statistic, and both mechanisms that produce the key-side term and the correction term are absent. Together with \Cref{thm:shifted-iproj}, this proves Items \ref{item:meta:ord:b} and \ref{item:meta:ord:c} of the Meta-Theorem for the masked model. Item \ref{item:meta:ord:a} is \Cref{prop:causal-rule}, given the masked Luce postulate and the lowest-order truncation of \Cref{rem:masked-luce}.

\begin{remark}[Last token versus autoregressive next token]\label{rem:last-vs-next}
\Cref{thm:shifted-last} gives the law of $X_n$ given $X_{1:n-1}$ \emph{and} $N=n$, i.e.\ the law of
the last token of a sentence whose length is known. It does not give the law of an
intermediate token given its past. For $1\le i\le n-2$ with $\nu(n)>0$, integrating out $X_{i
+2:n}$ in \Cref{def:shifted} gives
\[
\mrmP^\to_{\Lambda,V}\big(X_{i+1}\in\De y\,\big|\,N=n,\ X_{1:i}=x_{1:i}\big)\;\propto\;\exp\Big(\tfrac1{n-1}\,y^\top V\AttM{i}\Big)\,G_{i+1}(x_{1:i},y)\,\uvol(\De y),
\]
\[
G_{i+1}(x_{1:i},y)=\int\exp\Big(\frac1{n-1}\sum_{k=i+1}^{n-1}x_{k+1}^\top V\AttM{k}\Big)\,\uvol^{\otimes(n-i-1)}(\De x_{i+2:n})\Big|_{x_{i+1}=y},
\]
and $G_{i+1}$ depends on $y$, which is the query at position $i+1$ and a key of all later queries.
In particular the conditional law of intermediate tokens is not a von Mises--Fisher law with attention parameter:
a single attention head governs the law of the final token only; we plan to address the extension to intermediate tokens in future work.
\end{remark}

\begin{remark}[No self-pairing]\label{rem:shifted-selfpair}
Let $\Lambda=\kappa\Id$ with $\kappa\to\infty$, and consider an ordered sentence with pairwise distinct tokens, which is the case $\arrowN$-almost surely. The masked weights $n\Pi^{\mathrm M}_{ij}\propto e^{\kappa\,x_i^\top x_j}$, $j\le i$, concentrate on the largest exponent, which is attained only at $j=i$ since $x_i^\top x_i=1>x_i^\top x_j$ for $j\ne i$. Hence each token reads itself in the limit, $\Attn_{\kappa\Id}((x_i,i);\arrowmu)\to x_i$, and
\[
S^\to_{\kappa\Id}(\arrowmu)\to\frac1{n-1}\sum_{i<n}x_{i+1}x_i^\top=\frac1{n-1}XJX^\top,\qquad \Psi^\to_{\kappa\Id,V}(\arrowmu)\to\frac1{n-1}\sum_{i<n}x_{i+1}^\top Vx_i ,
\]
the bigram moment and the bigram readout. The limit readout is not constant, and it has mean $0$ under $\arrowN$ for every $V$, because $x_{i+1}$ is centered and independent of $x_i$, so the limit model $\propto e^{\frac1{n-1}\sum_ix_{i+1}^\top Vx_i}\arrowN$ is a non-trivial bigram model. 
By contrast, in the bidirectional model each read is paired with the reading token itself, so the limit model is $\propto e^{\frac1{n}\sum_ix_{i}^\top Vx_i}\mbfN$ which is a tilt of the null model without interaction between tokens.
This is the precise sense in which the shifted readout ``removes self-reading'': every token is still a key of itself in the masked choice rule, but no token is paired with its own read.
\end{remark}

In \Cref{thm:shifted-last} the concentration of the last-token law is $\abs{V^\star\AttM{n-1}}/(n-1)$, which vanishes for long sentences since a single matrix $V^\star$ serves all lengths. This is due to the normalization $\frac1{n-1}$ of the statistic. One may instead constrain the extensive statistic 
\[
T^\to_\Lambda(\arrowmu)\coloneqq(n-1)\,S^\to_\Lambda(\arrowmu)=\sum_{i=1}^{n-1}x_{i+1}\AttM{i}^\top.
\]
For this modified statistic we are able to show the following result.

\begin{proposition}[Extensive next-token statistic]\label{prop:rescaled}
Assume that $\nu(n)>0$ for some $n\ge2$ and that $\sum_n\nu(n)e^{tn}<\infty$ for every $t>0$ (e.g.\ a Poisson law shifted by one). For $\arrowmu=\iota(n,x_{1:n})$ let $\mrmP^{\Rightarrow}_{\Lambda,V}\propto e^{\ip{V}{T^\to_\Lambda}}\arrowN$.
\begin{enumerate}[(a)]
\item \label{prop:rescaled:a} $\Gamma^\Rightarrow(V)=\log\mbbE_{\arrowN}e^{\ip{V}{T^\to_\Lambda}}$ is finite and real-analytic on $\Mat$, and $V\mapsto\mbbE_{\mrmP^\Rightarrow_{\Lambda,V}}T^\to_\Lambda$ is a
diffeomorphism of $\Mat$ onto the interior of the closed convex hull $\mcK^\Rightarrow_\Lambda$
of the support of $(T^\to_\Lambda)_\#\arrowN$.
\item \label{prop:rescaled:b} For $c\in\inter\mcK^\Rightarrow_\Lambda$, $\mrmP^\Rightarrow_{\Lambda,V^\star}$
is the unique minimizer of $\rel{\mrmP}{\arrowN}$ under $\mbbE_\mrmP T^\to_\Lambda=c$, where $V^\star$ is the unique matrix such that $\mbbE_{\mrmP^\Rightarrow_{\Lambda,V^\star}}T^\to_\Lambda=c$.
\item \label{prop:rescaled:c} Under $\mrmP^\Rightarrow_{\Lambda,V}$, given $N=n\ge2$ with $\nu(n)>0$ and $X_{1:n-1}=x_{1:n-1}$, the last token has density 
\[p(y)\propto\exp\big(y^\top V\AttM{n-1}\big)\]
with respect to $\uvol$, and score $P^\perp_y\,V\AttM{n-1}$.
\end{enumerate}
\end{proposition}
\begin{proof}
We write $T=T^\to_\Lambda$ and $\Gamma=\Gamma^\Rightarrow$ for brevity, and proceed in four steps.

\smallskip
\noindent\emph{Step 1: the log-partition function is finite and real-analytic.}
On $\{N=n\}$ the matrix $T$ is a sum of $n-1$ matrices $x_{i+1}\AttM{i}^\top$ of Frobenius norm at most $1$, hence $\norm{T}_\Fr\le n-1$. Therefore, for every complex matrix $\eta$,
\[
\mbbE_{\arrowN}\abs{e^{\ip{\eta}{T}}}\le\sum_{n\ge1}\nu(n)\,e^{(n-1)\norm{\eta}_\Fr}<\infty
\]
by the assumption on $\nu$. As in the proof of \Cref{thm:iproj}, this domination allows us to differentiate under the integral sign: $V\mapsto\mbbE_{\arrowN}e^{\ip{V}{T}}$ extends to an entire function, $\Gamma$ is real-analytic on $\Mat$, and
\[
\nabla\Gamma(V)=\mbbE_{\mrmP^\Rightarrow_{\Lambda,V}}T,\qquad\nabla^2\Gamma(V)=\Cov_{\mrmP^\Rightarrow_{\Lambda,V}}(T).
\]

\smallskip
\noindent\emph{Step 2: non-degeneracy.}
We show that $\nabla^2\Gamma(V)$ is positive definite. Assume that $\ip{A}{T}$ is $\arrowN$-a.s.\ constant for some $A\in\Mat$, and fix $n\ge2$ with $\nu(n)>0$. On $\{N=n\}$ we have $\ip{A}{T}=(n-1)\ip{A}{S^\to_\Lambda}$, so $\ip{A}{S^\to_\Lambda}$ is $\uvol^{\otimes n}$-a.s.\ constant on $(\mbbS^d)^n$. Being continuous, it is constant everywhere on $(\mbbS^d)^n$, and \Cref{thm:shifted-iproj}-\ref{item:shifted:thm:a} gives $A=0$. Hence $\Cov_{\mrmP^\Rightarrow_{\Lambda,V}}(T)$ is positive definite and $\Gamma$ is strictly convex.

\smallskip
\noindent\emph{Step 3: proof of~\ref{prop:rescaled:a}.}
By Step~1 the natural parameter space of the exponential family $\{\mrmP^\Rightarrow_{\Lambda,V}\}_{V\in\Mat}$ is the whole of $\Mat$, which is open: the family is regular, hence steep, and it is minimal by Step~2. By \cite[Theorem~3.6]{brown1986}, the mean map $\nabla\Gamma$ is a bijection of $\Mat$ onto $\inter\mcK^\Rightarrow_\Lambda$; it is a diffeomorphism by the inverse function theorem, since its Jacobian $\nabla^2\Gamma$ is invertible.

\smallskip
\noindent\emph{Step 4: proof of~\ref{prop:rescaled:b}.}
Let $c\in\inter\mcK^\Rightarrow_\Lambda$ and let $V^\star$ be the unique matrix with $\nabla\Gamma(V^\star)=c$, given by Step~3. Then $\mrmP^\star\coloneqq\mrmP^\Rightarrow_{\Lambda,V^\star}$ satisfies the constraint and $\rel{\mrmP^\star}{\arrowN}=\ip{V^\star}{c}-\Gamma(V^\star)<\infty$.
Now let $\mrmP$ be admissible with $\rel{\mrmP}{\arrowN}<\infty$ (otherwise there is nothing to prove). By the entropy inequality $\mbbE_\mrmP f\le\rel{\mrmP}{\arrowN}+\log\mbbE_{\arrowN}e^{f}$, applied to $f=\norm{T}_\Fr$, and by Step~1,
\[
\mbbE_\mrmP\norm{T}_\Fr\le\rel{\mrmP}{\arrowN}+\log\sum_{n\ge1}\nu(n)e^{n-1}<\infty .
\]
Hence $\log\frac{\De\mrmP^\star}{\De\arrowN}=\ip{V^\star}{T}-\Gamma(V^\star)$ is $\mrmP$-integrable, with integral $\ip{V^\star}{c}-\Gamma(V^\star)$. Since $\mrmP\ll\arrowN$ and $\arrowN$ is equivalent to $\mrmP^\star$, we have $\log\frac{\De\mrmP}{\De\arrowN}=\log\frac{\De\mrmP}{\De\mrmP^\star}+\log\frac{\De\mrmP^\star}{\De\arrowN}$, and integrating against $\mrmP$ gives the Pythagorean identity
\[
\rel{\mrmP}{\arrowN}=\rel{\mrmP}{\mrmP^\star}+\ip{V^\star}{c}-\Gamma(V^\star)=\rel{\mrmP}{\mrmP^\star}+\rel{\mrmP^\star}{\arrowN}.
\]
Thus $\rel{\mrmP}{\arrowN}\ge\rel{\mrmP^\star}{\arrowN}$, with equality if and only if $\mrmP=\mrmP^\star$.

\smallskip
\noindent\emph{Proof of~\ref{prop:rescaled:c}.}
As in \Cref{thm:shifted-last}, given $N=n$ and $X_{1:n-1}=x_{1:n-1}$, the density of the last token is proportional to $y\mapsto e^{\ip{V}{T(\iota(n,(x_{1:n-1},y)))}}$. By \eqref{eq:shifted-facts}, the terms $i\le n-2$ of $T$ do not depend on $x_n=y$, and they cancel in the normalization. The remaining term $i=n-1$ equals $y^\top V\AttM{n-1}$.
\end{proof}

\appendix

\section{Non-degeneracy of attention readouts}\label{app:non-deg}

In this appendix we show the non-degeneracy of attention readouts, a property needed for applying \Cref{thm:iproj} to our $\KL$ minimization problem in $\mcP(\mcP(\mbbS^d))$.
We start with a technical lemma that computes how the attention cross-moment and readout behave on two-point sentences.

\begin{lemma}\label{lem:twopoint}
Let $\Lambda,V\in\Mat$, fix $x,y\in\mbbS^d$ and consider $\mu_t=(1-t)\delta_x+t\delta_y$, $t\in[0,1]$. Then
\begin{equation}\label{eq:twopoint-exact}
\Psi_{\Lambda,V}(\mu_t)=(1-t)\,x^\top Vx+t\,y^\top Vy+t(1-t)\Big[\frac{e^{x^\top\Lambda y}}{Z_x(t)}\,x^\top V(y-x)+\frac{e^{y^\top\Lambda x}}{Z_y(t)}\,y^\top V(x-y)\Big],\end{equation}
with $Z_x(t)=(1-t)\,e^{x^\top\Lambda x}+t\,e^{x^\top\Lambda y}$ and $Z_y(t)=(1-t)\,e^{y^\top\Lambda x}+t\,e^{y^\top\Lambda y}$ being the partition functions of the two queries associated to $\pi_{\mu_t}^\Lambda$. 
As a consequence we have 
\begin{equation}\label{eq:twopoint}
\frac{\De}{\De t}\Big|_{t=0}\Psi_{\Lambda,V}(\mu_t)=e^{x^\top\Lambda (y-x)}\,x^\top V(y-x)+y^\top Vx-x^\top Vx,\qquad \frac{\De}{\De t}\Big|_{t=0}S_\Lambda(\mu_t)=e^{x^\top\Lambda (y-x)}\,x(y-x)^\top+yx^\top-xx^\top.
\end{equation}
\end{lemma}

\begin{proof}
 Given $x,y\in\mbbS^d$ and $t\in[0,1]$, the corresponding choice rule $\pi^\Lambda_{\mu_t}=\sum_{a,b\in\{x,y\}}P_{ab}\,\delta_a\otimes\delta_b$
can be represented by the matrix
\begin{equation}\label{eq:twopoint-rule}
P=\begin{pmatrix}P_{xx}&P_{xy}\\[2pt]P_{yx}&P_{yy}\end{pmatrix}
=\begin{pmatrix}
(1-t)\,\dfrac{(1-t)\,e^{x^\top\Lambda x}}{Z_x(t)} & (1-t)\,\dfrac{t\,e^{x^\top\Lambda y}}{Z_x(t)}\\[12pt]
t\,\dfrac{(1-t)\,e^{y^\top\Lambda x}}{Z_y(t)} & t\,\dfrac{t\,e^{y^\top\Lambda y}}{Z_y(t)}
\end{pmatrix}\,,
\end{equation}
with rows indexed by the query and columns by the key. Consequently, with $X=(x,y)\in\R^{(d+1)\times2}$,
\begin{equation}\label{eq:twopoint-S}
S_\Lambda(\mu_t)=XPX^\top=(1-t)\,xx^\top+t\,yy^\top+t(1-t)\Big[\frac{e^{x^\top\Lambda y}}{Z_x(t)}\,x(y-x)^\top+\frac{e^{y^\top\Lambda x}}{Z_y(t)}\,y(x-y)^\top\Big],
\end{equation}
which combined with $\Psi_{\Lambda,V}(\mu_t)=\ip{V}{S_\Lambda(\mu_t)}$ shows \eqref{eq:twopoint-exact}. Since $Z_x(0)=e^{x^\top\Lambda x}$ and $Z_y(0)=e^{y^\top\Lambda x}$, differentiating at $t=0$ gives
\[
\frac{\De}{\De t}\Big|_{t=0}S_\Lambda(\mu_t)=\frac{e^{x^\top\Lambda y}}{e^{x^\top\Lambda x}}\,x(y-x)^\top+yx^\top-xx^\top\,,
\]
from which \eqref{eq:twopoint} immediately follows.
\end{proof}

We will see below that the expression of the derivatives in~\eqref{eq:twopoint} can be rearranged in order to take the form 
of combinations of trigonometric polynomials combined with exponentials. The main goal of this section is to show that if linear combinations of those derivatives are null, then the coefficients are equal to zero. The next lemma is the general linear independence result we are going to rely on, 
which states that exponentials of distinct first-degree trigonometric polynomials are linearly independent over trigonometric polynomials.

\begin{lemma}[Exponential polynomials on a circle]\label{lem:exppoly}
Let $(a_1,b_1),\dots,(a_m,b_m)\in\R^2$ be pairwise distinct and let $p_1,\dots,p_m$ be trigonometric polynomials. If $\sum_{k=1}^mp_k(\theta)\,e^{a_k\cos\theta+b_k\sin\theta}=0$ for every $\theta\in\R$, then $p_1=\dots=p_m=0$.
\end{lemma}
\begin{proof}

Put $z=e^{i\theta}$ and $\alpha_k=(a_k-ib_k)/2\in\C$. The map $(a,b)\mapsto(a-ib)/2$ is injective, so the $\alpha_k$ are pairwise distinct. Each trigonometric polynomial $p_k$ can be written $p_k(\theta)=L_k(e^{i\theta})$ for a Laurent polynomial $L_k(z)=\sum_{j=-J}^{J}c_{k,j}z^j$, with $J\in\N$ large enough for all $k$, and $L_k\equiv0$ if and only if $p_k\equiv0$. Consider
\[
H(z)=\sum_{k=1}^mL_k(z)\,\exp\big(\alpha_kz+\bar\alpha_k/z\big),\qquad z\in\C\setminus\{0\}.
\]
Each $L_k$ and each $z\mapsto\alpha_kz+\bar\alpha_k/z$ is holomorphic on $\C\setminus\{0\}$, hence so is $H$. For $\abs z=1$ we have $1/z=\bar z$, so that $\alpha_kz+\bar\alpha_k/z=2\,\mathrm{Re}(\alpha_kz)=a_k\cos\theta+b_k\sin\theta$ for $z=e^{i\theta}$, and therefore $H(e^{i\theta})=\sum_kp_k(\theta)\,e^{a_k\cos\theta+b_k\sin\theta}=0$ for every $\theta\in\R$. 
Since $\C\setminus\{0\}$ is open and connected and $H$ vanishes on the unit circle, which has accumulation points in $\C\setminus\{0\}$, the identity theorem gives $H\equiv0$ on $\C\setminus\{0\}$.

Assume, by contradiction, that not all $L_k$ vanish, and discard the indices with $L_k\equiv0$. We keep the notation $k=1,\dots,m$ for the remaining ones, so that every $L_k$ is a non-zero Laurent polynomial, the $\alpha_k$ are still pairwise distinct, and still $H\equiv0$. 
For every pair $j\ne k$, writing $\alpha_j-\alpha_k=\abs{\alpha_j-\alpha_k}e^{i\phi_{jk}}$, we have $\mathrm{Re}\big((\alpha_j-\alpha_k)e^{i\psi}\big)=\abs{\alpha_j-\alpha_k}\cos(\phi_{jk}+\psi)$, which vanishes for exactly two values of $\psi$ modulo $2\pi$. Choose $\psi\in\R$ outside these finitely many values, 
so that the numbers $\mathrm{Re}(\alpha_ke^{i\psi})$, $k\le m$, are pairwise distinct. Let $k^\star$ be the index at which the largest of them, $M^\star=\mathrm{Re}(\alpha_{k^\star}e^{i\psi})$, is attained, and let
\[
\eta=\min_{k\ne k^\star}\big(M^\star-\mathrm{Re}(\alpha_ke^{i\psi})\big)>0,
\]
with $\eta=+\infty$ if $m=1$.

For every $r>0$ we have $H(re^{i\psi})=0$. Dividing by $e^{\alpha_{k^\star}re^{i\psi}}$ and isolating the term $k^\star$, we obtain
\begin{align*}
L_{k^\star}(re^{i\psi})\,e^{\bar\alpha_{k^\star}/(re^{i\psi})}
&=-\sum_{k\neq k^\star}L_k(re^{i\psi})\,e^{(\alpha_k-\alpha_{k^\star})re^{i\psi}+\bar\alpha_k/(re^{i\psi})}\\
&=-\sum_{k\neq k^\star}L_k(re^{i\psi})\,e^{-r\left(M^\star-\mathrm{Re}(\alpha_ke^{i\psi})\right)}\,e^{\,ir\,\mathrm{Im}\left((\alpha_k-\alpha_{k^\star})e^{i\psi}\right)}\,e^{\bar\alpha_k/(re^{i\psi})},
\end{align*}
where we used $e^w=e^{\mathrm{Re}\,w}\,e^{i\,\mathrm{Im}\,w}$ and $\mathrm{Re}\big((\alpha_k-\alpha_{k^\star})e^{i\psi}\big)=-\big(M^\star-\mathrm{Re}(\alpha_ke^{i\psi})\big)$. 
The middle exponential has modulus one, and $\abs{e^{\bar\alpha_k/(re^{i\psi})}}=e^{\mathrm{Re}(\bar\alpha_ke^{-i\psi})/r}\in\big[e^{-\abs{\alpha_k}/r},e^{\abs{\alpha_k}/r}\big]$. Taking moduli, for $r\ge1$,
\begin{equation}\label{eq:exppoly-domination}
e^{-\abs{\alpha_{k^\star}}}\,\big|L_{k^\star}(re^{i\psi})\big|\le e^{-r\eta}\sum_{k\neq k^\star}e^{\abs{\alpha_k}}\,\big|L_k(re^{i\psi})\big| .
\end{equation}
For each $k$ let $q_k=\max\{j:\ c_{k,j}\ne0\}$, which exists because $L_k\not\equiv0$, and let $q=\max_kq_k$. Then $\abs{L_k(re^{i\psi})}=\abs{c_{k,q_k}}\,r^{q_k}\big(1+O(r^{-1})\big)$ as $r\to\infty$. 
Hence the left-hand side of \eqref{eq:exppoly-domination} is at least $\frac12e^{-\abs{\alpha_{k^\star}}}\abs{c_{k^\star,q_{k^\star}}}\,r^{q_{k^\star}}$ for $r$ large, whereas the right-hand side is $O\big(r^{q}e^{-r\eta}\big)$.
Since $r^{q_{k^\star}-q}e^{r\eta}\to\infty$ as $r\to\infty$, \eqref{eq:exppoly-domination} fails for $r$ large, which is a contradiction. Therefore $L_k\equiv0$ for every $k$, that is, $p_k\equiv0$ for every $k$.         
\end{proof}

Under $\mbfN$ only the sentences of lengths $m\in\supp(\nu)$ occur, so that, when $\supp(\nu)$ is finite, the support $\mcS_{\mbfN}$ is strictly smaller than $\mcP(\mbbS^d)$, 
and a statistic must be shown to be non-constant on the sets $\mcE_m$ of $m$-point empirical measures. Our argument only uses two-point measures $\mu_t=(1-t)\delta_x+t\delta_y$, through the derivative at $t=0$ (see the proof of \Cref{prop:independence} below), 
whereas on $\mcE_m$ only the weights $t\in\{0,\frac1m,\dots,1\}$ occur. The following lemma bridges the gap.

\begin{lemma}[Rational dependence on the weight]\label{lem:rational}
Fix $x,y\in\mbbS^d$ and let $\mu_t=(1-t)\delta_x+t\delta_y$. For $\Lambda\in\Mat$ let $q_\Lambda(t)=\big(1+t(e^{x^\top\Lambda(y-x)}-1)\big)\big(1+t(e^{y^\top\Lambda(y-x)}-1)\big)$, a polynomial of degree at most $2$ which is positive on $[0,1]$. For $\Lambda_1,\dots,\Lambda_H\in\Mat$ and $A_0,A_1,\dots,A_H\in\Mat$, the function
\[
f(t)=\ip{A_0}{m_{\mu_t}m_{\mu_t}^\top}+\sum_{k=1}^H\ip{A_k}{S_{\Lambda_k}(\mu_t)},\qquad t\in[0,1],
\]
is of the form $f=P/\prod_kq_{\Lambda_k}$ with $P$ a polynomial of degree at most $2H+2$, and of degree at most $2H+1$ if $A_0=0$. Consequently, if $m\ge2H+2$ (respectively $m\ge2H+1$ if $A_0=0$) and $f$ is constant on $\{0,\frac1m,\dots,1\}$, then $f$ is constant on $[0,1]$.
\end{lemma}

\begin{proof}
As in the proof of \Cref{lem:twopoint} we have
\[
S_\Lambda(\mu_t)=(1-t)\,xx^\top+t\,yy^\top+t(1-t)\Big[\frac{e^{x^\top\Lambda(y-x)}}{1+t(e^{x^\top\Lambda(y-x)}-1)}\,x(y-x)^\top+\frac{1}{1+t(e^{y^\top\Lambda(y-x)}-1)}\,y(x-y)^\top\Big].
\]
Multiplying by both denominators, each term $q_{\Lambda_k}\ip{A_k}{S_{\Lambda_k}(\mu_t)}$ becomes a polynomial of degree at most $3$, and $\ip{A_0}{m_{\mu_t}m_{\mu_t}^\top}$ is a polynomial of degree at most $2$, since $m_{\mu_t}=x+t(y-x)$. Multiplying by the remaining factors $q_{\Lambda_j}$ gives the degrees $3+2(H-1)=2H+1$ and $2+2H$. If $f=P/R$, with $R=\prod_kq_{\Lambda_k}>0$ on $[0,1]$, equals a constant $c$ at the $m+1$ distinct points $0,\frac1m,\dots,1$, then the polynomial $P-cR$, of degree at most $\max(\deg P,\deg R)$, has at least $m+1$ zeros. It therefore vanishes identically when $m+1$ exceeds this degree, which is at most $2H+2$ (respectively $2H+1$), since $\deg R\le2H$; then $f\equiv c$ on $[0,1]$.
\end{proof}

We are now ready to prove the linear independence of attention cross-moments, from which the non-degeneracy of attention readouts follows.

\begin{proposition}[Linear independence of attention cross-moments]\label{prop:independence}
Let $\Lambda_1,\dots,\Lambda_H\in\Mat\setminus\{0\}$ be pairwise distinct, $A_0,A_1,\dots,A_H\in\Mat$, and $m\ge2H+2$ ($m\ge2H+1$ if $A_0=0$). For $\mu\in\mcP(\mbbS^d)$ let $m_\mu=\int x\,\mu(\De x)$ be its mean. If the function
\[
\mu\longmapsto\ip{A_0}{m_\mu m_\mu^\top}+\sum_{k=1}^H\ip{A_k}{S_{\Lambda_k}(\mu)}
\]
is constant on $\mcE_m$, then $A_1=\dots=A_H=0$ and $A_0$ is antisymmetric. In particular, the same conclusion holds if the function is constant on $\mcP(\mbbS^d)$.
\end{proposition}

\begin{proof}
Fix $x,y\in\mbbS^d$, let $\mu_t=(1-t)\delta_x+t\delta_y$ and let $f(t)=\ip{A_0}{m_{\mu_t} m_{\mu_t}^\top}+\sum_{k=1}^H\ip{A_k}{S_{\Lambda_k}(\mu_t)}$, $t\in[0,1]$. Since $\mu_{j/m}=(1-\frac jm)\delta_x+\frac jm\delta_y\in\mcE_m$ for $0\le j\le m$, $f$ is constant on $\{0,\frac1m,\dots,1\}$, hence on $[0,1]$ by \Cref{lem:rational}. By \Cref{lem:twopoint}, $f$ is differentiable, so that $f'(0)=0$. Since $m_{\mu_t}=x+t(y-x)$ and $m^\top A_0m=m^\top(A_0)_sm$, with $(A_0)_s=\frac12(A_0+A_0^\top)$, the moment term contributes $2x^\top(A_0)_s(y-x)$ to $f'(0)$, while by \eqref{eq:twopoint} each attention term contributes $e^{x^\top\Lambda_k(y-x)}x^\top A_k(y-x)+y^\top A_kx-x^\top A_kx$. Hence, for all $x,y\in\mbbS^d$,
\begin{equation}\label{eq:indep-proof}
0=2x^\top(A_0)_s(y-x)+\sum_{k=1}^H\Big[e^{x^\top\Lambda_k(y-x)}\,x^\top A_k(y-x)+y^\top A_kx-x^\top A_kx\Big].
\end{equation}
Consider the set
\[
\mcX\coloneqq \big\{x\in\mbbS^d\,\colon\, \Lambda_k^\top x\neq 0\ \ \forall k\le H,\ \text{ and }\ (\Lambda_k-\Lambda_j)^\top x\ne0\ \ \forall j\neq k\le H\big\}.
\]
It is open and dense in $\mbbS^d$, since it is the complement of finitely many proper linear subspaces. Fix $x\in\mcX$, a unit vector $v\perp x$, and let $y=\cos\theta\,x+\sin\theta\,v\in\mbbS^d$, $\theta\in\R$. With $(a_k,b_k)\coloneqq(x^\top\Lambda_kx,\,x^\top\Lambda_kv)$ we have $x^\top\Lambda_k(y-x)=(\cos\theta-1)a_k+\sin\theta\,b_k$, and \eqref{eq:indep-proof} becomes
\[
\sum_{k=1}^He^{-a_k}\,e^{a_k\cos\theta+b_k\sin\theta}\Big[(\cos\theta-1)\,x^\top A_kx+\sin\theta\,x^\top A_kv\Big]+p_0(\theta)=0,\qquad\theta\in\R,
\]
where $p_0$ is a trigonometric polynomial, which collects the terms without exponential factors.

We now check that \Cref{lem:exppoly} applies. Its exponent pairs are $(a_k,b_k)$, $k\le H$, and $(0,0)$ for $p_0$, and they must be pairwise distinct: we need $(a_k,b_k)\ne(0,0)$ and $(a_k,b_k)\ne(a_j,b_j)$ for $j\ne k$. Both conditions have the form $(w^\top x,w^\top v)\ne(0,0)$, with $w=\Lambda_k^\top x$ and $w=(\Lambda_k-\Lambda_j)^\top x$ respectively, and $w\ne0$ because $x\in\mcX$. For such $w$, the pair $(w^\top x,w^\top v)$ vanishes only if $w\perp x$ and $w\perp v$. In that case $w\in x^\perp\setminus\{0\}$, and the unit vectors $v\in x^\perp$ with $v\perp w$ form a great subsphere of the unit sphere of $x^\perp$, which is closed with empty interior (and is empty when $d=1$). Since there are finitely many such $w$, the exponent pairs are pairwise distinct for every $v$ in an open dense subset of the unit sphere of $x^\perp$.

For such $v$, \Cref{lem:exppoly} shows that all the coefficients vanish identically, in particular $(\cos\theta-1)\,x^\top A_kx+\sin\theta\,x^\top A_kv=0$ for all $\theta$. Since $\cos\theta-1$ and $\sin\theta$ are linearly independent, $x^\top A_kx=x^\top A_kv=0$. By continuity in $v$, $x^\top A_kv=0$ for every unit $v\perp x$, so that the vector $A_k^\top x$ is orthogonal both to $x$ and to $x^\perp$, that is, $A_k^\top x=0$. This holds for every $x\in \mcX$, which is dense in $\mbbS^d$, hence for every $x\in\mbbS^d$ by continuity, and $A_k=0$ for every $k$.

Then \eqref{eq:indep-proof} reduces to $x^\top(A_0)_s(y-x)=0$ for all $x,y\in\mbbS^d$. With $y=-x$ it gives $x^\top(A_0)_sx=0$, hence $x^\top(A_0)_sy=0$ for all $x,y\in\mbbS^d$ and, by homogeneity, for all $x,y\in\R^{d+1}$, so that $(A_0)_s=0$, which is equivalent to $A_0$ being antisymmetric. The last claim follows since $\mcE_m\subset\mcP(\mbbS^d)$.
\end{proof}

Non-degeneracy holds for every non-zero choice matrix $\Lambda$, with no symmetry assumption: the proof only uses that $\{x:\ \Lambda_k^\top x\ne0\}$ is dense, which holds for every $\Lambda_k\ne0$, including antisymmetric ones. In particular $S_\Lambda$ and $S_{\Lambda^\top}$ are linearly independent when $\Lambda\ne\Lambda^\top$, in contrast with the kernel sums of \Cref{prop:additive}-\ref{item:d}. This explains why our model admits a query--key matrix $\Lambda=Q^\top K$ that is not necessarily symmetric. The mechanism is the softmax normalization: it produces the exponential factors in \eqref{eq:indep-proof}, which cannot be compensated by trigonometric polynomials (\Cref{lem:exppoly}).

\begin{corollary}[Non-degeneracy]\label{cor:nondeg}
  \,
\begin{enumerate}[(a)]
\item\label{item:nondeg:a} Let $\Lambda\ne0$. If $m\ge3$ and $\ip{A}{S_\Lambda}$ is constant on $\mcE_m$, then $A=0$. Consequently, if $\nu(m)>0$ for some $m\ge3$, then $S_\Lambda$ is $\nu$-non-degenerate: $\ip{A}{S_\Lambda}$ is constant on $\mcS_{\mbfN}$ only for $A=0$. The same holds on $\mcP(\mbbS^d)$.
\item\label{item:nondeg:b} Let $\Lambda=0$. If $m\ge2$, then $\ip{A}{S_0}$ is constant on $\mcE_m$ if and only if $A$ is antisymmetric. Consequently, if $\nu(m)>0$ for some $m\ge2$, then $\ip{A}{S_0}$ is constant on $\mcS_{\mbfN}$ if and only if $A$ is antisymmetric.
\end{enumerate}
\end{corollary}

\begin{proof}
 The first claim in \ref{item:nondeg:a} is \Cref{prop:independence} with $H=1$ and $A_0=0$, since $m\ge3=2H+1$. If $\ip{A}{S_\Lambda}$ is constant on $\mcS_{\mbfN}$, it is constant on $\mcE_m$ for every $m\in\supp(\nu)$, since $\mcE_m\subset\mcS_{\mbfN}$, and the first claim applies; the last claim follows from $\mcE_m\subset\mcP(\mbbS^d)$.

Claim~\ref{item:nondeg:b} can be proven as follows. For $\Lambda=0$ the choice rule is $\mu\otimes\mu$, so that $\Attn_0(x;\mu)=m_\mu$ and $S_0(\mu)=m_\mu m_\mu^\top$. Hence $\ip{A}{S_0(\mu)}=m_\mu^\top Am_\mu=m_\mu^\top A_sm_\mu$, with $A_s=\frac12(A+A^\top)$, 
which vanishes identically if $A$ is antisymmetric. Conversely, assume that $\ip{A}{S_0}=c$ on $\mcE_m$. At $\mu=\delta_x\in\mcE_m$ this gives $x^\top A_sx=c$ for every $x\in\mbbS^d$, so that $A_s=c\Id$. At $\mu=(1-\frac1m)\delta_x+\frac1m\delta_{-x}\in\mcE_m$, whose mean $(1-\frac2m)x$ has norm $\abs{1-\frac2m}<1$, 
it gives $c(1-\frac2m)^2=c$, so that $c=0$ and $A_s=0$. The last claim follows as in~\ref{item:nondeg:a}.
\end{proof}

\section{Finite vocabularies and the unembedding}\label{app:vocab}

In the body of the paper tokens are points of the sphere drawn from the uniform law $\uvol$, and the unembedding of \Cref{sec:unembedding} is obtained by projecting the output of the Langevin scheme onto the vocabulary through a non-linearity $\sigma$. In this appendix we show that, if the null model draws its tokens directly from a finite vocabulary, the causal model of \Cref{sec:shifted} gives the standard softmax unembedding exactly, with weight tying and without any sampler or non-linearity.

\medskip

Let $\mcV=\{v_1,\dots,v_{\abs\mcV}\}\subset\mbbS^d$ be a finite vocabulary of embedded tokens and let $\omega$ be a probability on $\mcV$ with $\omega(v)>0$ for every $v\in\mcV$ (e.g., the uniform law). Then, in this setting, the discrete null model $\arrowN_{\mcV}$ is the law of the ordered sentence $\iota(N,X_{1:N})$ when $N\sim\nu$ and, independently, $X_1,X_2,\dots$ are i.i.d.\ with law $\omega$. It is concentrated on the countable set $\iota\big(\bigsqcup_{n\ge1}\mcV^n\big)$, and
\[
\arrowN_{\mcV}\big(N=n,\ X_{1:n}=x_{1:n}\big)=\nu(n)\prod_{i=1}^n\omega(x_i),\qquad x_{1:n}\in\mcV^n .
\]
The masked choice rule, the shifted cross-moment $S^\to_\Lambda$ and the readout $\Psi^\to_{\Lambda,V}=\ip{V}{S^\to_\Lambda}$ are defined as in \Cref{sec:shifted}, and for $\Lambda,V\in\Mat$ we set $\mrmP^\to_{\Lambda,V;\mcV}\propto e^{\Psi^\to_{\Lambda,V}}\,\arrowN_{\mcV}$.
Finally, recall from \Cref{sec:unembedding} the definition of the unembedding matrix $W_{\mathrm{out}}\in\R^{\abs\mcV\times(d+1)}$, whose rows are $v_1^\top,\dots,v_{\abs\mcV}^\top$, so that $(W_{\mathrm{out}}\widehat{Y})_v=\ip{v}{\widehat{Y}}$ for $\widehat{Y}\in\R^{d+1}$.

\begin{proposition}[Finite vocabulary: exact softmax unembedding]\label{prop:vocab}
Let $\Lambda,V\in\Mat$.
\begin{enumerate}[(a)]
\item\label{item:vocab:a} For every $n\ge2$ with $\nu(n)>0$ and every context $x_{1:n-1}\in\mcV^{n-1}$, for any $v\in\mcV$
\[
\mrmP^\to_{\Lambda,V;\mcV}\big(X_n=v\,\big|\,N=n,\ X_{1:n-1}=x_{1:n-1}\big)=\mathrm{softmax}\Big(\tfrac1{n-1}\,W_{\mathrm{out}}V\AttM{n-1}+\log\omega\Big)_v,
\]
where $\log\omega=(\log\omega(v_1),\dots,\log\omega(v_{\abs\mcV}))$ is applied componentwise.
\item\label{item:vocab:b} Assume that $\nu(n)>0$ for some $n\ge2$ and that $\mcV$ is not contained in an affine hyperplane of $\R^{d+1}$. If $\ip{A}{S^\to_\Lambda}$ is $\arrowN_{\mcV}$-a.s.\ constant for some $A\in\Mat$, then $A=0$.
\item\label{item:vocab:c} Under the assumptions of~\ref{item:vocab:b}, let $\mcK^\to_{\Lambda;\mcV}$ be the closed convex hull of the support of $(S^\to_\Lambda)_\#\arrowN_{\mcV}$. For every $c\in\inter\mcK^\to_{\Lambda;\mcV}$ there is a unique $V^\star\in\Mat$ with $\mbbE_{\mrmP^\to_{\Lambda,V^\star;\mcV}}S^\to_\Lambda=c$, and $\mrmP^\to_{\Lambda,V^\star;\mcV}$ is the unique minimizer of $\rel{\mrmP}{\arrowN_{\mcV}}$ over the laws $\mrmP$ with $\mbbE_\mrmP S^\to_\Lambda=c$.
\end{enumerate}
\end{proposition}

\begin{proof}
Throughout, $n\ge2$ and we use the explicit form of the readout from \Cref{def:shifted},
\begin{equation}\label{eq:vocab:readout}
\Psi^\to_{\Lambda,V}(x_{1:n})=\frac1{n-1}\sum_{i=1}^{n-2}x_{i+1}^\top V\AttM{i}\;+\;\frac1{n-1}\,x_n^\top V\AttM{n-1}.
\end{equation}
By \eqref{eq:shifted-facts}, $\AttM{i}$ depends only on $x_{1:i}$. Hence the first sum depends only on the context $x_{1:n-1}$, and the last token $x_n$ enters only through the last term, linearly.

Since, by definition of $\mrmP^\to_{\Lambda,V;\mcV}$ and of $\arrowN_{\mcV}$, for $x_{1:n}\in\mcV^n$ we have
\[
\mrmP^\to_{\Lambda,V;\mcV}\big(N=n,\ X_{1:n}=x_{1:n}\big)\;\propto\;\nu(n)\prod_{i=1}^n\omega(x_i)\;e^{\Psi^\to_{\Lambda,V}(x_{1:n})},
\]
for any fixed context $x_{1:n-1}$ and any $v\in\mcV$ it holds
\begin{equation*}
  \begin{aligned}
\mrmP^\to_{\Lambda,V;\mcV}\big(X_n=v\,\big|\,N=n,\ X_{1:n-1}=x_{1:n-1}\big)\;\propto&\;\omega(v)\,\exp\Big(\tfrac1{n-1}\,v^\top V\AttM{n-1}\Big)\\
\propto &\,\exp\Big(\tfrac1{n-1}\big(W_{\mathrm{out}}V\AttM{n-1}\big)_v+\log\omega(v)\Big).
  \end{aligned}
\end{equation*}
Normalizing over $v\in\mcV$ gives \ref{item:vocab:a}.

\smallskip
Item~\ref{item:vocab:b} can be proven as follows. Fix $n\ge2$ with $\nu(n)>0$, and assume that $\ip{A}{S^\to_\Lambda}$ is constant $\arrowN_{\mcV}$-a.s., and therefore it is constant
for every $x_{1:n}\in\mcV^n$ ($\arrowN_{\mcV}$ gives positive mass $\nu(n)\prod_i\omega(x_i)$ to every $x_{1:n}\in\mcV^n$).

Now, if we apply \eqref{eq:vocab:readout} with $V=A$ and $x_{1:n-1}\in\mcV^{n-1}$ fixed, since $\ip{A}{S^\to_\Lambda}$ is constant, for any $v,v'\in\mcV$ we immediately get 
$(v-v')^\top A\, \AttM{n-1}=0$. Since $\mcV$ is not contained in an affine hyperplane, the differences $\{v-v_1:\ v\in\mcV\}$ span $\R^{d+1}$, which means that $A\,\AttM{n-1}=0$.

When considering the constant context $x_1=\dots=x_{n-1}=v\in\mcV$, where every key of the query $(v,n-1)$ equals $v$ so that $\AttM{n-1}=v$, we conclude that $Av=0$.
This is enough to conclude that $A=0$: indeed $\mcV$ is not contained in any affine hyperplane and therefore it spans the whole $\mbbR^{d+1}$, which means that $A$ vanishes on $\R^{d+1}$.
This concludes the proof of~\ref{item:vocab:b}.

\smallskip
The proof of Claim~\ref{item:vocab:c} is similar to that of \Cref{thm:iproj} and relies on \cite[Theorem~3.6]{brown1986}.
Since $\norm{S^\to_\Lambda}_\Fr\le1$, the log-partition function $\Gamma_{\mcV}(V)=\log\mbbE_{\arrowN_{\mcV}}e^{\ip{V}{S^\to_\Lambda}}$ is finite on the whole of $\Mat$, which is open, so the exponential family $\{\mrmP^\to_{\Lambda,V;\mcV}\}_{V\in\Mat}$ is regular, hence steep. By~\ref{item:vocab:b} it is minimal. Then \cite[Theorem~3.6]{brown1986} shows that $V\mapsto\nabla\Gamma_{\mcV}(V)=\mbbE_{\mrmP^\to_{\Lambda,V;\mcV}}S^\to_\Lambda$ is a bijection of $\Mat$ onto $\inter\mcK^\to_{\Lambda;\mcV}$, which gives existence and uniqueness of $V^\star$.
Set $\mrmP^\star=\mrmP^\to_{\Lambda,V^\star;\mcV}$. Its density $e^{\ip{V^\star}{S^\to_\Lambda}-\Gamma_{\mcV}(V^\star)}$ with respect to $\arrowN_{\mcV}$ is bounded away from $0$ and $\infty$, so $\mrmP^\star$ and $\arrowN_{\mcV}$ are equivalent. Let $\mrmP$ satisfy the constraint with $\rel{\mrmP}{\arrowN_{\mcV}}<\infty$. Then $\mrmP\ll\mrmP^\star$, and integrating $\log\frac{\De\mrmP}{\De\arrowN_{\mcV}}=\log\frac{\De\mrmP}{\De\mrmP^\star}+\ip{V^\star}{S^\to_\Lambda}-\Gamma_{\mcV}(V^\star)$ against $\mrmP$ gives the Pythagorean identity
\[
\rel{\mrmP}{\arrowN_{\mcV}}=\rel{\mrmP}{\mrmP^\star}+\ip{V^\star}{c}-\Gamma_{\mcV}(V^\star)=\rel{\mrmP}{\mrmP^\star}+\rel{\mrmP^\star}{\arrowN_{\mcV}} .
\]
From this last identity we conclude that $\rel{\mrmP}{\arrowN_{\mcV}}\ge\rel{\mrmP^\star}{\arrowN_{\mcV}}$, with equality if and only if $\mrmP=\mrmP^\star$.
\end{proof}

\begin{remark}[On the assumptions and on weight tying]\label{rem:vocab}\,
\begin{enumerate}[(i)]
\item The geometric assumption of~\ref{item:vocab:b} requires $\abs\mcV\ge d+2$, and it holds as soon as the embedded vocabulary is not contained in an affine hyperplane, which is the generic situation since in practice $\abs\mcV\gg d+1$. 
\item In~\ref{item:vocab:a} the unembedding matrix is the matrix of the vocabulary itself. If the vocabulary is obtained from one-hot vectors through an embedding matrix $E\in\R^{(d+1)\times\abs\mcV}$ followed by normalization, $v_i=Ee_i/\abs{Ee_i}$, then $W_{\mathrm{out}}=\mathrm{diag}\big(\abs{Ee_1},\dots,\abs{Ee_{\abs\mcV}}\big)^{-1}E^\top$, i.e.\ the unembedding is tied to the embedding. The reference law $\omega$ only adds the output bias $\log\omega$, which vanishes, up to an additive constant, for the uniform law.
\item The same proof applies verbatim to the extensive statistic $T^\to_\Lambda=(n-1)S^\to_\Lambda$ of \Cref{prop:rescaled}, under the exponential-moment assumption on $\nu$ made there. The logits then become $W_{\mathrm{out}}V\AttM{n-1}+\log\omega$, without the factor $\frac1{n-1}$.
\end{enumerate}
\end{remark}

\bibliographystyle{alpha}
\bibliography{biblio_final}

\end{document}

%% file: header.tex
\usepackage[T1]{fontenc}	
\usepackage{totpages}
\usepackage{graphicx}
\usepackage{amsmath, amsthm, amssymb}
\usepackage{upgreek}
\usepackage{mathrsfs}
\usepackage{pdfsync}
\usepackage[a4paper,top=3cm,bottom=3cm,left=2.cm,right=2.cm, bindingoffset=0mm]{geometry}
\usepackage{cite}
\usepackage[dvipsnames]{xcolor}
\usepackage[inline,shortlabels]{enumitem}
\usepackage[hyphens]{url}									%per gli url in bliografia
\usepackage{verbatim}								%per commentare cose lunghe
\usepackage{dsfont}									%doublestroke \mathds
\usepackage{bbm}
\usepackage{bm}
\usepackage[all]{xy}

\usepackage{mathtools}
\usepackage{xparse}
\usepackage{microtype}

\usepackage{subcaption}
\usepackage{graphicx}
\usepackage[font=small]{caption}
\usepackage{nicefrac}
\usepackage{longtable}

\usepackage{booktabs,array}

\usepackage[pdftex,bookmarks,
						hidelinks,					%nasconde i link: antagonista di colorlinks (s 1 E no 2)
						colorlinks,				%link a colori invece che riquadrati
						breaklinks,%unicode,
						citecolor=OliveGreen,
						linkcolor=Maroon,
						]{hyperref} %hyperref deve stare per ultimo

\usepackage{mathtools}

\usepackage{aliascnt}
\usepackage{cleveref}
\usepackage{enumitem}

\RequirePackage{silence}
\allowdisplaybreaks

\usepackage[textwidth=1.5cm]{todonotes}

\numberwithin{equation}{section}

\theoremstyle{plain}
\newtheorem{theorem}{Theorem}[section]
\newtheorem*{theorem*}{Theorem}
\crefname{theorem}{theorem}{theorems}
\Crefname{theorem}{Theorem}{Theorems}

\newaliascnt{proposition}{theorem}
\newtheorem{proposition}[proposition]{Proposition}
\aliascntresetthe{proposition}
\crefname{proposition}{proposition}{propositions}
\Crefname{proposition}{Proposition}{Propositions}

\newaliascnt{lemma}{theorem}
\newtheorem{lemma}[lemma]{Lemma}
\aliascntresetthe{lemma}
\crefname{lemma}{lemma}{lemmas}
\Crefname{lemma}{Lemma}{Lemmata}

\newaliascnt{corollary}{theorem}
\newtheorem{corollary}[corollary]{Corollary}
\aliascntresetthe{corollary}
\crefname{corollary}{corollary}{corollaries}
\Crefname{corollary}{Corollary}{Corollaries}

\newaliascnt{definition}{theorem}
\newtheorem{definition}[definition]{Definition}
\aliascntresetthe{definition}
\crefname{definition}{definition}{definitions}
\Crefname{definition}{Definition}{Definitions}

\newaliascnt{notation}{theorem}

\aliascntresetthe{notation}
\crefname{notation}{notation}{notations}
\Crefname{notation}{Notation}{Notations}

\newtheorem*{defs*}{Definition}

\crefname{formal}{formal statement}{formal statements}
\Crefname{formal}{Formal statement}{Formal statements}	

\newaliascnt{conj}{theorem}

\aliascntresetthe{conj}	
\crefname{conj}{conjecture}{conjectures}
\Crefname{conj}{Conjecture}{Conjectures}

\newaliascnt{example}{theorem}
\newtheorem{example}[theorem]{Example}%[section]
\aliascntresetthe{example}
\crefname{example}{example}{examples}
\Crefname{example}{Example}{Examples}

\newaliascnt{remark}{theorem}
\newtheorem{remark}[remark]{Remark}%[section]
\aliascntresetthe{remark}
\crefname{remark}{remark}{remarks}
\Crefname{remark}{Remark}{Remarks}

\crefname{assumption}{\textbf{H}}{\textbf{H}}
\newtheorem*{ass*}{Assumption}%[section]

\crefname{mainthm}{Theorem}{Theorems}

\newtheorem*{metathm}{Meta-Theorem}
\crefname{mainthm}{Meta-Theorem}{Meta-Theorems}

\renewcommand{\paragraph}[1]{\medskip\emph{#1}.\quad}

\newcommand{\bes}{\begin{equation*}}
\newcommand{\ees}{\end{equation*}}
\newcommand{\beas}{\begin{eqnarray}}
\newcommand{\eeas}{\end{eqnarray}}
\newcommand{\bea}{\begin{eqnarray}}
\newcommand{\eea}{\end{eqnarray}}
\newcommand{\be}{\begin{equation}}
\newcommand{\ee}{\end{equation}}
\newcommand{\bei}{\begin{itemize}}
\newcommand{\eei}{\end{itemize}}
\newcommand{\bec}{\begin{cases}}
\newcommand{\eec}{\end{cases}}
\newcommand{\ben}{\begin{enumerate}}
\newcommand{\een}{\end{enumerate}}

\usepackage{xparse}

\ExplSyntaxOn
\NewDocumentCommand{\makeabbrev}{mmm}
 {
  \yoruk_makeabbrev:nnn { #1 } { #2 } { #3 }
 }

\cs_new_protected:Npn \yoruk_makeabbrev:nnn #1 #2 #3
 {
  \clist_map_inline:nn { #3 }
   {
    \cs_new_protected:cpn { #2 } { #1 { ##1 } }
   }
 }
 \ExplSyntaxOff

\makeabbrev{\textbf}{tbf#1}{a,b,c,d,e,f,g,h,i,j,k,l,m,n,o,p,q,r,s,t,u,v,w,x,y,z,A,B,C,D,E,F,G,H,I,J,K,L,M,N,O,P,Q,R,S,T,U,V,W,X,Y,Z}

\makeabbrev{\textbf}{bf#1}{a,b,c,d,e,f,g,h,i,j,k,l,m,n,o,p,q,r,s,t,u,v,w,x,y,z,A,B,C,D,E,F,G,H,I,J,K,L,M,N,O,P,Q,R,S,T,U,V,W,X,Y,Z}

\makeabbrev{\textsf}{tsf#1}{a,b,c,d,e,f,g,h,i,j,k,l,m,n,o,p,q,r,s,t,u,v,w,x,y,z,A,B,C,D,E,F,G,H,I,J,K,L,M,N,O,P,Q,R,S,T,U,V,W,X,Y,Z}

\makeabbrev{\mathsf}{mss#1}{a,b,c,d,e,f,g,h,i,j,k,l,m,n,o,p,q,r,s,t,u,v,w,x,y,z,A,B,C,D,E,F,G,H,I,J,K,L,M,N,O,P,Q,R,S,T,U,V,W,X,Y,Z}

\makeabbrev{\mathfrak}{mf#1}{a,b,c,d,e,f,g,h,i,j,k,l,m,n,o,p,q,r,s,t,u,v,w,x,y,z,A,B,C,D,E,F,G,H,I,J,K,L,M,N,O,P,Q,R,S,T,U,V,W,X,Y,Z,
sl,gl
}

\makeabbrev{\mathrm}{mrm#1}{a,b,c,d,e,f,g,h,i,j,k,l,m,n,o,p,q,r,s,t,u,v,w,x,y,z,A,B,C,D,E,F,G,H,I,J,K,L,M,N,O,P,Q,R,S,T,U,V,W,X,Y,Z}

\makeabbrev{\mathbf}{mbf#1}{a,b,c,d,e,f,g,h,i,j,k,l,m,n,o,p,q,r,s,t,u,v,w,x,y,z,A,B,C,D,E,F,G,H,I,J,K,L,M,N,O,P,Q,R,S,T,U,V,W,X,Y,Z}

\makeabbrev{\mathcal}{mc#1}{A,B,C,D,E,F,G,H,I,J,K,L,M,N,O,P,Q,R,S,T,U,V,W,X,Y,Z}

\makeabbrev{\mathbb}{mbb#1}{A,B,C,D,E,F,G,H,I,J,K,L,M,N,O,P,Q,R,S,T,U,V,W,X,Y,Z}

\makeabbrev{\mathscr}{ms#1}{A,B,C,D,E,F,G,H,I,J,K,L,M,N,O,P,Q,R,S,T,U,V,W,X,Y,Z}

\makeabbrev{\mathrm}{#1}{
id,ran,rk,diag,stab,ann,conv,pr,ev,tr,End,Hom,sgn,im,op,can,fin,ext,red,tot,lex,Aut,Inn,unit,
rot,usc,lsc,Lip,lip,bSymLip,osc,AC,loc,coz,z,
supp,Opt,Adm,Cpl,Geo,GeoOpt,GeoAdm,GeoCpl,reg,res,
bd,co,Ric,Exp,dExp,dist,seg,Seg,cut,fcut,Cut,SDiff,Iso,Isom,diam,cl,Homeo,Diff,Der,vol,dvol,inj,relint, Graph, sub,
var,law,Var,Poi,Gam,pa,so,iso,fs,inv,pqi,mix,erg,form,
TestF,
ob,cod,inp,
}

\makeabbrev{\mathsf}{#1}{CD,BE,MCP,Ent,wMTW,MTW,Ch,RCD,EVI,Rad,dRad,SL,cSL,dSL,ScL,Irr,SC,wFe,VA,MetMeas,UMeas,CSMet,Met,USp,Meas,Mbl,alg,Alg}

\makeabbrev{\mathsc}{#1}{mmaf,cg}

\newcommand{\inter}{\mathrm{int}}

\renewcommand{\iint}{\int\!\!\!\!\!\int}

\newcommand{\mathsc}[1]{\text{\textsc{#1}}}

\newcommand{\abs}[1]{\left\lvert#1\right\rvert}						%Modulo
\newcommand{\norm}[1]{\left\lVert#1\right\rVert}					%Norma
\newcommand{\N}{{\mathbb N}}
\newcommand{\R}{{\mathbb R}}

\newcommand{\C}{{\mathbb C}} 	%renew perchè \C è un comando nell'estensione "unicode" del pacchetto hyperref: togliendo l'estensione (o il pacchetto) si può usare newcommand

\newcommand{\Id}{\mssI}

\usepackage{scrextend}						%KOMA: confligge quindi deve stare qui

\let\temp\phi
\let\phi\varphi
\let\varphi\temp

\let\temp\epsilon
\let\epsilon\varepsilon
\let\varepsilon\temp

\newcommand{\De}{\mathrm{d}}

\newcommand{\KL}{\mathrm{KL}}
\newcommand{\Cov}{\mathrm{Cov}}

\newcommand{\ip}[2]{\langle #1,#2\rangle}
\newcommand{\rel}[2]{\KL(#1\,|\,#2)}

\DeclareMathOperator*{\argmax}{argmax}

\newcommand{\uvol}{\underline{\vol}}

\newcommand{\Attn}{\mathrm{Attn}}
\newcommand{\Mat}{\mathbb{M}_D}
\newcommand{\Fr}{\mathrm{F}}

\newcommand{\arrowmu}{\overset{\rightarrow}{\mu}}
\newcommand{\arrowN}{\overset{\rightarrow}{\mbfN}}

\renewcommand{\Mat}{\mbbM_{d+1}}

\newcommand{\AttM}[2][\Lambda]{\Attn_{#1}((x_{#2},#2);\arrowmu)}